\documentclass[dissertacao]{theme/inf-ufg}

\usepackage{amsmath, amssymb}
\usepackage{theme/math-notation}
\pgfplotsset{compat=1.18}

\usepackage{booktabs}
\usepackage{pgfgantt}

\usepackage{pifont}
\newcommand{\cmark}{\ding{51}}

\usepackage{graphicx}
\usepackage{multirow}
\usepackage{pgf-pie}
\usepackage{makecell}
\usepackage{tabularx}
\usepackage{xurl}
\usepackage[frozencache=true,cachedir=minted-cache]{minted}
\usepackage{footnote}
\usepackage{svg}
\usepackage{tcolorbox}
\usepackage{soul}
\usepackage{ulem}
\usepackage{scalerel}
\usepackage{pdfpages}
\usepackage{siunitx}
\usepackage{threeparttable}

\usepackage{multicol}

\usepackage{csquotes}
\MakeOuterQuote{"}

\newcommand{\novo}[1]{#1}
\newcommand{\correcaoPrevia}[1]{#1}
\newcommand{\correcao}[1]{#1}

\begin{document}

\autor{Juliana Resplande Sant'anna Gomes}
\autorR{Gomes, Juliana} 

\titulo{\textbf{Verificação Semi-Automática de Fatos em Português}: Enriquecimento de \textit{Corpus} via Busca e Extração de Alegação}

\cidade{Goiânia}
\dia{10}\mes{04}\ano{2025}

\orientador{Arlindo Rodrigues Galvão Filho}
\orientadorR{Filho, Arlindo}
\universidade{Universidade Federal de Goiás (UFG)}
\uni{UFG}
\unidade{Instituto de Informática}
\departamento{}
\programa{Programa de Pós-Graduação em Ciência da Computação}
\area{Ciência da Computação}
\concentracao{Sistemas Inteligentes e Aplicações}

\capa
\includepdf{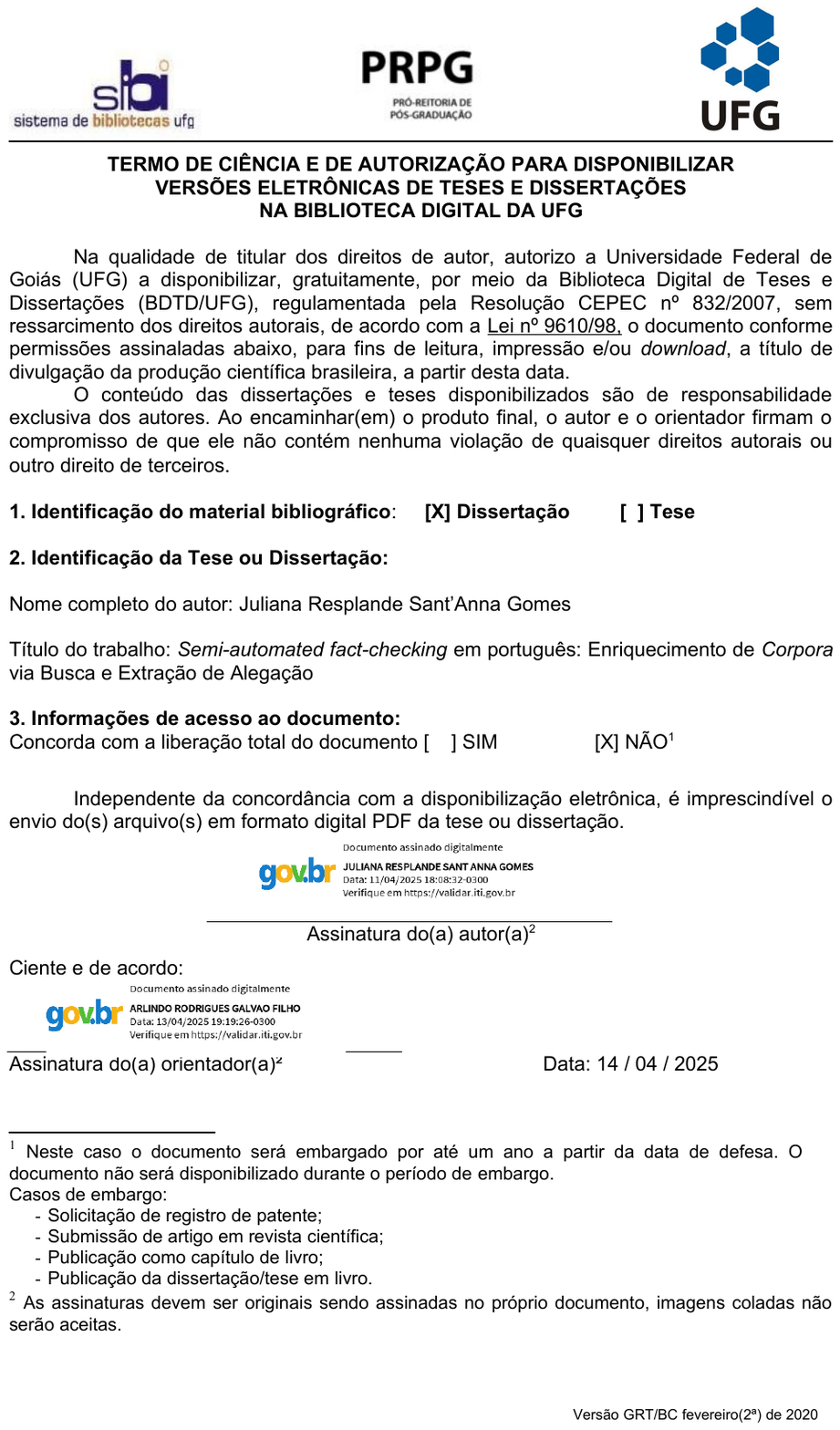}
\rosto 
\includepdf{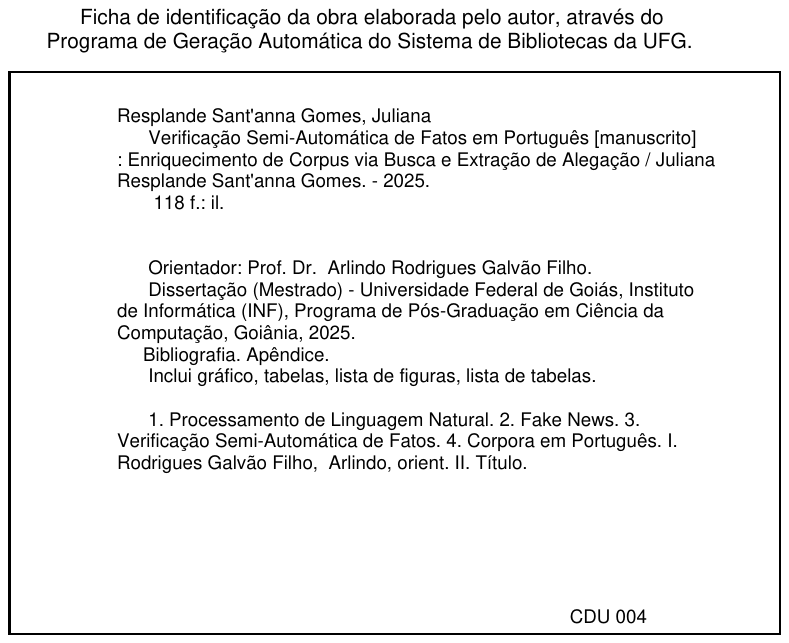}

\begin{aprovacao}
\profa{Telma Woerle de Lima Soares}{Instituto de Informática -- UFG}
\banca{Eliomar Araújo de Lima}{Instituto de Informática -- UFG}
\end{aprovacao}

\direitos{Mestranda pela Universidade Federal de Goiás (UFG), filiada ao Centro de Excelência em Inteligência Artificial (CEIA). Meus interesses de pesquisa são obtenção e tradução de corpora ao português, verificação de fake news, sistemas de Question Answering, geração de texto.}

\begin{dedicatoria}
 Dedico este trabalho à minha mãe, que com imensa força e cumplicidade, esteve ao meu lado em cada desafio, formando comigo o pilar central da nossa jornada.
\end{dedicatoria}

\begin{agradecimentos}
A conclusão desta dissertação de mestrado marca o fim de uma jornada desafiadora e gratificante, e seria impossível alcançá-la sem o apoio e a contribuição de diversas pessoas e instituições a quem desejo expressar minha mais sincera gratidão.

Em primeiro lugar, agradeço profundamente ao meu orientador, Prof. Dr. Arlindo Rodrigues Galvão Filho, pela orientação segura, pelos valiosos ensinamentos, pela paciência e pela confiança depositada em meu trabalho ao longo desta caminhada. Sua expertise foi fundamental para o desenvolvimento desta pesquisa. Aos membros da banca examinadora, Prof. Dr. Eliomar Araújo de Lima e Prof.ª Dr.ª Telma Woerle de Lima Soares, minha gratidão pelas leituras atentas, críticas construtivas e sugestões perspicazes que enriqueceram significativamente este estudo e abriram novos horizontes para reflexão.

Meu reconhecimento se estende ao Centro de Competência EMBRAPII em Tecnologias Imersivas (Advanced Knowledge Center for Immersive Technologies - AKCIT) e ao Centro de Excelência em Inteligência Artificial (CEIA). O apoio institucional, as discussões estimulantes e a colaboração frutífera foram cruciais para o avanço e a concretização desta pesquisa. Sou igualmente grata ao Instituto de Informática da Universidade Federal de Goiás (INF-UFG) e à Universidade Federal de Goiás (UFG) como um todo, por fornecerem a infraestrutura, os recursos e o ambiente acadêmico propício para a realização deste mestrado.

No âmbito pessoal, palavras são insuficientes para expressar minha eterna gratidão à minha mãe, por seu amor incondicional, por ser a base da minha formação e por sempre me incentivar tanto na esfera profissional quanto pessoal. Seus sacrifícios e apoio foram meu alicerce.

Ao meu companheiro de mestrado e de vida, Eduardo Augusto Santos Garcia, meu mais profundo agradecimento. Sua compreensão, apoio constante e as valiosas revisões e discussões ao longo de todo o processo, foram essenciais. Sua presença tornou cada desafio mais leve.

Aos meus amigos, pela amizade leal, pelo incentivo vibrante, pelas palavras de conforto nos momentos de dificuldade e pela celebração de cada conquista. Vocês foram um refúgio e uma fonte de energia.
\end{agradecimentos}

\epigrafe{"Nós não estamos somente a pandemia, nós estamos combatendo uma infodemia."}
{Tedros Adhanom Ghebreyesus - diretor geral da Organização Mundial da Saúde}
{Conferência de Segurança de Munique (MSC) em 2020}

\chaves{Processamento de Linguagem Natural, \textit{Fake News}, Verificação Semi-Automática de Fatos, \textit{Corpora} em português.}

\begin{resumo}
A disseminação acelerada de desinformação excede a capacidade da verificação manual de fatos, evidenciando a necessidade de sistemas de Verificação Semi-Automática de Fatos (AFC). \novo{No contexto da língua portuguesa, constata-se uma carência de conjuntos de dados (\textit{corpora}) publicamente disponíveis que integrem evidências externas}, um componente essencial para o desenvolvimento de sistemas robustos de AFC, uma vez que muitos recursos existentes focam apenas na classificação baseada em características intrínsecas do texto.

Esta dissertação aborda essa lacuna \novo{desenvolvendo, aplicando e analisando uma metodologia para enriquecer \textit{corpora} de notícias em português (Fake.Br, COVID19.BR, MuMiN-PT) com evidências externas}. A abordagem \novo{simula o processo de verificação de um usuário}, empregando Modelos de Linguagem Grandes (LLMs, especificamente Gemini 1.5 Flash) \novo{para extrair a alegação principal dos textos} e APIs de mecanismos de busca (API de busca do Google, API de busca de alegações do Google FactCheck) para recuperar documentos externos relevantes (evidências). \novo{Adicionalmente, um processo de validação e pré-processamento de dados, incluindo detecção de quase duplicatas, é introduzido para aprimorar a qualidade dos \textit{corpora} base.} 

\novo{Os principais resultados demonstram a viabilidade da metodologia, fornecendo \textit{corpora} enriquecidos e análises que confirmam a utilidade da extração de alegações, a influência das características dos dados originais no processo, e o impacto positivo do enriquecimento no desempenho de modelos de classificação (Bertimbau e Gemini 1.5 Flash), especialmente com ajuste fino. Este trabalho contribui com recursos valiosos e \textit{insights} para o avanço da AFC em português.}
\end{resumo}

\keys{Natural Language Processing, Fake news, Semi-Automated Fact-checking, Portuguese \textit{Corpora}.}

\begin{abstract}{Semi-automated Fact-checking in Portuguese: \textit{Corpora} Enrichment using Retrieval with Claim extraction}
The accelerated dissemination of disinformation often outpaces the capacity for manual fact-checking, highlighting the urgent need for Semi-Automated Fact-Checking (SAFC) systems. \novo{Within the Portuguese language context, there is a noted scarcity of publicly available datasets (\textit{corpora}) that integrate external evidence}, an essential component for developing robust AFC systems, as many existing resources focus solely on classification based on intrinsic text features.

This dissertation addresses this gap by \novo{developing, applying, and analyzing a methodology to enrich Portuguese news \textit{corpora} (Fake.Br, COVID19.BR, MuMiN-PT) with external evidence}. The approach \novo{simulates a user's verification process}, employing Large Language Models (LLMs, specifically Gemini 1.5 Flash) \novo{to extract the main claim from texts} and search engine APIs (Google Search API, Google FactCheck Claims Search API) to retrieve relevant external documents (evidence). \novo{Additionally, a data validation and preprocessing framework, including near-duplicate detection, is introduced to enhance the quality of the base \textit{corpora}.} 

\novo{The main results demonstrate the methodology's viability, providing enriched \textit{corpora} and analyses that confirm the utility of claim extraction, the influence of original data characteristics on the process, and the positive impact of enrichment on the performance of classification models (Bertimbau and Gemini 1.5 Flash), especially with fine-tuning. This work contributes valuable resources and insights for advancing SAFC in Portuguese.}
\end{abstract}

\tabelas[figtab]

\chapter{Introdução}
\label{c:intro}

A era digital, especialmente a partir de 2015, redefiniu o consumo de informação, caracterizada pelo uso massivo das mídias sociais. Nesse cenário, observa-se que conteúdos circulam frequentemente sem os critérios de rigor e qualidade associados ao jornalismo tradicional. Em países como Reino Unido e Estados Unidos, constata-se uma mudança geracional, na qual os jovens adotam crescentemente as redes sociais digitais como principal fonte de notícias, em detrimento de meios consolidados como a televisão \cite{aimeur2023review}. Contudo, essa mesma dinâmica digital facilitou a disseminação de desinformação e \textit{fake news}, que se tornaram ferramentas potentes de manipulação, capazes de infligir danos significativos a reputações \textit{corpora}tivas, governamentais e a grupos sociais \cite{meel2020survey,schlicht2023health_systematic}.

Diante desse desafio, agências de checagem de fatos (\textit{fact-checking}), como a Agência Lupa e o Boatos.org no Brasil, desempenham um papel crucial ao investigar manualmente a veracidade das informações \cite{faustini2019bracisfake, couto2021central_de_fatos}. No entanto, a velocidade viral com que a desinformação se propaga excede em muito a capacidade de verificação humana, um problema exacerbado durante a pandemia de COVID-19, período em que o confinamento intensificou o uso da internet e a circulação de conteúdos duvidosos \cite{varma2021health}.

Essa limitação intrínseca da verificação manual impulsionou a pesquisa e o desenvolvimento de ferramentas de verificação automática ou semi-automática de fatos, área conhecida como \textit{Automated Fact-Checking} (AFC). Essas abordagens buscam analisar a veracidade de alegações comparando-as com fontes externas de conhecimento, integrando técnicas de Recuperação de Informação (IR) e Processamento de Linguagem Natural (PLN) \cite{guo2022survey_afc}. 

\novo{Reconhece-se, no entanto, que a automação completa ainda enfrenta desafios significativos, especialmente na interpretação de nuances contextuais e na avaliação da credibilidade das fontes. Por essa razão, muitos sistemas operam de forma semi-automática, onde a tecnologia auxilia o especialista humano, que realiza a validação final \cite{martin2022factercheck, villarrodríguez2024distrack}. Adotou-se, nesta dissertação, o termo Verificação Semi-Automática de Fatos para refletir essa interação colaborativa.}

No contexto da língua portuguesa, apesar da existência de pesquisas e recursos para a detecção de \textit{fake news}, uma lacuna significativa persiste. Uma análise de 18 \textit{corpora} publicamente disponíveis (detalhados na Seção \ref{sec:related_pt}) revela que a maioria se concentra na classificação da notícia com base em suas características intrínsecas (estilo de escrita, parcialidade), como exemplificado pelo \textit{corpus} Fake.br \cite{monteiro2018fakebr} e abordagens iniciais em inglês \cite{guo2022survey_afc}.  São escassos os conjuntos de dados em português que fornecem as evidências externas associadas às alegações, um componente crucial para treinar e avaliar sistemas semi-automáticos de AFC robustos que se baseiam na verificação factual contra fontes externas, em vez de apenas classificar o texto isoladamente.

Esta dissertação visa abordar diretamente essa lacuna. O objetivo central deste trabalho é desenvolver uma método para enriquecer conjuntos de dados de notícias em português já existentes, agregando a eles evidências contextuais relevantes recuperadas de fontes externas. Para este fim, foram selecionados três \textit{corpora} proeminentes com distintas características de fonte, método de coleta e temporalidade: Fake.Br \cite{monteiro2018fakebr}, notícias de páginas \correcaoPrevia{da} web de domínio geral; COVID19.BR \cite{martins2021covid19br}, mensagens do WhatsApp sobre saúde e MuMiN-PT \cite{nielsen2024mumin}, \textit{tweets} de domínio geral.

A abordagem proposta simula o processo cognitivo de um usuário que busca informações adicionais para verificar a veracidade de uma notícia. Para tanto, foram empregados Modelos de Linguagem Grandes (LLMs), especificamente o Gemini 1.5 Flash, para extrair a alegação principal contida no texto original, especialmente quando a busca direta não encontrava correspondência forte. Essa alegação serve como consulta otimizada para mecanismos de busca, como a API de busca do Google (CSE) \novo{e a API de busca de alegações do Google FactCheck}, na recuperação de documentos externos relevantes (evidências), que são então associados ao item original do \textit{corpus}.

O resultado principal desta pesquisa é a criação de versões enriquecidas destes \textit{corpora}, acompanhadas de uma análise detalhada do processo de coleta, das características dos dados obtidos (como a prevalência de quase duplicatas e a natureza das evidências recuperadas) e do impacto das características originais dos conjuntos de dados nas análises subsequentes. \correcao{Esta dissertação, portanto, não se limita a apresentar os conjuntos de dados enriquecidos como produto final, mas detalha e analisa criticamente todo o fluxo de trabalho operacional — desde a validação dos dados brutos, passando pela extração de alegações, até a recuperação e avaliação das evidências —, oferecendo insights sobre os desafios e as decisões metodológicas em cada etapa.}

\novo{Uma análise qualitativa dos dados enriquecidos identificou padrões recorrentes na corroboração ou refutação de alegações, incluindo o reconhecimento de exemplos dos \textit{corpora} em publicações acadêmicas.} \correcaoPrevia{Além isso, o processo de validação semi-automático dos dados, que incluiu a detecção de quase duplicatas e checagens de consistência de rótulos, mostrou-se fundamental para melhorar a qualidade dos dados base antes do enriquecimento.}

\novo{De forma a avaliar experimentalmente o impacto do processo, comparou-se o desempenho do ajuste fino do Bertimbau base e de \textit{few-shot prompting} do Gemini 1.5 Flash em diferentes configurações de dados (originais, validados, e validados e enriquecidos). Os resultados indicaram que, embora os dados apenas validados pudessem apresentar desempenho inferior aos originais devido ao aumento da complexidade da tarefa, o enriquecimento com conteúdo externo geralmente melhorou o desempenho sobre os dados apenas validados, especialmente para o Bertimbau e Gemini no COVID19.BR. Consistentemente, o ajuste fino do Bertimbau superou o \textit{few-shot learning} com Gemini. Estes achados sugerem que o enriquecimento adiciona contexto valioso, mas sua eficácia depende da qualidade da busca, da extração de alegações, da cobertura das APIs e da temporalidade da informação, apontando para o potencial de abordagens híbridas.}

Este trabalho foi motivado pela participação da autora em um projeto de pesquisa sobre \textit{Fake News} em colaboração com a Agência Nacional de Telecomunicações (ANATEL) e Fundação de Amparo à Pesquisa do Estado de Goiás (FAPEG), \correcaoPrevia{em que se publicou uma revisão terciária rápida sobre a área de detecção de fake news \cite{gomes2023erigo}}. 

A dissertação também conta com o apoio do Centro de Excelência em Inteligência Artificial (CEIA) e do Centro de Competência EMBRAPII em Tecnologias Imersivas (Advanced Knowledge Center for Immersive Technologies - AKCIT) do Instituto de Informática da Universidade Federal de Goiás (INF-UFG), \correcaoPrevia{no qual a autora participou da competição CLEF CheckThat! 2025 \cite{CheckThat:ECIR2025, clef-checkthat:2025-lncs}. O sistema proposto, utilizando o ajuste fino do modelo Mono-PTT5 \cite{10.1007/978-3-031-79032-4_23}, provou-se altamente competitivo ao alcançar a terceira posição para o português (METEOR 0.5290) \cite{clef-checkthat:2025:task2}.}

A seguir, na Seção \ref{sec:hypo}, são apresentadas as hipóteses que nortearam esta investigação. Os objetivos específicos decorrentes dessas hipóteses são detalhados na Seção \ref{sec:objectives}, e as principais contribuições do trabalho estão listadas na Seção \ref{sec:contribution}. A estrutura restante desta dissertação é organizada da seguinte forma:

\begin{description}
    \item[Capítulo \ref{c:background}] Apresenta a fundamentação teórica sobre \textit{fake news}, Processamento de Linguagem Natural (com foco em Modelos de Linguagem Grandes - LLMs) e Recuperação de Informação (IR).
    \item[Capítulo \ref{c:related}] Discute os trabalhos relacionados, abordando o uso de LLMs na verificação de fatos e pesquisas em PLN para detecção de \textit{fake news} em português.
    \item[Capítulo \ref{c:methods}] Descreve detalhadamente os métodos propostos para \novo{a validação semi-automático,} enriquecimento \novo{e avaliação experimental} dos \textit{corpora}.
    \item[Capítulo \ref{c:develop}] Apresenta o desenvolvimento prático do enriquecimento dos \textit{corpora}, a análise exploratória dos dados originais, a análise quantitativa e qualitativa dos dados enriquecidos, \novo{e os resultados da avaliação experimental}.
    \item[Capítulo \ref{c:conclusion}] Sumariza as conclusões do trabalho e aponta direções para pesquisas futuras.
\end{description}

\section{Hipóteses}
\label{sec:hypo}
Com o intuito de analisar o cenário em língua portuguesa de verificação de \textit{fake news} com conteúdo externo, foram elaboradas as seguintes hipóteses:

\begin{enumerate}[start=1,label={\bfseries H\arabic*:}]
 \item \textbf{Há uma escassez de \textit{Corpora} em português acessível para detecção de \textit{fake news} usando conteúdo externo, técnica conhecida como verificação de fatos}. Dentre 18 conjuntos de dados identificados, poucos fornecem evidências associadas, e os que o fazem podem apresentar limitações de acesso ou escopo.

\item\textbf{A partir dos conjuntos de dados em português de detecção de \textit{fake news} que possuem somente alegações, é possível enriquecê-los com evidências}. Utilizando mecanismos de busca e, quando necessário, extração de alegação via LLM, pode-se obter material contextual para cada notícia, embora se reconheçam limitações temporais (relevância da evidência ao longo do tempo, mudança de veracidade de fatos) e de verificabilidade (alegações subjetivas, "meias-verdades") inerentes ao processo.

\item\textbf{A extração de alegação via LLM auxilia no processo de verificação de \textit{fake news} em português, otimizando a busca por evidências}. \novo{Acredita-se que esta abordagem melhora a precisão da busca por evidências, especialmente para textos menos diretos, embora sua necessidade varie conforme a clareza do texto original e a natureza do \textit{corpus}.}

\item\textbf{A natureza da coleta dos dados (\textit{top-down} vs. \textit{bottom-up}) interfere no processo de enriquecimento e nas características dos \textit{corpora}}. \novo{Conjuntos de dados \textit{bottom-up} (ex. MuMiN-PT), originados de verificações de agências, tendem a ter mais exemplos falsos e maior correspondência com resultados de busca de verificações, demandando menos extração de alegação.}

\item{\textbf{O veículo de publicação impacta nas características distintivas entre \textit{fake news} e textos verdadeiros.}} \novo{Conforme observado na análise exploratória, o meio original de publicação da notícia (e.g., Twitter, sites de notícias, WhatsApp) influencia características textuais como tamanho, \novo{homogeneidade, presença de URLs e a natureza das quase duplicatas}, que podem diferenciar notícias verdadeiras de falsas de maneiras específicas para cada plataforma.}

\item\novo{\textbf{O processo de validação e enriquecimento dos dados \correcaoPrevia{auxiliam} o desempenho de modelos de detecção de \textit{fake news}.}} \novo{A validação, ao remover certos sinais (e.g., URLs), pode tornar a tarefa mais desafiadora. O enriquecimento com conteúdo externo, por outro lado, visa fornecer contexto adicional que pode melhorar a capacidade de generalização e o desempenho dos modelos, embora a qualidade e relevância da informação externa sejam cruciais}.

\end{enumerate}

\section{Objetivos}
\label{sec:objectives}
Com base no problema identificado e nas hipóteses formuladas, o objetivo geral desta dissertação é \textbf{\correcao{desenvolver} uma metodologia para enriquecer conjuntos de dados de notícias em português com evidências externas recuperadas automaticamente}. Os objetivos específicos são:

\begin{enumerate}
    \item Realizar um levantamento abrangente e uma análise comparativa aprofundada dos \textit{corpora} existentes para detecção de \textit{fake news} em português, focando em suas características relevantes para a verificação baseada em evidências (ex., fonte, método de coleta, disponibilidade de evidências, características textuais\novo{, prevalência de quase duplicatas}).
    \item \novo{Projetar e implementar um fluxo de trabalho para o enriquecimento de \textit{corpora}, incluindo uma etapa de validação semi-automática dos dados base}, investigando e selecionando técnicas de engenharia de \textit{prompt} para extração de alegações via LLMs e utilizando APIs de mecanismos de busca para a recuperação de evidências.

    \item Aplicar a metodologia \novo{de validação} e enriquecimento em \textit{corpora} selecionados em português (Fake.Br, COVID19.BR, MuMiN-PT), gerando conjuntos de dados validados e enriquecidos com referências externas.

    \item \correcao{Avaliar} os resultados do processo de enriquecimento, considerando aspectos como \novo{a aplicabilidade da extração de alegação em diferentes \textit{corpora}}, a \novo{distribuição e natureza das fontes de evidência recuperadas (busca web geral vs. API de Fact Check, incluindo domínios governamentais, de mídia e de checagem)}, e as características dos dados gerados, \novo{relacionando-as às propriedades originais dos conjuntos de dados e identificando padrões qualitativos nas evidências.}

    \item \novo{Comparar experimentalmente a eficácia das etapas propostas de validação semi-automática dos dados e de enriquecimento com conteúdo externo no desempenho de modelos de PLN (Bertimbau e Gemini 1.5 Flash) na tarefa de classificação de veracidade.}
\end{enumerate}

\section{\novo{Contribuição}}
\label{sec:contribution}
\noindent As contribuições desta dissertação são:

\begin{enumerate}
    \item \novo{Um levantamento e \correcao{comparação} \textit{corpora} de \textit{fake news} em português (18 conjuntos de dados), destacando características frequentemente negligenciadas como métodos de coleta (\textit{top-down} vs. \textit{bottom-up}) \novo{e a prevalência e impacto de exemplos quase duplicados,} e a influência das fontes de informação primárias (notícias da internet, Twitter, WhatsApp).}

    \item \novo{Um processo de validação e pré-processamento de dados, incorporando a detecção de quase duplicatas (MinHash LSH) e checagens de consistência de rótulos, visando melhorar a confiabilidade dos \textit{corpora} base antes do enriquecimento, com a remoção de vieses como URLs explícitas.}
    
    \item \novo{O \correcao{desenvolvimento} de uma metodologia para enriquecer conjuntos de dados de notícias em português com informações contextuais externas, utilizando LLMs (Gemini 1.5 Flash) para extração de alegações e APIs de mecanismos de busca (API de Busca do Google e API de Busca de alegações do Google FactCheck) para recuperação de evidências.}
    
    \item \novo{Uma análise aprofundada dos \textit{corpora} enriquecidos sobre a eficácia da extração de alegações em diferentes contextos, a natureza e distribuição dos domínios das fontes de evidência recuperadas (incluindo fontes jornalísticas, governamentais, acadêmicas e de checagem), e como as características originais dos \textit{corpora} (fonte, temporalidade, método de coleta) influenciam o processo de enriquecimento e a natureza das evidências. Inclui-se uma análise qualitativa dos padrões de evidências e a identificação de publicações acadêmicas que referenciam exemplos dos \textit{corpora}.}

    \item \novo{Uma avaliação experimental do impacto das etapas de validação e enriquecimento no desempenho de modelos de detecção de \textit{fake news} (Bertimbau e Gemini 1.5 Flash), demonstrando o potencial do enriquecimento contextual, mas também suas nuances e dependências da qualidade e relevância da informação externa.}
\end{enumerate}

\chapter{Fundamentos} \label{c:background}
Este capítulo apresenta os conceitos essenciais para a compreensão do problema da \textit{fake news} e das abordagens computacionais para seu combate. Inicialmente, explora-se o fenômeno das \textit{fake news} \correcao{na Seção \ref{sec:fake}}, detalhando sua taxonomia e as técnicas empregadas para sua detecção, com ênfase na verificação de fatos (\textit{fact-checking}). Em seguida, abordam-se os fundamentos das áreas de conhecimento que viabilizam a automação ou semi-automação desse processo: o Processamento de Linguagem Natural (PLN) na \correcao{Seção \ref{sec:pln}}, e a Recuperação de Informação (RI) em \correcao{\ref{sec:ir}}.

No âmbito do PLN, discute-se a evolução dos modelos de linguagem, desde abordagens estatísticas até os Modelos de Linguagem Pequenos (SLMs), como o BERT e o Bertimbau, e os Modelos de Linguagem Grandes (LLMs), como o Gemini. Destaca-se, neste contexto, a técnica de engenharia de \textit{prompts}, \novo{fundamental para interagir com LLMs na tarefa de extração de alegações, como realizado neste trabalho}. Finalmente, explora-se a Recuperação de Informação, definindo seus conceitos gerais e sua interconexão com o PLN, a verificação de fatos (especificamente na busca por evidências) e os mecanismos de busca, como o buscador Google, \novo{cujas APIs são frequentemente empregadas em sistemas de verificação}.

\section{\textit{Fake news}}
\label{sec:fake}
\textit{Fake news} podem ser compreendidas como conteúdo fabricado que mimetiza o formato de notícias genuínas, com o intuito de enganar o leitor \cite{wu2022review}. É relevante notar que o termo "\textit{fake news}" carece de uma definição única e universalmente aceita, sendo sua interpretação sujeita a variações contextuais e disciplinares \cite{aimeur2023review}.

A Seção \ref{subsec:taxonomy} aprofunda a terminologia associada. Posteriormente, a Seção \ref{subsec:techs} delineia as principais abordagens tecnológicas para a detecção e análise de \textit{fake news}, com base em \cite{hangloo2022survey}.

\subsection{Taxonomia}
\label{subsec:taxonomy}
Sharma et al. (2019) definem \textit{fake news} como notícias ou mensagens publicadas e disseminadas pela mídia que contêm informações falsas, independentemente dos meios ou motivações por trás de sua propagação. Esta definição abrange diversos tipos de conteúdo problemático identificados na literatura \cite{sharma2019survey}.

Com base nessa perspectiva, é possível categorizar as \textit{fake news} em subtipos como: conteúdo fabricado (totalmente falso), conteúdo enganoso (uso de informação para distorcer um fato ou indivíduo), conteúdo impostor (fontes genuínas falsamente atribuídas), conteúdo manipulado (informação ou imagem genuína alterada para enganar), conexão falsa (títulos, imagens ou legendas não condizentes com o conteúdo) e contexto falso (conteúdo genuíno compartilhado com informação contextual incorreta) \cite{sharma2019survey}.

Adicionalmente, a informação falsa pode ser classificada pela intenção. A \textit{misinformation} (informação falsa, não intencional) refere-se à disseminação involuntária de falsidades, que pode advir de vieses cognitivos, falta de compreensão ou atenção. Em contraste, a \textit{disinformation} (desinformação, intencional) envolve a criação e propagação deliberada de informações falsas com o objetivo explícito de enganar \cite{aimeur2023review, sharma2019survey}.

\noindent Outras formas de conteúdo relacionadas incluem:
\begin{itemize}
    \item \textbf{Paródia e Sátira:} Frequentemente associadas ao humor, onde sátiras podem imitar o estilo da mídia tradicional para criticar ou expor algo, sem a intenção primária de enganar, embora possam ser mal interpretadas \cite{hangloo2022survey}.
    \item \textbf{\textit{Clickbaits}:} Títulos ou chamadas sensacionalistas projetados para atrair cliques e direcionar tráfego para um \textit{site}, muitas vezes com fins de monetização por publicidade, mesmo que o conteúdo não entregue o prometido \cite{hangloo2022survey}.
    \item \textbf{Propaganda:} Informação, muitas vezes enviesada ou enganosa, usada para promover uma causa ou ponto de vista político específico, visando influenciar a opinião pública \cite{aimeur2023review}.
    \item \textbf{Teorias da Conspiração:} Explicações para eventos que invocam tramas secretas por parte de grupos poderosos, geralmente carecendo de evidências robustas e resistindo à refutação \cite{aimeur2023review}.
\end{itemize}

\subsection{Técnicas de Detecção}
\label{subsec:techs}
\novo{Dada a escala e complexidade da disseminação de \textit{fake news}}, diversas abordagens tecnológicas têm sido desenvolvidas para sua detecção e análise. As principais se baseiam em aprendizado de máquina (ML), aprendizado profundo (DL), processamento de linguagem natural (PLN), verificação de fatos (\textit{fact-checking} ou FC, \novo{incluindo a Verificação Automática de Fatos - AFC e a Verificação Semi-Automática de Fatos - SAFC}), \textit{crowdsourcing} (CDS), \textit{blockchain} (BKC) e redes neurais de grafos (GNN) \cite{aimeur2023review}.

\textbf{Inteligência Artificial (IA):} É um termo guarda-chuva frequentemente empregado no contexto de ML ou DL \cite{garg2020ml_vs_dl, hangloo2022survey}.
\begin{itemize}
    \item \textbf{Aprendizado de Máquina (ML):} Refere-se a métodos clássicos onde características (\textit{features}) relevantes, como contagem de palavras, comprimento de sentenças ou frequência de certos termos, são frequentemente extraídas ou projetadas manualmente (\textit{feature engineering}) para treinar modelos preditivos.
    \item \textbf{Aprendizado Profundo (DL):} Utiliza redes neurais com múltiplas camadas para aprender representações hierárquicas dos dados. Geralmente, a extração de características é feita automaticamente pela própria rede durante o treinamento. Métodos de DL costumam exigir maior volume de dados e capacidade computacional, mas podem alcançar desempenho superior em tarefas complexas.
\end{itemize}

\textbf{Processamento de Linguagem Natural (PLN):} Engloba técnicas focadas na análise do conteúdo textual das notícias. Podem-se examinar aspectos:
\begin{itemize}
    \item \textbf{Lexicais e Sintáticos:} Análise de frequência de palavras, uso de classes gramaticais (POS \textit{tagging}), comprimento de sentenças, complexidade sintática, presença de erros ortográficos \cite{reis2019nlp}.
    \item \textbf{Semânticos:} Análise do significado do texto, incluindo a detecção de tópicos, análise de sentimentos, e a representação do texto através de \textit{embeddings} (vetores numéricos) que capturam relações semânticas.
    \item \textbf{Psicolinguísticos:} Extração de características relacionadas a processos psicológicos e sociais, utilizando ferramentas como o LIWC (Linguistic Inquiry and Word Count) \cite{tausczik2019liwc}.
\end{itemize}

Classificadores tradicionais de ML, como Regressão Logística, podem ser combinados com representações textuais como TF-IDF (Frequência do Termo–Inverso da Frequência do Documento) ou com \textit{embeddings} gerados por modelos de linguagem neural para a classificação de notícias.

\textbf{Verificação de Fatos (\textit{Fact-Checking}, FC):} Consiste em avaliar a veracidade de alegações factuais comparando-as com fontes de informação externas e confiáveis. Este processo pode ser:
\begin{itemize}
    \item \textbf{Manual:} Realizado por jornalistas especializados ou checadores de fatos, como os das agências listadas na Tabela \ref{tab:rotulos}. Também pode envolver \textbf{\textit{Crowdsourcing} (CDS)}, onde tarefas de verificação são distribuídas a um grande número de pessoas \cite{ali2022survey, kondamudi2021systematic, hangloo2022survey, aimeur2023review}.
    \item \novo{\textbf{Automático ou Semi-Automático (AFC / SAFC):}} Busca automatizar partes ou todo o processo de verificação usando técnicas computacionais, principalmente de PLN e RI. A abordagem semi-automática (SAFC), \textbf{adotada como foco conceitual nesta dissertação}, reconhece a importância da intervenção humana em etapas críticas, como na validação final ou no tratamento de casos complexos \cite{martin2022factercheck, ni2024afacta, villarrodríguez2024distrack}.
\end{itemize}

\textbf{Blockchain (BKC):} É uma tecnologia de registro distribuído e imutável. Dados são armazenados em blocos encadeados criptograficamente em múltiplos servidores, sem uma autoridade central. Uma vez adicionado, um registro não pode ser alterado ou excluído. Propriedades como imutabilidade, descentralização, resistência à adulteração e transparência (controlada) tornam o blockchain uma ferramenta potencial para rastrear a proveniência e a integridade de conteúdos digitais, ajudando a verificar sua autenticidade \cite{dhall2021blockchain, ahmad2022systematic}.

\textbf{Redes Neurais de Grafos (GNN):} São modelos de DL projetados para operar sobre dados estruturados em grafos. No contexto de \textit{fake news}, podem modelar as relações entre usuários, notícias e fontes, ou a propagação de informação em redes sociais, para identificar padrões associados à \textit{fake news} \cite{aimeur2023review}.

\begin{figure}[htb]
     \centering
        \includegraphics[width=\textwidth]{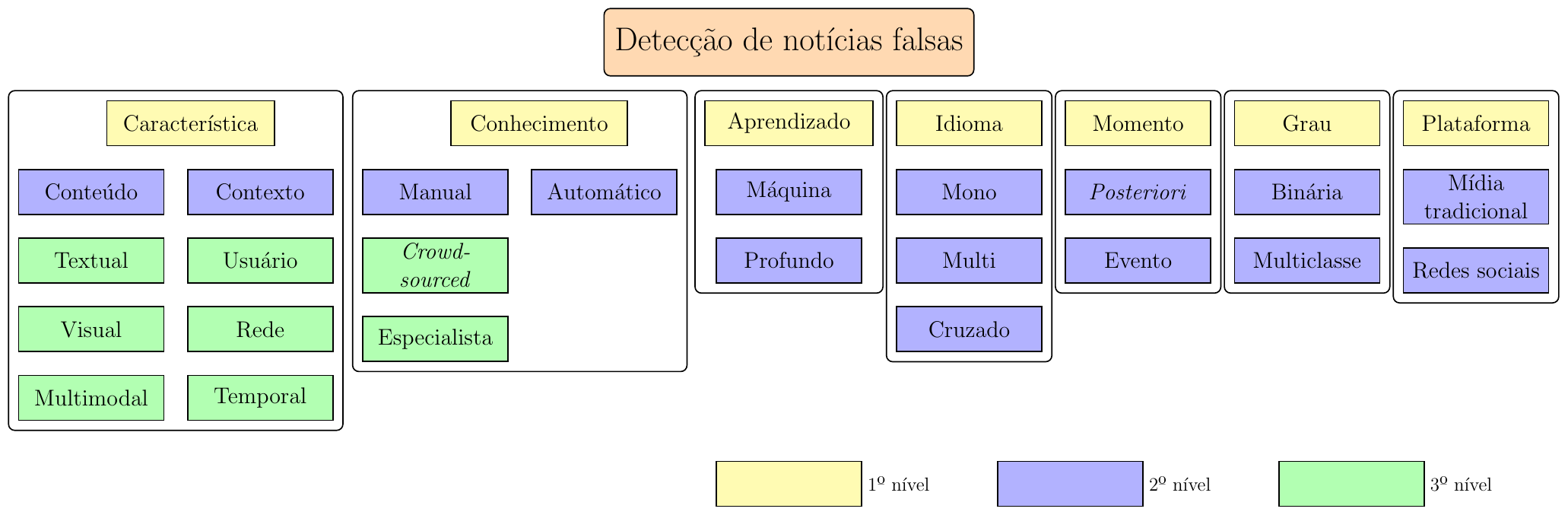}
     \caption{Taxonomia de técnicas de detecção de \textit{fake news}, adaptado de \cite{hangloo2022survey}.}
     \label{fig:taxonomia}
 \end{figure}

As técnicas de detecção podem ser agrupadas conforme seu foco principal \cite{ali2022survey, kondamudi2021systematic, hangloo2022survey}, como ilustrado na Figura \ref{fig:taxonomia}. A taxonomia divide as abordagens em quatro categorias principais: Baseadas em Características, Baseadas em Conhecimento e Baseadas em Aprendizado, complementadas por aspectos como idioma, granularidade da detecção, granularidade da verdade e plataforma.

\textbf{Abordagens Baseadas em Características:} Examinam propriedades intrínsecas ou extrínsecas da notícia para identificar padrões suspeitos.
\begin{itemize}
    \item \textbf{Baseadas em Conteúdo:} Focam nos elementos da própria notícia: texto, imagem, vídeo (ou combinações multimodais). Análise textual (PLN), análise de sentimento (viés negativo), detecção de manipulação em imagens ou vídeos (\textit{deepfakes}) são exemplos. Conteúdo textual pode ser extraído de áudio/vídeo via Reconhecimento Automático de Fala (ASR) ou de imagens via Reconhecimento Óptico de Caracteres (OCR).
    \item \textbf{Baseadas em Contexto Social:} Analisam o ambiente e a forma como a notícia se propaga. Inclui o estudo de padrões de disseminação em redes sociais, reações de usuários (comentários, compartilhamentos), características dos perfis que compartilham (detecção de bots), e a estrutura da rede de interações (usando GNNs ou análise de redes). \novo{Fatores temporais, como a recorrência de certos rumores, e a credibilidade atribuída à fonte da notícia também são considerados} \cite{aimeur2023review, ali2022survey, kondamudi2021systematic}.
\end{itemize}

\textbf{Abordagens Baseadas em Conhecimento:} Correspondem essencialmente às técnicas de Verificação de Fatos (FC), que confrontam as alegações da notícia com fontes externas de conhecimento (bases de dados, artigos de checagem, fontes confiáveis). \novo{Um desafio central nesta abordagem é a coleta e organização das evidências relevantes, especialmente para idiomas com menos recursos, como o português, onde há carência de bases de dados que integrem alegações e suas respectivas evidências externas.} Conforme mencionado, pode ser manual, automática (AFC) ou semi-automática (SAFC) \cite{ali2022survey, kondamudi2021systematic, hangloo2022survey}.

\textbf{Abordagens Baseadas em Aprendizado:} Referem-se ao uso de IA (ML e DL). Métodos de ML clássico são geralmente mais interpretáveis e requerem menos dados, mas dependem de engenharia de características manual. Métodos de DL podem aprender representações complexas automaticamente a partir de grandes volumes de dados, frequentemente alcançando maior acurácia, mas são computacionalmente mais custosos e podem ser menos transparentes ("caixas pretas") \cite{garg2020ml_vs_dl}.

Outros aspectos relevantes na detecção de \textit{fake news} incluem:
\begin{itemize}
    \item \textbf{Idioma:} A detecção pode ser monolíngue ou multilíngue. Técnicas de aprendizado translingual (\textit{cross-lingual learning}) buscam transferir conhecimento de um idioma rico em recursos (como o inglês) para outro com menos recursos (como o português) \cite{hangloo2022survey}.
    \item \textbf{Momento da Detecção:} A classificação pode ocorrer \textit{a posteriori} (após a notícia circular) ou, idealmente, em estágios iniciais de sua disseminação (\textit{early detection}). A maioria dos métodos atuais opera \textit{a posteriori} \cite{hangloo2022survey}.
    \item \textbf{Granularidade da Verdade:} A classificação pode ser binária (verdadeiro/falso) ou multiclasse, incorporando níveis intermediários como "enganoso", "fora de contexto", "impreciso", etc. \cite{hangloo2022survey}. O conjunto de dados LIAR, por exemplo, usa seis níveis. A Tabela \ref{tab:rotulos} mostra a variedade de rótulos usados por agências brasileiras, refletindo diferentes graus de veracidade.
    \item \textbf{Plataforma:} A análise pode focar em uma plataforma específica (Twitter, Facebook, WhatsApp) ou em fontes de notícias tradicionais (web). As características da \textit{fake news} e sua propagação podem variar significativamente entre plataformas \cite{statista2024social_media}.
\end{itemize}

A Tabela \ref{tab:rotulos} indica os rótulos usados por agências brasileiras de checagem de fatos do conjunto de dados Central de Fatos \cite{couto2021central_de_fatos}. \correcaoPrevia{O mais comum é “falso/fake/boato”.}

Somente a agência Boatos.org possui uma categoria que seria falsa: \enquote{boato}. As outras cinco agências categorizam meias-verdades, em que quatro possuem mais de uma classe associada a meia-verdade. O Projeto Comprova, por exemplo, usa duas categorias para indicar meias-verdades, \enquote{enganoso} e \enquote{contexto errado}. O Estadão Verifica, de forma análoga, utiliza as categorias correspondentes \enquote{enganoso} e \enquote{fora de contexto}. A agência Lupa e Aos Fatos usam maior granularidade de \enquote{meias verdades}, como exagerado, \enquote{verdadeiro mas}, \enquote{ainda é cedo para dizer}, subestimado e exagerado, no caso da agência Lupa, e distorcido, exagerado e impreciso, no caso do Aos Fatos.

\begin{table}
\begin{tabularx}{\textwidth}{@{}lX}
\toprule
Agência & Rótulos \\ \midrule
\small
Projeto Comprova & enganoso (183), \textbf{falso (159)}, comprovado (9), evidência comprovada (6), contexto errado (4) \\
Estadão Verifica & \textbf{falso (299)}, enganoso (160), fora de contexto (134) \\
Aos Fatos & \textbf{falso (2522)}, verdadeiro (637), impreciso (394), exagerado (239), distorcido (125), insustentável (120), contraditório (65) \\
Boatos.org & \textbf{boato (5523)} \\
G1: Fato ou Fake & \textbf{fake (1098)}, fato (394), não é bem assim (346) \\
Agência Lupa & \textbf{falso (3209)}, verdadeiro (1469), exagerado (866), verdadeiro mas (723), de olho (222), contraditório (189), subestimado (108), ainda é cedo para dizer (108), insustentável (97) \\ \bottomrule
\end{tabularx}
    \caption{Agências brasileiras de verificação de fatos e os rótulos associados no conjunto de dados Central de Fatos \cite{couto2021central_de_fatos}  \correcaoPrevia{(rótulo mais comum destacado em
negrito)}.}
    \label{tab:rotulos}
\end{table}

\subsubsection{Verificação de Fatos (\textit{Fact-Checking})}
\label{subsubsec:fc}
Como introduzido, a Verificação de Fatos (FC) é uma técnica baseada em conhecimento que avalia a veracidade de uma alegação textual comparando-a com evidências externas provenientes de fontes confiáveis \cite{guo2022survey_afc}.

\begin{figure}[ht]
    \centering
    \includegraphics[width=\textwidth]{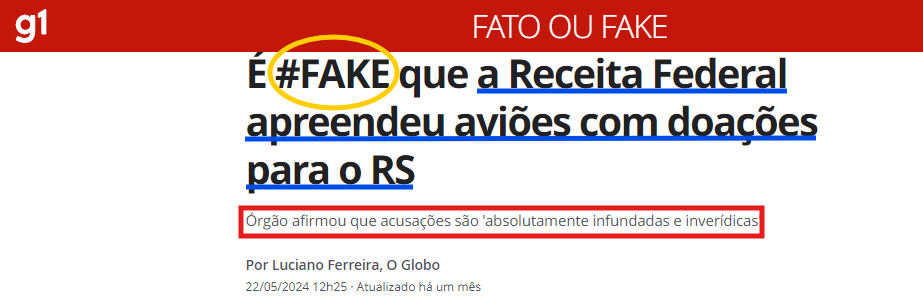}
    \caption[]{Exemplo de verificação manual de fatos realizada pela agência G1: Fato ou Fake, destacando a alegação (azul), o rótulo de veracidade (amarelo) e a evidência utilizada (vermelho) \footnotemark.}
    \label{fig:fact-checking}
\end{figure}

Historicamente realizada manualmente por jornalistas e organizações especializadas como mostrado na Figura \ref{fig:fact-checking}, a escala da desinformação online impulsionou a pesquisa em Verificação Automática de Fatos (AFC). \novo{Entretanto, dada a complexidade da linguagem natural, nuances contextuais e a necessidade de julgamento crítico, muitos sistemas atuais operam de forma semi-automática (SAFC), onde ferramentas computacionais auxiliam o verificador humano, que retém um papel fundamental no processo} \cite{martin2022factercheck, villarrodríguez2024distrack}. O \textit{crowdsourcing} também pode ser empregado para escalar partes do processo manual ou para anotar dados para sistemas automáticos \cite{kondamudi2021systematic}.

\novo{Será utilizado, nesta dissertação, o conceito de verificação semi-automática. Isso porque esses processos não são por sua totalidade automatizados, necessitando do papel fundamental de especialistas da área} \cite{martin2022factercheck, villarrodríguez2024distrack}. Quando realizada de forma manual, a técnica pode ser realizada por \textit{crowd-sourcing} ou por especialistas da área. \textit{Crowd-sourcing} pode ser realizada com o apoio de plataformas como o Amazon Mechanical Turk \footnote{\url{https://www.mturk.com/}}, em que se contratam pessoas de forma remota para a anotação dos dados. 

\footnotetext{\tiny \url{https://g1.globo.com/fato-ou-fake/noticia/2024/05/22/e-fake-que-a-receita-federal-apreendeu-avioes-com-doacoes-para-o-rs.ghtml}}

O processo de AFC/SAFC é tipicamente modelado como um \textit{pipeline} composto pelas seguintes etapas \cite{guo2022survey_afc}:

\begin{enumerate}
    \item \textbf{Identificação/Extração de Alegação (\textit{Claim Detection/Extraction}):} Identificar sentenças ou trechos dentro de um texto (e.g., artigo de notícia, postagem em rede social) que contenham alegações factuais verificáveis, distinguindo-as de opiniões ou conteúdo não factual. \novo{Esta etapa é crucial e, como explorado nesta dissertação, pode ser abordada com o uso de LLMs}.
    
    \item \textbf{Recuperação de Evidências (\textit{Evidence Retrieval}):} Dada uma alegação identificada, buscar em um vasto repositório de informações (como a web, bases de conhecimento ou arquivos de notícias) por documentos ou trechos de texto que possam servir como evidência para confirmar ou refutar a alegação. \novo{Esta etapa depende fundamentalmente de técnicas de Recuperação de Informação (RI), muitas vezes utilizando APIs de mecanismos de busca}.

    \item \textbf{Predição de Veredito (\textit{Verdict Prediction}):} Com base na alegação e nas evidências recuperadas, determinar a veracidade da alegação. Isso envolve analisar a relação entre a alegação e cada evidência (suporta, refuta, não relacionado) e, em seguida, agregar essas análises para chegar a um veredito final (e.g., verdadeiro, falso, enganoso). \novo{Modelos de PLN são usados para avaliar a relação semântica entre alegação e evidência}.

    \item \textbf{Geração de Justificativa (\textit{Justification Generation}):} Explicar o porquê de um determinado veredito ter sido atribuído, idealmente selecionando os trechos de evidência mais relevantes que o suportam. Esta etapa visa aumentar a transparência e a interpretabilidade do processo de verificação.
\end{enumerate}

\novo{A construção de \textit{corpora} anotados com alegações, evidências e vereditos é fundamental para treinar e avaliar modelos para cada uma dessas etapas, especialmente para a predição de veredito e a análise da relação alegação-evidência}. \correcao{Com isso, \textit{pipeline} canônico serve como a inspiração fundamental para a metodologia de enriquecimento proposta neste trabalho, conforme detalhado no Capítulo \ref{c:methods}. As etapas de Identificação/Extração de Alegação e Recuperação de Evidências são o foco central do nosso fluxo, apresentado na Figura \ref{fig:process}.}

\section{Processamento de Linguagem Natural (PLN)}
\label{sec:pln}
O Processamento de Linguagem Natural (PLN) é um campo interdisciplinar, envolvendo Ciência da Computação, Inteligência Artificial e Linguística, que se dedica a capacitar computadores a processar, analisar, compreender e gerar linguagem humana (texto e fala) de forma significativa \cite{bpln2024pln}. O PLN é fundamental para diversas aplicações, incluindo tradução automática, análise de sentimentos, sistemas de diálogo, sumarização de texto e, para este trabalho, a análise e verificação de informações. 

Sistemas de PLN são frequentemente desenvolvidos e avaliados com base em \textbf{tarefas} específicas, como classificação de texto, reconhecimento de entidades nomeadas, resposta a perguntas (\textit{question answering}) ou geração de texto. \novo{As tarefas centrais abordadas nesta dissertação, no contexto da verificação de fatos, são a extração de alegações e a futura análise da relação alegação-evidência}.

O desenvolvimento e a avaliação de modelos de PLN dependem de conjuntos de dados textuais, conhecidos como \textbf{\textit{corpus}} (singular) ou \textbf{\textit{corpora}} (plural). \novo{A qualidade e as características dos \textit{corpora} (e.g., tamanho, diversidade, tipo de anotação) influenciam diretamente o desempenho dos modelos treinados sobre eles}. Os \textit{corpora} utilizados e enriquecidos neste trabalho são detalhados posteriormente (ver Seção \ref{sec:eda}).

\subsection{Modelo de Linguagem (LM)}
\label{subsec:LM}

Um Modelo de Linguagem (LM) é um modelo computacional probabilístico ou neural treinado para compreender e/ou gerar linguagem natural. Fundamentalmente, um LM atribui probabilidades a sequências de palavras ou aprende representações vetoriais (\textit{embeddings}) de palavras, sentenças ou documentos, capturando propriedades sintáticas e semânticas da língua \cite{bpln2024lm}. Dada uma entrada textual $x$, o LM a transforma em uma representação numérica $y$ (um vetor ou conjunto de vetores) que pode ser utilizada por algoritmos de aprendizado de máquina para realizar tarefas de PLN.

\begin{figure}[htb]
    \centering
    \includegraphics[width=\textwidth]{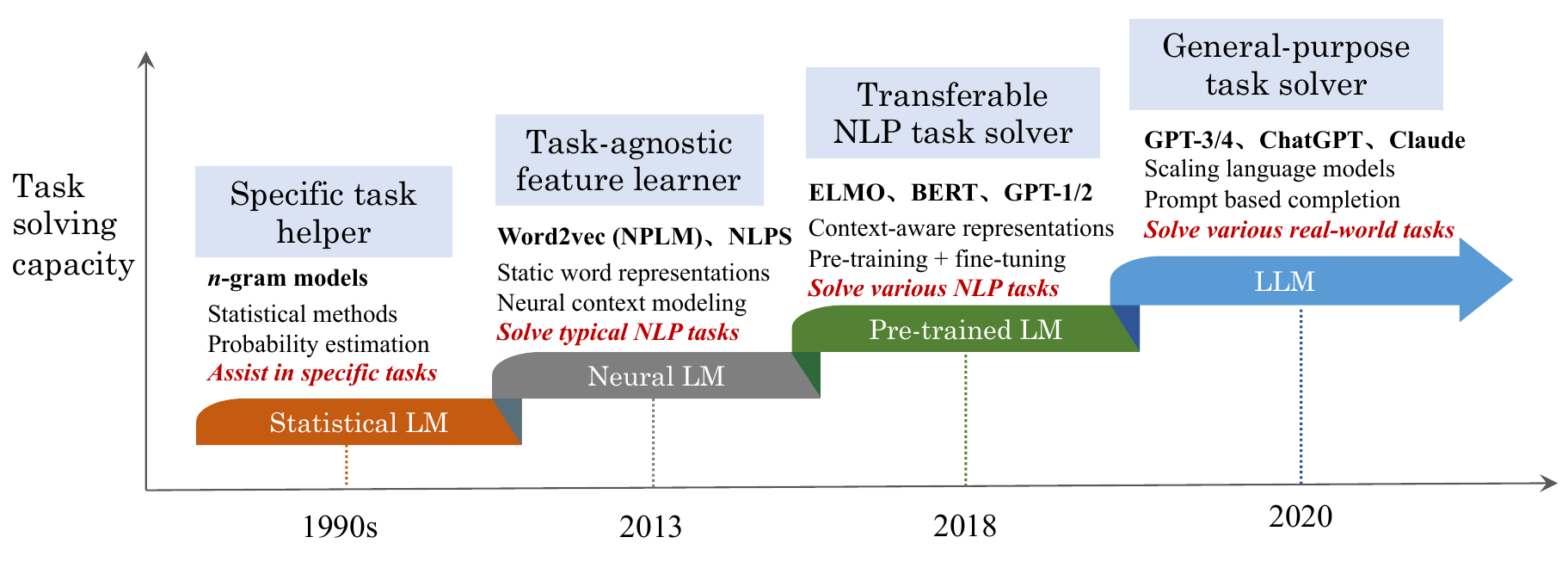}
    \caption{Linha do tempo ilustrando a evolução dos modelos de linguagem e sua crescente capacidade de resolver tarefas complexas \cite{zhao2023LLMSurvey}.}
    \label{fig:lms}
\end{figure}

A Figura \ref{fig:lms} ilustra a evolução dos modelos de linguagem. Em cada geração de arquitetura, aumentou-se a habilidade dos sistemas de realizarem tarefas complexas, geralmente à custa de maior necessidade de dados e recursos computacionais. As principais famílias de LMs incluem:

\textbf{LMs Estatísticos:} Modelos baseados em contagem de sequências de palavras ($n$-gramas) ou frequências de termos (e.g., TF-IDF, BM25). São computacionalmente eficientes e não requerem (ou requerem pouco) treinamento supervisionado, mas têm dificuldade em capturar o significado semântico profundo ou lidar com a ordem das palavras e dependências de longo alcance. Motores de busca tradicionais, como o Lucene\footnote{\url{https://lucene.apache.org/}}, frequentemente utilizam LMs estatísticos em seus índices \cite{bpln2024ir}.

\begin{figure}
    \centering
    \includegraphics[width=0.5\textwidth]{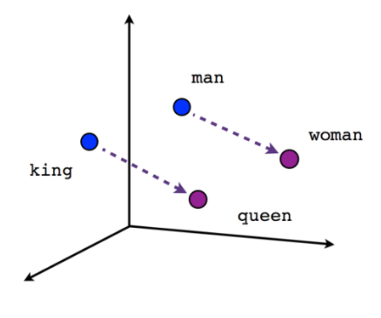}
    \caption[]{Analogia na representação de LMs estáticos \footnote{\url{https://ai.engin.umich.edu/2018/07/23/word-embeddings-and-how-they-vary/}.}}
    \label{fig:analogy}
\end{figure}

\textbf{LMs (Neurais) Estáticos:} Primeiros modelos de linguagem baseados em redes neurais que aprenderam representações vetoriais densas (\textit{embeddings}) para palavras, como o Word2vec \cite{mikolov2013word2vec}, GloVe \cite{pennington2014glove}, FastText \cite{bojanowski2017fasttext}). Esses modelos capturam relações semânticas e sintáticas (como a analogia "rei - homem + mulher = rainha", ilustrada na Figura \ref{fig:analogy}), mas atribuem um único vetor a cada palavra, independentemente do contexto em que ela aparece, limitando sua capacidade de lidar com polissemia (múltiplos significados) e ambiguidades. Em português, o NILC disponibilizou \textit{embeddings} estáticos treinados em grandes \textit{corpora} \cite{hartmann2017embeddings}.

\footnotetext{\url{https://ai.engin.umich.edu/2018/07/23/word-embeddings-and-how-they-vary/}}

\textbf{LMs (Pré-treinados) Contextuais:} Modelos baseados na arquitetura Transformer \cite{vawasni2017attention}, como BERT \cite{devlin2019bert} e RoBERTa \cite{liu2019roberta}, constituem uma classe importante de modelos de linguagem. Estes são pré-treinados em vastas quantidades de texto não rotulado da ordem de terabytes utilizando tarefas auto-supervisionadas, sendo a mais comum a predição de palavras mascaradas (\textit{Masked Language Modeling} - MLM).

O mecanismo de auto-atenção, intrínseco aos Transformers, permite que a representação vetorial (\textit{embedding}) de cada \textit{token} (palavra ou sub-palavra) seja computada dinamicamente com base em seu contexto textual. Isso possibilita a geração de \textit{embeddings} contextuais que capturam nuances semânticas, distinguindo, por exemplo, entre diferentes acepções de uma palavra polissêmica (como "banco" enquanto assento ou instituição financeira). \novo{Tipicamente compostos por centenas de milhões de parâmetros, esses modelos geralmente necessitam de uma etapa subsequente de ajuste fino (\textit{fine-tuning}) sobre um conjunto de dados rotulados específico da tarefa alvo (frequentemente na ordem de milhares de exemplos) para otimizar seu desempenho em aplicações específicas \cite{devlin2019bert}}.

\novo{No domínio da língua portuguesa, destaca-se o BERTimbau \cite{souza2020bertimbau}, uma adaptação do BERT otimizada para o português brasileiro. Desenvolvido a partir do modelo BERT multilíngue (\textit{base}), o BERTimbau foi submetido a pré-treinamento adicional utilizando o \textit{corpus} BrWAC \cite{wagner2018brwac}, composto por aproximadamente 3,5 milhões de páginas web em português. Na presente pesquisa, a versão \textit{base} do BERTimbau é empregada como um modelo representativo da abordagem de \textit{fine-tuning},} \correcao{conforme detalhado nas Seções \ref{sec:estrategia_avaliacao} e \ref{sec:ambiente_avaliacao}}.

Na literatura mais recente, modelos com a escala de parâmetros do BERT e seus derivados são frequentemente classificados como Modelos de Linguagem Pequenos (\textit{Small Language Models} - SLMs), em contraste com os Modelos de Linguagem Grandes (\textit{Large Language Models} - LLMs) \cite{zhao2023LLMSurvey}. Apesar da denominação "pequenos", os SLMs continuam a representar o estado da arte em diversas tarefas de Processamento de Linguagem Natural, especialmente na classificação de textos, quando se dispõe de um volume suficiente de dados rotulados para o ajuste fino \cite{qiu2024chatgpt_bert, yang2024llm_survey, vannguyen2024slm}.

\textbf{Modelos de Linguagem Grandes (LLMs):} Os Modelos de Linguagem Grandes (\textit{Large Language Models} - LLMs) constituem uma evolução dos LMs contextuais, distinguindo-se principalmente pela sua escala massiva, com um número de parâmetros na ordem de bilhões ou mesmo trilhões, e por serem treinados em volumes de dados textuais ainda mais extensos \cite{zhao2023LLMSurvey}. Exemplos proeminentes incluem a família de modelos GPT \cite{brown2020gpt3}. 

Uma característica marcante dos LLMs são suas habilidades emergentes, notadamente a capacidade de realizar tarefas novas com pouquíssima ou nenhuma demonstração prévia (\textit{zero-shot} ou \textit{few-shot learning}), respondendo diretamente a instruções fornecidas em linguagem natural (\textit{prompts}). Diferentemente dos SLMs, os LLMs não requerem, necessariamente, um ajuste fino específico para cada nova tarefa, o que os torna modelos mais generalistas e flexíveis. \novo{Nesta dissertação, o LLM Gemini 1.5 Flash \cite{geminiteam2024gemini, reid2024gemini} é empregado especificamente para duas finalidades: a extração de alegações factuais dos \textit{corpora} e a avaliação do impacto do enriquecimento de dados na classificação de notícias, utilizando a abordagem de \textit{few-shot learning},}\correcao{ conforme detalhado nas Seções \ref{sec:estrategia_avaliacao} e \ref{sec:ambiente_avaliacao}}.

\section{Modelos de linguagens grandes (LLM)}
\begin{figure}[ht]
    \centering
    \includegraphics{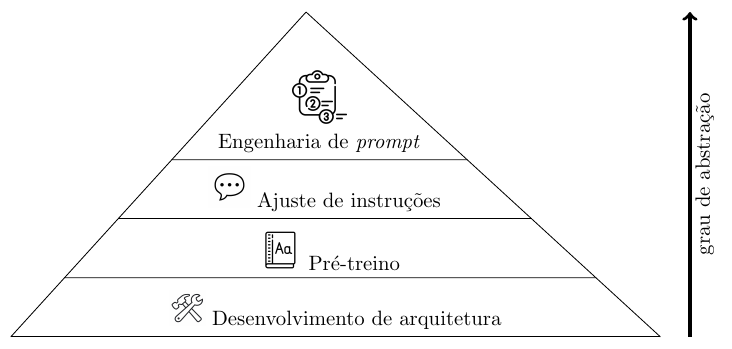}
    \caption[]{Hierarquia da customização de um LLM\footnote{\url{https://www.linkedin.com/pulse/part-2-llm-training-data-kedar-bhumkar-pv9wc/}}.}
    \label{fig:pyramid}
\end{figure}

A Figura \ref{fig:pyramid} mostra a hierarquia de customização de um LLM. Inicialmente, o modelo é pré-treinado em trilhões de palavras, o que equivaleria a um humano lendo sem parar por 8 mil anos, de forma a aprender um ou mais idiomas. São tarefas auto-supervisionadas, em que o próprio processamento do texto é diretamente uma resposta, como predizer a próxima palavra. Essa etapa representa 90\% do custo computacional, o que representa a eletricidade consumida por 200 brasileiros em um ano \cite{nogueira2023lecture}. 

Posteriormente, o modelo é feito o ajuste fino para tarefas conversacionais para se tornar um assistente com 1 milhão de exemplos. A última etapa seria como adaptar a instrução do modelo no uso em inferência para gerar a resposta, conhecido como engenharia de \textit{prompt} \cite{nogueira2023lecture}.

\subsection{Engenharia de \textit{Prompt}}
\begin{figure}[ht]
     \centering
\includegraphics{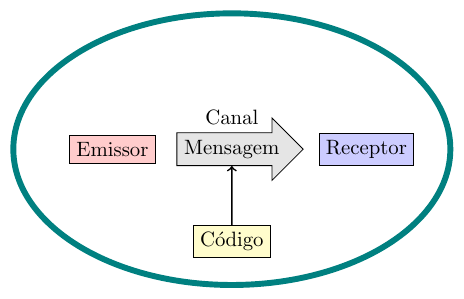}
     \caption{Os elementos básicos da comunicação segundo Jakobson \cite{jakobson1961linguistics}, adaptado de \cite{mirsarraf2017elements}. Estes elementos podem inspirar a construção de \textit{prompts} eficazes para LLMs.}
     \label{fig:communication}
 \end{figure}

A Engenharia de \textit{Prompt} (\textit{Prompt Engineering}) é a prática de projetar cuidadosamente as entradas (instruções ou \textit{prompts}) fornecidas a um LLM para obter as saídas desejadas \cite{zhao2023LLMSurvey}. Como LLMs são sensíveis à forma como a tarefa é descrita, um bom \textit{prompt} pode melhorar significativamente o desempenho do modelo em uma tarefa específica, sem a necessidade de re-treinamento. A formulação de um \textit{prompt} eficaz pode ser vista como análoga aos elementos da comunicação descritos por Jakobson \cite{jakobson1961linguistics}, como ilustrado na Figura \ref{fig:communication}.

\begin{description}
    \item[\colorbox{pink}{Emissor}:] A entidade que formula a instrução para o LLM.
    \item[\colorbox{black!20}{Mensagem}:] O conteúdo do \textit{prompt}, incluindo a tarefa, o contexto e os exemplos.
    \item[\colorbox{brown!50}{Canal}:] A interface de entrada do LLM.
    \item[\colorbox{blue!30}{Receptor}:] O LLM que processa o \textit{prompt}.
    \item[\colorbox{yellow!30}{Código}:] A linguagem natural (e possivelmente formatos específicos) compreendida pelo LLM.
    \item[\colorbox{teal!40}{Contexto}:] Informações adicionais ou restrições que guiam a resposta do LLM.

\end{description}

A Figura \ref{fig:communication} demonstra um exemplo de \textit{prompt} aplicando técnicas como a definição de um papel (\textit{role} - Emissor) e a especificação do formato e tom da resposta (Mensagem, Código, Contexto) direcionada a um público específico (Receptor). \novo{A clareza, especificidade e, por vezes, a inclusão de exemplos (\textit{few-shot prompting}) são cruciais para guiar LLMs eficazmente} \cite{meta2024prompt}, especialmente em tarefas como a extração precisa de alegações factuais, central neste trabalho.

\begin{figure}[ht]
    \centering
\begin{tcolorbox}[colback=black!1]
 \colorbox{pink}{Sou o CEO de uma empresa de médio porte}. \\[2mm]
 Escreva um \colorbox{brown!50}{e-mail} \colorbox{teal!40}{curto, bem-humorado e profissional} para \colorbox{blue!30}{meu gerente regional}.\\[2mm]
 Peça a ele que:
\begin{itemize}
\setlength\itemsep{2mm}
\item Me envie uma lista atualizada do nosso estoque de papel.
\item Organize uma reunião esta semana com outros gerentes regionais.
\item Me informe como foi o \textit{workshop} de IA em toda a empresa em seu escritório.
\end{itemize}
\end{tcolorbox}
    \caption[]{Exemplo de um \textit{prompt} para um LLM, destacando diferentes componentes inspirados nos elementos da comunicação. Adaptado de \url{https://learnprompting.org/docs/intro}.}
    \label{fig:prompt}
\end{figure}

\novo{A clareza, especificidade e, por vezes, a inclusão de exemplos são cruciais para guiar LLMs eficazmente} \cite{meta2024prompt}, especialmente em tarefas complexas como a extração precisa de alegações factuais de textos noticiosos.

\section{Recuperação de Informação (RI)}
\label{sec:ir}
A Recuperação de Informação (RI), ou \textit{Information Retrieval} (IR), é a área de estudo focada em encontrar material (geralmente documentos) de natureza não estruturada (geralmente texto) que satisfaça uma necessidade de informação a partir de grandes coleções (geralmente armazenadas em computadores) \cite{manning2008ir_book}. A RI situa-se na interseção entre ciência da computação, banco de dados, PLN e ciência da informação \cite{bpln2024ir}.

\novo{No contexto da verificação de fatos, a RI é a tecnologia central para a etapa de Recuperação de Evidências} \cite{guo2022survey_afc}. Dada uma alegação, um sistema de RI é usado para buscar e ranquear documentos (e.g., artigos de notícias, páginas web, posts de blog) de um índice que possam conter informações relevantes para verificar essa alegação.

Quando a busca é textual, o mecanismo de busca está diretamente atrelado aos modelos de linguagem. A consulta (\textit{query}) é transformada em uma representação vetorial pelo modelo de linguagem e comparada via métricas de similaridade com as representações vetoriais dos documentos candidatos \cite{karpukhin2020dpr}. No contexto de LLMs, a integração de RI para fornecer informações externas ao modelo antes da geração de uma resposta é conhecida como \textit{Retrieval Augmented Generation} (RAG) \cite{lewis2020rag, reimers2019sentencebert, bpln2024ir}.

Mecanismos de busca web, como o buscador Google, são implementações de sistemas de RI em larga escala. Além da relevância textual (calculada usando técnicas de RI), eles consideram muitos outros fatores para ranquear os resultados, como a popularidade e autoridade da página (historicamente medida por algoritmos como o PageRank \cite{page1999pagerank}), a localização e o histórico do usuário, o frescor do conteúdo, entre outros \footnote{ \url{https://developers.google.com/search/docs/fundamentals/how-search-works\#ranking}}. \novo{APIs fornecidas por esses mecanismos de busca (como a API de Busca do Google e a API de busca de alegações do Google FactCheck, utilizadas neste trabalho)}\correcao{, conforme mecionado na Seção \ref{sec:search},}{ são ferramentas valiosas para implementar a etapa de recuperação de evidências em sistemas de AFC/SAFC, permitindo acesso programático a vastos índices da web e a repositórios de verificações de fatos existentes}.
\chapter{Trabalhos Relacionados}
\label{c:related}
Neste capítulo, são introduzidos os trabalhos relacionados à classificação de \textit{fake news} utilizando técnicas de PLN. De forma geral, apresentamos as técnicas na Seção \ref{sec:related_nlp}. Na Seção \ref{sec:related_llm}, são explanadas pesquisas de detecção de notícias falsas com o uso de IA generativa em texto e, na Seção \ref{sec:related_pt} são explicados trabalhos em português para checagem de texto.

\section{PLN para detecção de notícias falsas}
\label{sec:related_nlp}
De forma geral, as técnicas em PLN para detecção de notícias falsas podem ser agrupadas pelas seguintes abordagens de análise:

\noindent\textbf{Linguística e estilo.} As técnicas que analisam a linguística e o estilo examinam dados sintáticos, léxicos, psico-linguísticos e semânticos. No domínio sintático, é examinado o uso de categorias de palavras como substantivos, verbos e adjetivos pela técnica de \textit{POS-tagging}. No domínio léxico, são extraídas diversas características como número de palavras únicas e suas frequências, número de frases e erros gráficos para detectar textos suspeitos. As informações psico-linguísticas podem ser obtidas por meio do sistema estatístico LIWC \cite{reis2019nlp}.

\noindent\textbf{Análise de sentimentos.} Esta abordagem é semântica e procura detectar vieses, como discurso de ódio e toxicidade \cite{addullah2023sent}.  O estado da arte na subárea de análise de sentimentos, assim como na grande área de PLN, envolve o uso de modelos contextuais baseados em Transformers \cite{wang2018glue}. 

\noindent\textbf{Verificação de fatos (\textit{Fact-Checking}).} O processo de \textit{fact-checking} automatizado, definido na Seção \ref{subsec:techs}, é, em alguns aspectos, análogo a sistemas de perguntas e respostas, nos quais, dada uma pergunta (ou alegação), procuram-se referências para gerar uma resposta (ou veredito) \cite{karpukhin2020dpr}. Dentre as abordagens utilizadas no processo de \textit{fact-checking} automatizado, algumas técnicas notáveis são discutidas abaixo.
 
\noindent\textbf{Busca estatística.} A busca por evidências pode utilizar métodos baseados em palavras-chave, como o TF-IDF, ou procedimentos análogos ao sistema de busca PageRank \cite{reddy2022zero_shot}.
 
\noindent\textbf{Busca contextual.} A busca por contra-evidências é frequentemente realizada por modelos contextuais do tipo Encoder, como os Sentence-Transformers, adaptados para tarefas de busca e recuperação de informação semanticamente relevante \cite{reddy2022zero_shot, deka2022medical_afc, reimers2019sentencebert, karpukhin2020dpr}.

\noindent\textbf{Gerador de resposta extrativo.} A resposta final é gerada pela extração das partes relevantes da contra-evidências pelos modelos Transformers Encoder treinados em tarefas análogas ao conjunto de dados SQuAD~\cite{deka2022medical_afc, devlin2019bert}. 
 
\noindent\textbf{Modelos generativos.} \novo{Como será detalhado na próxima seção,} modelos generativos podem ser utilizados em diversas etapas da cadeia de PLN associada a \textit{fake news}, como gerar explicações para as contra-evidências ou gerar uma pergunta que explicite a alegação feita pela notícia, facilitando o processo de busca e verificação \cite{chen2022gere, ousidhoum2022varifocal}.

\section{Geração na verificação de fatos}
\label{sec:related_llm}

Modelos de Linguagem Grandes (LLMs) têm se tornado peças centrais em tarefas de verificação de fatos. \novo{Atuam como uma alternativa mais econômica em comparação a anotadores humanos especializados, como jornalistas ou especialistas da área}. Em muitos casos, o desempenho dos LLMs \novo{é comparável ao de anotações realizadas via \textit{crowd-sourcing}} \cite{maia2024extension, ni2024afacta}. Além disso, LLMs permitem a geração de textos explicativos \novo{e, frequentemente, não necessitam de treinamento supervisionado específico para a tarefa (abordagem \textit{zero-shot})} \cite{gangi2022claim_zero, kim2024llm_explain, zeng2024justilm, chen2023llm_fake_survey}.

Contudo, modelos como o GPT-3.5 podem apresentar resultados inferiores a modelos de linguagem menores (SLMs) treinados especificamente para a tarefa, como o BERT, na predição de \textit{fake news} sem o uso de evidências externas \cite{hu2024llm_fake}. Isso indica que, para tarefas de classificação de texto, o estado da arte ainda pode favorecer SLMs treinados com exemplos suficientes (tipicamente entre 100 e 1000 exemplos) \cite{qiu2024chatgpt_bert, yang2024llm_survey}.

Ao utilizar um LLM para processar alegações com base em seu conhecimento de mundo intrínseco (sem evidência externa) e, subsequentemente, usar um modelo como o BERT para a verificação final, os resultados podem superar o uso isolado do BERT \cite{hu2024llm_fake}. Incluir no \textit{prompt} instruções para o LLM considerar o estilo de escrita e o conhecimento de mundo também demonstrou melhorar os resultados \cite{liu2024llm_fake}.

\novo{No processo semi-automatizado de verificação de fatos \cite{guo2022survey_afc}, LLMs podem ser utilizados em diversas etapas}: identificação de alegações \cite{kotitsas2024lora_claim, ni2024afacta}, recuperação de evidências (muitas vezes delegada a mecanismos de busca externos como Google ou Bing \cite{wang2024factcheckbench, li2024self_check, tan2023llm_pipeline}), verificação da alegação (a detecção em si) \cite{tan2023llm_pipeline, choi2024claim_matching, krishna2022proofver, wang2024factcheckbench} e justificativa da verificação \cite{kim2024llm_explain, zeng2024justilm, krishna2022proofver, wang2024factcheckbench}. A seguir, algumas etapas são detalhadas:
\\\
\noindent\textbf{Identificação de alegação}. Nesta etapa, o modelo de linguagem extrai as informações que devem ser verificadas (alegações) do texto. Essa etapa será detalhada na Seção \ref{subsec:claim_extraction}.

\noindent\textbf{Verificação da Alegação}. Com a alegação e as evidências relevantes, o LLM pode ser utilizado para detectar se o que foi alegado é falso ou não \cite{tan2023llm_pipeline, choi2024claim_matching, tan2025coling}. A detecção pode considerar também o raciocínio lógico associado, como o uso de operações lógicas \cite{krishna2022proofver, wang2023reasoning} e técnicas de \textit{prompting} ligadas ao raciocínio, como a cadeia de pensamento (\textit{chain-of-thought}) \cite{choi2024claim_matching, wang2023reasoning} e o LLM como um juiz (\textit{LLM-as-a-judge}) \cite{ni2024afacta}.

\noindent\textbf{Justificativa de verificação}. A conclusão da verificação pode ser justificada com o uso de LLMs, que interpretam os insumos (alegação, evidências, veredito) e geram uma explicação para o usuário \cite{kim2024llm_explain, zeng2024justilm}. Para textos complexos, como na área médica, também são sugeridos processos automáticos de simplificação de textos complexos para torná-los acessíveis à população leiga \cite{schlicht2023health_systematic}.

\subsection{Extração de Alegação via LLM}

\label{subsec:claim_extraction}

Na identificação de alegação, também conhecida como extração de alegação, podem ser extraídas uma alegação principal ou múltiplas alegações. Alternativamente, como a próxima etapa é buscar pela alegação, essa etapa também pode ser vista como geração de consulta (\textit{query generation}) ou geração de perguntas (\textit{question generation}) \cite{cho2022qg, ousidhoum2022varifocal}.

\correcaoPrevia{Uma técnica intimamente relacionada à extração de alegações é a normalização de alegações (\textit{claim normalization}), que visa simplificar textos, especialmente de redes sociais, removendo redundâncias e ambiguidades para obter a essência da afirmação \cite{sundriyal2023}. A relevância desta tarefa foi destacada na conferência CheckThat! 2025, que organizou uma tarefa compartilhada sobre normalização de alegações em 20 idiomas, incluindo o português. Como produto paralelo desta dissertação de mestrado, a equipe AKCIT-FN participou desta competição, alcançando o pódio em quinze idiomas e obtendo o terceiro lugar em português \cite{CheckThat:ECIR2025, clef-checkthat:2025-lncs, clef-checkthat:2025:task2}.}

A extração de uma única alegação é geralmente denominada detecção de alegação (\textit{claim detection}) \cite{kotitsas2024lora_claim, ni2024afacta}. No sentido de extrair múltiplas, chama-se decomposição de alegação (\textit{claim decomposition}) ou quebra de alegações (\textit{claim split}) \cite{tang2024minicheck, kamoi2023wice, scirè2024fenice}. Na quebra em múltiplas contextualizações, uma abordagem é remover a dependência semântica entre as alegações, processo conhecido como descontextualização (\textit{decontextualization}) \cite{choi2021decontextualization, wang2024factcheckbench}.

Pode-se gerar diferentes alegações focando em diferentes entidades no texto, formando distintos pontos focais para gerar mais informações relevantes \cite{ousidhoum2022varifocal}. Outra forma de realizar a quebra seria por meio de \textit{chain-of-thought} ou lógica de primeira ordem para decompor a alegação em múltiplas perguntas \cite{wang2023reasoning}.

Único conjunto de dados em português que possui múltiplas alegações é FactNews, em que cada notícia é analisada ao nível de frase. As alegações são extraídas com o uso de anotadores humanos e os dados são avaliados com o modelo BERT \cite{vargas2023factnews}.

Ademais, Ni et al. argumentam que existem discrepâncias sobre o que constitui uma alegação: alguns pesquisadores consideram apenas afirmações factuais, enquanto outros incluem também opiniões com impacto social. A maioria dos estudos considera uma alegação como algo digno de verificação (\textit{check-worthy}), mas a definição do que é relevante para checar também é considerada subjetiva \cite{ni2024afacta}.

Nesse contexto, Ni et al. listam explicitamente os tipos de fatos a serem considerados como alegações em discursos políticos e instruem o LLM a pensar passo a passo na extração de alegação, usando \textit{chain-of-thought} na extração da alegação \cite{ni2024afacta}. A primeira etapa do raciocínio seria extrair a alegação em si e depois indicar qual tipo de fato a alegação extraída contém. Na Seção \ref{sec:claim_extraction}, são explicitados os principais modelos de \textit{prompts} encontrados na literatura.

\section{Trabalhos em português}
\label{sec:related_pt}
Em setembro de 2023, foram identificados 18 conjuntos de dados de \textit{fake news} em língua portuguesa mencionados em artigos científicos e disponíveis publicamente, conforme apresentado na Tabela \ref{tab:found}. Esses conjuntos de dados exibem uma variedade de domínios, fontes e tarefas, embora o ano de publicação esteja concentrado entre 2018 e 2020.

\begin{table}[ht]
\centering
\resizebox{0.95\textwidth}{!}{%
\begin{tabular}{@{}clccccc@{}}
\toprule
\textbf{Publicação} & \textbf{Nome} & \textbf{Domínio} & \textbf{Veículo} & \textbf{Alegação} & \textbf{Evidência} & \textbf{Ano dos dados} \\ \midrule
12/2018 & Fake.Br \cite{monteiro2018fakebr} & Geral & Notícias &  & \cmark & 2016 - 2018 \\
10/2019 & Factck.BR \cite{moreno2019factck.br} & - & Notícias &  & \cmark & - \\
10/2019 & FakeTweet.Br \cite{cordeiro2019faketweetbr} & - & Twitter & \cmark & \cmark & - \\
12/2019 & Bracis2019FakeNews \cite{faustini2019bracisfake} & Política & WhatsApp, Twitter &  & \cmark & 2018 - \\
07/2020 & fake news Multilabel \cite{morais2020fakenewsmultilabel} & Eleições & Notícias &  & \cmark & - \\
11/2020 & FakeNewsSetGen \cite{silva2020FakeNewsSetGen} & - & Twitter & \cmark & \cmark & 2017 - 2020 \\
11/2020 & MM-COVID-PT \cite{li2020mmcovid} & COVID-19 & Twitter & \cmark  & \cmark & - \\
06/2020 & FakeCovid-PT \cite{shahi2020fakecovid} & COVID-19 & Redes sociais Digitais & \cmark & \cmark & 2020 \\
04/2021 & FakeWhatsApp.Br \cite{cabral2021fakewhastapp} & Eleições & WhatsApp & \cmark &  & 2018 \\
07/2021 & Dataset-fake-news \cite{almeida2021dataset-fake-news} & Política & Notícias & \cmark &  & - \\
10/2021 & COVID19.BR \cite{martins2021covid19br} & COVID-19 & WhatsApp & \cmark  &  & 2020 \\
10/2021 & Central de Fatos \cite{couto2021central_de_fatos} & Geral & Notícias &  & \cmark & 2013 - 2021 \\
02/2022 & MuMiN-PT \cite{nielsen2024mumin} & Geral & Twitter & \cmark & \cmark & 2020-2022 \\
03/2022 & FakeRecogna \cite{garcia2022FakeRecogna} & Geral & Notícias & \cmark &  & 2019 - 2021 \\
03/2022 & Fakepedia \cite{charles2022Fakepedia} & Geral & Notícias &  & \cmark & 2013 - 2021 \\
06/2022 & Fact-check\_tweet-PT \cite{kazemi2022Fact-check_tweet} & - & Twitter & \cmark  & \cmark & - \\
06/2022 & SIRENE-news \cite{barbosa2022SIRENE-news} & Geral & Notícias &  & \cmark & 2019 \\ 
09/2023 & FactNews \cite{vargas2023factnews} & Gerak & Notícias & \cmark &  &  2006 - 2007 e 2021 - 2022 \\ \bottomrule
\end{tabular}
}
\caption{Conjuntos de dados de \textit{fake news} em português identificados (até Set/2023).}
\label{tab:found}
\end{table}

Esses conjuntos de dados em língua portuguesa exibem uma variedade de domínios, fontes e tarefas, embora o ano de publicação esteja concentrado entre 2018 e 2020. A análise revela que aproximadamente metade dos trabalhos verifica alegações provenientes de notícias de fontes estabelecidas, enquanto os demais focam em contextos de redes sociais digitais, notadamente Twitter e WhatsApp. Quando mencionados, os domínios dos dados distribuem-se amplamente, com ênfase em tópicos como saúde (especialmente COVID-19), política/eleições, entretenimento, e assuntos gerais nacionais e internacionais.

No que tange às tarefas de PLN e aos dados fornecidos, a maioria dos \textit{corpora} (12 dos listados) fornece apenas as alegações a serem verificadas, sem evidências associadas. Dentre estes, somente o FactNews decompõe cada exemplo em múltiplas alegações ao nível de frase \cite{vargas2023factnews}. 

Por outro lado, alguns conjuntos \cite{cordeiro2019faketweetbr, silva2020FakeNewsSetGen, shahi2020fakecovid, nielsen2024mumin} oferecem as alegações e as respectivas evidências associadas, geralmente provenientes de agências de verificação de fatos ou fontes confiáveis que validam ou refutam as alegações. \novo{No entanto, mesmo os conjuntos que  frequentemente fornecem apenas identificadores de tweets ou referências indiretas, tornando o acesso direto às evidências desafiador ou inviável, especialmente após as mudanças na API do Twitter/X em 2023.}

\begin{table}[ht]
\centering
\resizebox{0.5\textwidth}{!}{%
\begin{tabular}{lcc}
\toprule
Agência & Região & Menções \\ \midrule
Lupa\footnote{\url{https://lupa.uol.com.br/}} & Brasil & 8 \\
Boatos.org\footnote{\url{https://boatos.org/}} & Brasil & 7 \\
Aos fatos\footnote{\url{https://www.aosfatos.org/}} & Brasil & 5  \\
Projeto Comprova\footnote{\url{https://projetocomprova.com.br/}} & Brasil & 4 \\
AFP: Checamos\footnote{\url{https://checamos.afp.com/}} & Brasil & 3 \\
G1: Fato ou Fake\footnote{\url{https://g1.globo.com/fato-ou-fake/}} & Brasil & 2 \\
Estadão Verifica\footnote{\url{https://www.estadao.com.br/estadao-verifica/}} & Brasil & 1 \\
UOL Confere\footnote{\url{https://noticias.uol.com.br/confere/}} & Brasil & 1 \\
Pública: Truco\footnote{\url{https://apublica.org/tag/truco/}} & Brasil & 1  \\
Observador: Fact-checks\footnote{\url{https://observador.pt/factchecks/}} & Portugal & 1  \\
 E-farsas\footnote{\url{https://www.e-farsas.com/}} & Brasil & 1  \\ 
\bottomrule
\end{tabular}
}%
\caption[]{Agências e menções nos \textit{corpora}.}
\label{tab:agencies}
\end{table}

As agências verificadoras de \textit{fake news} mencionadas explicitamente nos trabalhos são listadas na Tabela \ref{tab:agencies}. Dentre elas, a Lupa e a Boatos.org são as mais citadas, sendo empregadas diretamente em sete e oito trabalhos, respectivamente. Três agências são das maiores mídias jornalísticas tradicionais brasileiras, como sendo G1: Fato ou Fake, Estadão Verifica e UOL Confere. 

Três são de portuguesas como Observador: \textit{Fact-checks}, Polígrafo e Viral. AFP: Checamos é uma filial brasileira da agência francesa AFP. Além disso, o Viral é o única agência jornalística de domínio específico, no qual é saúde.

É importante ressaltar que os conjuntos de dados MM-COVID, FakeCovid, MuMinLarge e Fact-check\_tweet são multilíngues e incorporam agências de verificação de fatos internacionais provenientes de listas confiáveis, como a Polynter\footnote{\url{https://www.poynter.org/ifcn/}} e o Google Fact Check\footnote{\url{https://toolbox.google.com/factcheck/explorer}} \cite{li2020mmcovid, shahi2020fakecovid, nielsen2024mumin, kazemi2022Fact-check_tweet}. Por exemplo, a Polynter lista agências como Lupa, Observador, AFP e Estadão Verifica. Portanto, embora não mencionem explicitamente as agências lusófonas, esses conjuntos de dados as utilizam indiretamente.

Entre os conjuntos de dados monolíngues em português, o Fake-Recogna extrai informações de agências confiáveis de uma lista internacional chamada Duke Reporter's Labs\footnote{\url{https://reporterslab.org/fact-checking/}} \cite{couto2021central_de_fatos}. O conjunto de dados Central de Fatos menciona a lista Polynter, mas também utiliza verificadores não presentes na listagem \cite{couto2021central_de_fatos}.

\novo{A dificuldade em acessar evidências consistentes e diretamente utilizáveis na maioria dos conjuntos de dados em português existentes motiva a abordagem deste trabalho, que visa não apenas identificar alegações, mas também recuperar e associar novas evidências da web a essas alegações, contribuindo para o enriquecimento dos recursos disponíveis para a pesquisa em verificação de fatos na língua portuguesa.}
\chapter{Métodos}
\label{c:methods}

Este capítulo detalha os procedimentos metodológicos empregados para o enriquecimento de \textit{corpora} destinados à detecção de notícias falsas (\textit{fake news}) em língua portuguesa. O enriquecimento é realizado por meio da incorporação de informações contextuais externas, provenientes de Modelos de Linguagem Grandes (LLMs) e mecanismos de busca.

A Seção \ref{sec:general_flow} apresenta o fluxo geral do processo de enriquecimento dos dados. Subsequentemente, a Seção \ref{sec:data} descreve os critérios para a seleção dos conjuntos de dados, o pré-processamento aplicado e o processo de validação semi-automática realizado. A Seção \ref{sec:claim_extraction} define o método de extração de alegações centrais dos textos. Adiante, a Seção \ref{sec:search} elucida a estratégia adotada para a recuperação de evidências, englobando a busca inicial na \textit{web}, a extração condicional de alegações e a utilização de Interfaces de Programação de Aplicações (APIs) específicas para verificação de fatos. Por fim, a Seção \ref{sec:estrategia_avaliacao} explicita a metodologia de avaliação experimental, concebida para mensurar a eficácia das etapas propostas de validação e de enriquecimento com conteúdo externo.

\section{Fluxo de Enriquecimento}
\label{sec:general_flow}

\correcaoPrevia{O fluxo de enriquecimento proposto é fundamentalmente inspirado no \textit{pipeline} canônico de Verificação Semi-Automática de Fatos (SAFC), conforme detalhado na Seção \ref{subsubsec:fc}. O processo de SAFC trenvolve tradicionalmente astapas de Extração de Alegação, Recuperação de Evidências, Predição de Veredito e Geração de Justificativa. O método foca nas duas primeiras etapas — Extração de Alegação e Recuperação de Evidências — essenciais para coletar o contexto externo necessário para a verificação.}

\correcaoPrevia{A principal inovação metodológica desta dissertação reside na implementação de um fluxo adaptativo, que busca otimizar a eficiência e os custos computacionais. A premissa central é que a etapa de Extração de Alegação, embora crucial para textos longos ou coloquiais, é redundante para textos que já são, em sua essência, alegações concisas e verificáveis. Por exemplo, uma postagem em rede social repleta de opiniões e ruído necessita que sua alegação factual seja isolada para uma busca eficaz \cite{clef-checkthat:2025:task2, sundriyal2023}. Em contrapartida, um texto curto de um \textit{corpus} de verificação de fatos, como "O novo coronavírus pode ser transmitido através de encomendas enviadas da China" da Figura \ref{fig:exemplo1}, já funciona como uma consulta de alta qualidade para um mecanismo de busca.}

\begin{figure}
    \centering
    \includegraphics{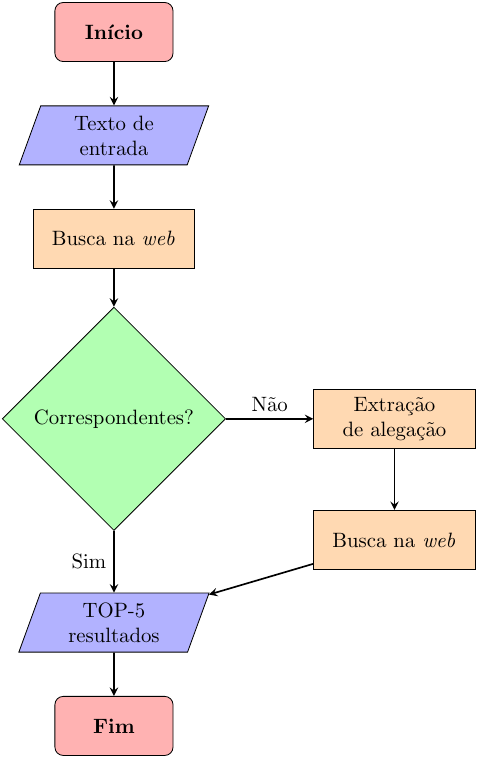}
    \caption{Diagrama de fluxo geral do processo de enriquecimento dos \textit{corpora}.}
    \label{fig:process}
\end{figure}

\correcaoPrevia{Para operacionalizar essa lógica, o} fluxo geral do processo de enriquecimento é ilustrado na Figura \ref{fig:process}.  Inicialmente, uma consulta derivada do texto de entrada é submetida a um mecanismo de busca na \textit{web}. Avalia-se a correspondência entre os resultados obtidos (considerando os cinco primeiros) e o texto original. Se for identificada uma correspondência significativa, assume-se que a busca inicial recuperou informações diretamente relevantes, e os resultados são armazenados. Caso contrário, procede-se à etapa de extração de alegação, na qual um LLM é utilizado para identificar a afirmação central do texto de entrada. Esta alegação extraída é, então, empregada como uma nova consulta para uma segunda busca na \textit{web}. Adicionalmente, realiza-se uma busca específica em uma API de verificação de fatos.

\textbf{Busca na web.} Optou-se pelo uso da API oficial de busca do Google, Google Custom Search Engine (CSE), por conta dos créditos gratuitos iniciais. \novo{Adicionalmente, os textos foram buscados na API de busca por alegações do Google, Google FactCheck Claim Search,
\footnote{\scriptsize \url{https://developers.google.com/fact-check/tools/api\#the-google-factcheck-claim-search-api}} de forma gratuita. Caso o texto inicial a ser buscado no Google FactCheck não retornar resultados e houver alegação previamente extraída na busca principal, a alegação será buscada no Google FactCheck.}

\textbf{Correspondência.} \correcaoPrevia{Avalia-se a presença dos termos da consulta com exceção de \textit{stopwords}) nos fragmentos destacados pelos marcadores HTML \texttt{<b>} e \texttt{</b>} dos resultados da API do Google CSE. Na prática, se 80\% ou mais desses termos estiverem presentes nos fragmento em destaque, o processo de extração de afirmações não é executado.}

\textbf{Extração de alegação.}  Para a extração de alegação, são \correcaoPrevia{extraídas} até as 75 primeiras palavras. Dos cinco primeiros resultados obtidos. Optou-se pelo uso do Gemini 1.5 Flash por conta dos créditos gratuitos \correcaoPrevia{do} Google Cloud. Foi utilizado também o framework Langchain\cite{chase2022langchain} para a engenharia de \textit{prompt}. Para a extração de alegação, são extraídas até as 75 primeiras palavras.

\correcao{A metodologia foi desenhada para extrair uma única alegação central do texto, simplificando o processo a uma única busca subsequente por evidências. Consequentemente, a abordagem não contempla a separação de múltiplas alegações (\textit{claim splitting}), uma simplificação intencional do fluxo que será discutida como limitação na Seção \ref{sec:limitations}. Adicionalmente, a reprodutibilidade e a generalização dos resultados são afetadas pela dependência de tecnologias específicas (Gemini 1.5 Flash, APIs do Google), um desafio agravado pela ausência de uma semente determinística (\textit{seed}) na API do Gemini e pela natureza personalizada da busca do Google.}

\section{Conjunto de dados}
\label{sec:data}
Selecionaram-se conjuntos de dados de detecção de \textit{fake news} em textos com os seguintes critérios: (i) públicos e acessíveis, (ii) possuam pelo menos um mil exemplos, (iii) possuam artigos publicados. Foram excluídos conjuntos de dados que consistiam somente em reportagens de agências verificadoras sobre \textit{fake news}, em vez das notícias ou alegações originais.

\begin{table}[ht]
\resizebox{\textwidth}{!}{
\begin{tabular}{lcccccc}
\toprule
Conjunto de dados & Domínio & Veículo & Abordagem \cite{hangloo2022survey} & Tamanho por rótulo & Ano dos dados & Agências mencionadas \\ \midrule
Fake.br \cite{monteiro2018fakebr}  & Geral & \makecell{Páginas\\da web} & \textit{bottom-up} & \makecell{\correcaoPrevia{3600 \texttt{true}}\\\correcaoPrevia{3,600 \texttt{fake}}} & 2016 -- 2018 & \makecell{Estadão\\Folha\\G1} \\[8mm]
COVID19.BR \cite{martins2021covid19br} & Saúde & WhatsApp & \textit{bottom-up} & \makecell[c]{\correcaoPrevia{1987 \texttt{true}}\\\correcaoPrevia{912 \texttt{fake}}} & 2020 & \makecell{Boatos.org\\Lupa} \\[7mm]
MuMiN-PT \cite{nielsen2024mumin} & Geral & \makecell{Twitter\\(atual X)} & \textit{top-down} & \makecell[c]{\correcaoPrevia{1339 \texttt{fake}}\\\correcaoPrevia{65 \texttt{true}}} & 2020 -- 2022 & \makecell{AFP\\Aos Fatos\\Comprova\\Observador\\O Globo\\Piauí\\Uol} \\ \bottomrule
\end{tabular}
}
\caption{Conjuntos de dados selecionados para o estudo. \correcaoPrevia{As abordagens de coleta seguem a classificação por \cite{hangloo2022survey}, em que a \textbf{top-down} envolve a coleta de publicações para \textit{fake news} conhecidos e de longa data (frequentemente a partir de sites de checagem de fatos), e a \textbf{bottom-up} consiste em reunir todas as publicações relevantes de um determinado período para identificar a \textit{fake news} emergente}.  As publicações originais listam fontes de verificação, mas não fornecem os excertos específicos de evidência usados para checar cada item.}
\label{tab:datasets}
\end{table}

Com base nesses critérios, três \textit{corpora} em língua portuguesa foram selecionados, cujas características são resumidas na Tabela \ref{tab:datasets}:
\begin{enumerate}
    \item \textbf{Fake.br} \cite{monteiro2018fakebr}: Composto por notícias de páginas web, abrangendo domínios gerais, coletadas entre 2016 e 2018. Utiliza uma abordagem \textit{bottom-up}, coletando diretamente o conteúdo das páginas.
    \novo{Para cada \textit{fake news} obtida, foi extraídas uma notícia verdadeira correspondente. Com isso, o \textit{corpus} consiste em 3600 pares de \textit{fake news} e a notícia verdadeira associada.}
    
    \item \textbf{COVID19.BR} \cite{martins2021covid19br, sa2021iceis}: Contém mensagens da plataforma WhatsApp de 236 grupos públicos focadas no tema da saúde (COVID-19) entre junho e abril de 2020. Também segue uma abordagem \textit{bottom-up}. Possui o campo de fonte da agência mas é nulo em 98\% dos dados.
    
    \item \textbf{MuMiN-PT} \cite{nielsen2024mumin}: Formado por \textit{tweets} (da plataforma X, anteriormente Twitter) de domínio geral, coletados entre 2020 e 2022. \correcaoPrevia{Ele representa um contraste metodológico crucial com os outros, pois foi coletado por meio de uma abordagem top-down \cite{hangloo2022survey}}, encontrando publicações correspondentes a alegações verificadas por checadores de fatos. Trata-se de um subconjunto em língua portuguesa do \textit{corpus} multilíngue MuMiN, extraído utilizando a ferramenta Lingua \cite{stahl2023lingua}.
\end{enumerate}

\novo{
A diversidade desses conjuntos de dados -- em termos de fonte (notícias web, WhatsApp, X/Twitter), domínio temático (geral, saúde), método de coleta (\textit{top-down}, \textit{bottom-up}) e período temporal -- foi considerada vantajosa para avaliar a generalização da abordagem de enriquecimento proposta.}

\subsection{\correcao{Limpeza} e Validação Semi-automática dos Dados}
\label{subsec:validation}

\correcao{A qualidade e a confiabilidade dos dados foram garantidas por um pipeline de validação semiautomático. Essa metodologia integrou rotinas automatizadas a uma curadoria manual detalhada, realizada por um dos autores, visando identificar e retificar inconsistências. Exemplos práticos que ilustram cada etapa do tratamento são detalhados no Apêndice \ref{append:validation_examples}. As fases do processo foram:}

\begin{enumerate}[noitemsep, topsep=0pt]
\item \textbf{Filtragem Inicial Automatizada:} Remoção de duplicatas exatas, exemplos compostos inteiramente por URLs e textos excessivamente curtos \footnote{Com menos de 15 tokens removendo \textit{emojis}, URLs, \textit{stopwords} e pontuações ao utilizar o tokenizador \texttt{cl100k\_base} da biblioteca tiktoken \url{https://github.com/openai/tiktoken}}.

\item \textbf{Filtragem de Idioma:} \correcao{Identificação automática do idioma português com a biblioteca Lingua \cite{stahl2023lingua}, seguida de verificação manual e exclusão de exemplos em outros idiomas.}

\item \textbf{\correcao{Remoção} de Rótulos Conflitantes em Pares Relacionados:} \correcao{Revisão manual de instâncias com alta similaridade textual, mas com rótulos de veracidade divergentes. A detecção desses pares foi feita por dois métodos:}
        \begin{enumerate}[label=\roman*., noitemsep, topsep=0pt, leftmargin=*]
            \item \textbf{Quase-duplicatas:} \correcao{Identificadas via algoritmo MinHash, como explicado em detalhe na Seção \ref{sec:correpondence}.}
            \item \textbf{Referências de URL:} \correcao{Identificadas quando textos distintos mencionavam a mesma URL.}
        \end{enumerate}

\item \textbf{\novo{Verificação de Rótulos com Fonte Externa:}} \correcao{Revisão manual de exemplos cujos rótulos conflitavam com informações do Google FactCheck.}

\item \textbf{Inspeção de Subconjunto Aleatório:} Verificação manual de um subconjunto aleatório de cada conjunto de dados para avaliar a qualidade geral.

\item \textbf{\novo{Tratamento Específico para o Fake.br:}}
\begin{enumerate}[label=(\alph*), noitemsep, topsep=0pt]
    \item \textbf{Remoção de Quase-Duplicatas da Mesma Fonte:} \correcao{Exclusão de textos quase-idênticos que provinham da mesma URL de origem, uma vez que, neste \textit{corpus}, a fonte primária de cada notícia é uma URL}
        \item \textbf{Correção de Exemplos Incompletos:} \correcao{Remoção de exemplos da versão normalizada do \textit{corpus} que não continham o trecho original completo da notícia \footnote{\url{https://github.com/roneysco/Fake.br-Corpus/issues/7}}.}
    \item \novo{Remoção de textos quase duplicados provenientes da mesma URL de origem, uma vez que, neste \textit{corpus}, a fonte primária de cada notícia é uma URL.}
    \item \textbf{\correcao{Remoção de par correspondente:}} \novo{Dado que o Fake.br é estruturado em pares de notícias (uma falsa e uma verdadeira sobre o mesmo evento), exemplos cujo par correspondente havia sido removido em etapas anteriores do pré-processamento também foram excluídos para manter a integridade da estrutura pareada.}
\end{enumerate}

\novo{Como etapa final de pré-processamento, todas as URLs mencionadas explicitamente nos textos originais foram removidas. Esta medida visou mitigar potenciais vieses introduzidos pelos domínios das URLs, que poderiam ser aprendidos pelos modelos de forma espúria. Uma análise preliminar no \textit{corpus} COVID19.BR, por exemplo, indicou que a simples presença de certos domínios estava fortemente correlacionada com o rótulo de veracidade (conforme Tabela \ref{tab:domain_covid})}.

\end{enumerate}

\novo{A Tabela \ref{tab:validation} detalha a quantidade de exemplos que foram corrigidos ou removidos em cada etapa do fluxo de validação semi-automático descrito: }

\begin{table}[ht]
    \centering
    \begin{tabular}{lccc}
    \toprule
     Etapa de Validação & {COVID19.BR} & {Fake.br} & {MuMiN-PT} \\
    \midrule
    {Filtragem Inicial Automatizada} & 804 & 1 & 0 \\
    {Filtragem de idioma} & 8 & 0 & 11801 \\
    {Resolução de Contradições} & 20 & 0 & 0 \\
    {Verificação Externa de Rótulos} & 23 & 0 & 4 \\
    {Inspeção de Subconjunto} & 88 & 0 & 0 \\
    {Tratamento específico Fake.br} & 0 & 61 & 0 \\
    \bottomrule
    \end{tabular}
    \caption{\novo{Quantidade de exemplos corrigidos ou removidos nos \textit{corpora} durante o fluxo de validação semi-automático.}}
    \label{tab:validation}
\end{table}

\section{Extração de alegação}
\label{sec:claim_extraction}

Foram pesquisados quais \textit{prompts} estão reportados na literatura na área de extração de alegação e de extração da consulta (\textit{query extraction}), definidos na Seção \ref{subsec:claim_extraction}. Buscou-se no Google Scholar em meados de maio de 2024 por palavras-chave \texttt{claim detection}, \texttt{claim split}, \texttt{claim decomposition}, \texttt{query generation} e \texttt{query extraction}. Foram encontrados seis padrões principais de \textit{prompts}:
\begin{description}
    \item[Detecção de alegação \cite{kotitsas2024lora_claim}] O que o texto alega no geral? 
    \item[Separação de alegação \cite{tang2024minicheck, wang2024factcheckbench, kamoi2023wice, scirè2024fenice}] Definimos uma afirmação como uma "unidade elementar de informação em uma sentença, que não precisa ser dividida ainda mais." Segmente os fatos do texto a seguir. 
    \item[Uso de perspectiva (\textit{role}) \cite{kotitsas2024lora_claim}] Você é um assistente que ajuda os jornalistas verificarem \textit{fake news}.
    \item[Explicitar as categorias de fatos \cite{ni2024afacta}] Categorias de fatos: C1. Mencionar que alguém (incluindo ele(a) próprio(a)) fez ou está fazendo algo. C2. Citando quantidades e estatísticas. C3. Alegando correlação ou relação causal. C4. Afirmando leis existentes ou regras de operação. C5. Prometer um plano específico para o futuro ou fazer previsões específicas sobre o futuro.  
    \item[Extração de consulta \cite{abbasiantaeb2024generateretrieve}] Imagine que você está navegando na internet e topa com uma notícia e dá uma lida rápida. O que você buscaria no Google para ver se o que você entendeu no geral da notícia é verdadeira? Extraia o que você buscaria do texto a seguir.
    \item[\textit{Few-shot} \cite{scirè2024fenice}] ENTRADA: O rover Perseverance da NASA descobriu vida microbiana antiga em Marte, de acordo com um estudo recente publicado na revista Science. SAÍDA: {'afirmações': ["O rover Perseverance da NASA descobriu vida microbiana antiga.", "Vida microbiana antiga foi descoberta em Marte.", "A descoberta foi feita de acordo com um estudo recente.", "O estudo foi publicado na revista Science."]}
\end{description}

Para cada padrão de \textit{prompt} foram testadas algumas variações em poucos exemplos dos conjuntos de dados. Não foi utilizado o cenário de múltiplas alegações, o qual envolve a separação de alegação (\textit{claim split}), definido em \ref{sec:claim_extraction}, pois geraria mais de uma busca para ser verificada. Também não foi testado o impacto de explicitar a categoria de fatos.

\begin{figure}[ht]
    \centering
\begin{tcolorbox}[colback=black!1]
 Qual principal fato exposto no texto?
\begin{enumerate}
    \item Extraia um trecho de até 20 palavras do texto a seguir.
    \item Retorne somente a alegação \novo{e sem título}
\end{enumerate}
Texto: \texttt{\{TEXTO DE ENTRADA\}} \\
Alegação:
\end{tcolorbox}
    \caption{\textit{Prompt} final para a extração de alegação.}
    \label{fig:prompt_extraction}
\end{figure}

A Figura \ref{fig:prompt_extraction} mostra o \textit{prompt} desenvolvido para a extração de alegação. Verificou-se que não foi necessário acrescentar o uso de perspectiva (\textit{role}) ou exemplos de \textit{few-shot} para que o modelo extraísse corretamente as alegações em um conjunto de 10 exemplos de teste. 

Para simular uma leitura rápida e fornecer contexto ao LLM, utilizou-se como \texttt{\{TEXTO DE ENTRADA\}} até os três primeiros parágrafos do texto original ou, aproximadamente, até 50 palavras, caso o texto fosse mais longo. Para o retorno, exigiu-se que a saída do modelo (\textit{Alegação}) contivesse até 20 palavras, visando uma consulta de busca concisa e focada.

\novo{Ao utilizar o Gemini 1.5 Flash para extração de alegação, a configuração de segurança foi desativada\footnote{\url{https://ai.google.dev/gemini-api/docs/safety-settings}} para evitar filtragem excessiva de conteúdo. Adicionalmente, observou-se que a documentação da API do LLM (consultada em 29/06/2024) não oferecia opção para fixar uma semente (\textit{seed}) de geração, o que impede o determinismo completo dos resultados para fins de reprodutibilidade\footnote{\scriptsize \url{https://cloud.google.com/vertex-ai/generative-ai/docs/model-reference/gemini?hl=pt-br}}.}

\section{Mecanismos de Busca}
\label{sec:search}

\novo{A recuperação de evidências externas é realizada por meio de dois mecanismos de busca principais: a API Google Custom Search Engine (CSE) para buscas genéricas na \textit{web} e a API Google FactCheck Claims Search, especializada na busca por alegações já verificadas. Ambas as APIs são consultadas com o texto pré-processado (ou a alegação extraída) em português, solicitando-se os cinco primeiros resultados.}

\subsection{Pré-processamento da Consulta de Busca}

\correcao{Para simular a criação de uma consulta única a partir de texto puro, foi desenvolvido um pipeline de pré-processamento baseado em heurísticas.} Este processo é implementado pela função \texttt{preprocess\_query} em Python, apresentada na Figura \ref{fig:prepro}.

\correcao{Em textos com até 20 palavras, o conteúdo integral foi usado como consulta. Para textos mais longos, a primeira sentença era extraída. Se ela fosse longa o suficiente para conter a alegação principal (7 ou mais palavras), tornava-se a consulta. Contudo, se a primeira sentença fosse muito curta (menos de 7 palavras), a consulta era formada pelo que fosse mais longo: o primeiro parágrafo completo ou as 20 primeiras palavras do texto.}

\begin{figure}
    \centering
\begin{minted}[
    breaklines,
    encoding=utf8,
    fontsize=\footnotesize,
    frame=single,
    framesep=3mm,
    python3=true
]
{python}
from nltk import sent_tokenize
import re

SPACES = re.compile(r"\s+")
QUOTES = #todas as representações unicode de aspas

def preprocess_query(text: str) -> str:
    text = text.strip()
    for quote_char in QUOTES:
        text = text.replace(quote_char, "")
    
    words = re.split(SPACES, text)
    
    if len(words) <= 20:
        query = text
    else:
        fst_sent = sent_tokenize(text, language='portuguese')[0]
    
        if len(re.split(SPACES, fst_sent.strip())) >= 7:
            query = fst_sent
        else:
            fst_paragraph = re.split("\n+", text)[0]
            
            if len(re.split(SPACES, fst_paragraph.strip())) < 20:
                query = " ".join(words[:20])
            else:
                query = fst_paragraph

    return query
\end{minted}
    \caption{Código de pré-processamento da frase de busca inicial.}
    \label{fig:prepro}
\end{figure}

Na frase de busca, foram removidos os caracteres UNICODE de aspas simples e duplas, pois alteram o comportamento da busca do Google, forçando correspondência exata de frases\footnote{\scriptsize \url{https://blog.google/products/search/how-were-improving-search-results-when-you-use-quotes/}}. \novo{Também foram retirados emojis\footnote{\url{https://home.unicode.org/emoji/about-emoji/}}, pois experimentalmente notou-se que esses símbolos pioraram alguns resultados de buscas como o da Figura \ref{fig:clickbait}.}

\textbf{Google Custom Search Engine (CSE).} A API Google CSE é utilizada para buscas genéricas na \textit{web}. Além da consulta pré-processada e da solicitação dos cinco primeiros resultados (\texttt{num=5}), especificam-se os seguintes parâmetros para refinar a busca: geolocalização do usuário como Brasil (\texttt{gl=pt-BR}) e idioma preferencial dos resultados como português (\texttt{lr=lang\_pt})\footnote{\scriptsize \url{https://developers.google.com/custom-search/v1/reference/rest/v1/cse/list}}. A Figura \ref{fig:search_result_cse_example} ilustra um exemplo da estrutura de retorno da API CSE. Desta estrutura, são armazenados o título (\texttt{title}), o \textit{link} (\texttt{link}) e o trecho de texto relevante (\texttt{snippet}) de cada resultado.

\begin{figure}
    \centering
\begin{minted}{json}
{
    "htmlTitle": "Cadastro em aplicativo do SUS é recomendado, mas não é ...",
    "link": "https://lupa.uol.com.br/jornalismo/2021/01/12/verificamos-cadastro-sus",
    "htmlSnippet": "Jan 12, 2021 <b>...</b> ... todos os brasileiros precisam se cadastrar no aplicativo Conecte ... “<b>Pessoal</b>, <b>todo mundo precisa se cadastrar no conectesus para vacinar</b>.",
    "pagemap": {
        "metatags": [
        {
            "og:type": "article",
            "og:title": "Cadastro em aplicativo do SUS é recomendado, mas não é obrigatório para a vacinação",
            "og:description": "Circula pelo WhatsApp que todos os brasileiros precisam se cadastrar no aplicativo Conecte SUS para receber a vacina contra o novo coronavírus — caso contrário, não serão imunizados. Por WhatsApp, leitores da Lupa sugeriram que esse conteúdo fosse analisado. Confira a seguir o trabalho de verificação : “Pessoal, todo mundo precisa se cadastrar no conectesus para […]",
        }
        ],
        "ClaimReview": [
        {
            "datePublished": "2021-01-12",
            "claimReviewed": "Pessoal, todo mundo precisa se cadastrar no",
        }
        ]
    }
}
\end{minted}
    \caption{Exemplo de um item da lista de resultados da API Google CSE. Alguns campos foram omitidos para concisão. Consulta realizada em 09/07/2024 para a frase "Pessoal, todo mundo precisa se cadastrar no conectesus para vacinar.".}
    \label{fig:search_result_cse_example}
\end{figure}

\textbf{\novo{Google FactCheck Claims Search}}  \novo{A API Google FactCheck Claims Search é empregada para buscar alegações específicas que já foram verificadas por organizações de checagem de fatos. Os parâmetros utilizados na consulta são o idioma (\texttt{languageCode=pt-BR}) e o número de resultados (\texttt{pageSize=5}). Diferentemente da CSE, esta API não oferece argumentos de localidade adicionais para o português. O pré-processamento do texto de busca é idêntico ao utilizado para a CSE. A Figura \ref{fig:claim_search} apresenta um exemplo da estrutura de retorno desta API. Para cada alegação encontrada, armazena-se apenas o primeiro objeto dentro da lista \texttt{claimReview}, que contém a avaliação da veracidade e o texto da alegação verificada\footnote{\scriptsize \url{https://developers.google.com/fact-check/tools/api\#the-google-factcheck-claim-search-api}}.}

\begin{figure}
    \centering
\begin{minted}{json}
{
    "claims": [
        {
            "claimDate": "2021-04-20T16:52:25Z",
            "claimReview": [
                {
                    "languageCode": "pt",
                    "publisher": {
                        "name": "Observador",
                        "site": "observador.pt"
                    },
                    "reviewDate": "2021-06-29T09:21:38Z",
                    "textualRating": "Errado",
                    "title": "Fact Check. Vacinas “não criam imunidade contra a Covid-19”?",
                    "url": "https://observador.pt/factchecks/..../"
                }
            ],
            "claimant": "Utilizador de Facebook",
            "text": "\"Vacinas contra a Covid-19 não criam imunidade\""
        }
    ]
}
\end{minted}
    \caption{Exemplo de um item da lista de resultados da API Google FactCheck Claims Search. Alguns campos foram omitidos para concisão. Consulta realizada em 08/01/2025 para a frase "Vacinas contra a Covid-19 não criam imunidade.".}
    \label{fig:claim_search}
\end{figure}

\section{\novo{Avaliação dos dados}}
\label{sec:estrategia_avaliacao}

\novo{Com o objetivo de mensurar a eficácia das etapas propostas de validação semi-automática dos dados e de enriquecimento com conteúdo externo, delineou-se um conjunto de configurações experimentais, aplicadas ao COVID19.BR e ao Fake.br, para permitir uma avaliação comparativa e incremental do impacto de cada etapa.  O \textit{corpus} MuMiN-PT foi excluído desta fase experimental devido ao severo desbalanceamento de classes resultante após a filtragem para o idioma português, onde restaram apenas 65 exemplos da classe minoritária (notícias verdadeiras), conforme detalhado na análise exploratória (ver Figura \ref{fig:balancing})}

\novo{As configurações avaliadas foram:}

\begin{enumerate}
    \item \textbf{\novo{Dados Originais (Linha de Base)}}: 
        \novo{Representa os conjuntos de dados com o mínimo pré-processamento necessário para a modelagem, servindo como ponto de partida comparativo. O processamento aplicado limitou-se à remoção de menções a URLs nos textos, justificada pela análise de viés de domínio, discutida na Seção \ref{subsec:validation} e exemplificada na Tabela \ref{tab:domain_covid}. Especificamente para o \textit{corpus} Fake.br, foi empregado o texto normalizado em tamanho, em conformidade com a metodologia de classificação adotada pelos autores originais \cite{monteiro2018fakebr}.}

    \item \textbf{\novo{Dados Validados:}} \novo{Corresponde aos conjuntos de dados resultantes da aplicação integral do fluxo de validação semi-automático detalhado na Seção \ref{subsec:validation}. Esta configuração visa isolar o efeito da melhoria da qualidade intrínseca dos dados (remoção de ruídos, inconsistências, etc.) antes do enriquecimento.}

    \item \textbf{\novo{Dados Validados e Enriquecidos com Conteúdo Externo}}: 
        \novo{Corresponde aos conjuntos de dados obtidos após a execução do fluxo completo ilustrado na \ref{fig:process}, incorporando os resultados de busca relacionada como informação contextual adicional. Essa configuração foi avaliada em dois cenários distintos para investigar o impacto de diferentes tipos de fontes externas}:
        \begin{enumerate}
            \item \textbf{\novo{Enriquecimento Completo:}} \novo{Considera o primeiro resultado de busca obtido através das APIs do Google (CSE e FactCheck).}
            \item \textbf{\novo{Enriquecimento Filtrado (Sem Redes Sociais):}} \novo{Exclui, dos resultados de busca da CSE, aqueles identificados como provenientes de plataformas de redes sociais (e.g., Twitter/X, Facebook). Esta variação permite avaliar se o conteúdo de redes sociais, frequentemente menos curado, introduz ruído ou benefício ao processo de classificação.}
        \end{enumerate}
\end{enumerate}

\novo{Para assegurar a comparabilidade entre as configurações, adotou-se uma divisão padronizada dos dados em conjuntos de treino (80\%), validação (10\%) e teste (10\%) para o COVID19.BR e o Fake.br. Essa padronização foi necessária porque os autores de COVID19.BR e Fake.br não disponibilizaram suas divisões originais, reportando apenas o uso de validação cruzada \textit{5-fold} \cite{martins2021covid19br_modelo, monteiro2018fakebr}. A re-divisão garante proporções consistentes para avaliação.}

\novo{Uma etapa adicional foi implementada para o \textit{corpus} Fake.br:  durante a divisão, garantiu-se que os pares de notícias (a \textit{fake news} e sua correspondente notícia verdadeira sobre o mesmo evento) fossem mantidos juntos na mesma partição (treino, validação ou teste). Esta medida é crucial para prevenir o vazamento de informação (\textit{data leakage}) entre os conjuntos e evitar uma avaliação excessivamente otimista do desempenho do modelo.}

\novo{O desempenho dos modelos em cada configuração de dados foi aferido por meio de duas abordagens distintas de aprendizado de máquina, descritas na Seção \ref{subsec:LM}. Estas abordagens foram escolhidas para representar diferentes paradigmas de treinamento e capacidades de modelos de linguagem:}
\begin{enumerate}
    \item \novo{Treinamento supervisionado completo (\textit{fine-tuning}) do modelo de linguagem Bertimbau (versão \textit{base}), um representante de Modelos de Linguagem Pequenos (SLMs) adaptados ao português.} 

    \item \novo{Aprendizado com poucos exemplos (\textit{few-shot learning}) utilizando o LLM Gemini 1.5 Flash.}
\end{enumerate}

\section{\correcao{Resumo dos métodos}}
\label{sec:methods_summary}

\correcao{Este capítulo detalhou a metodologia empregada para o enriquecimento de \textit{corpora} de detecção de notícias falsas em língua portuguesa, utilizando contexto externo proveniente de mecanismos de busca e Modelos de Linguagem Grandes (LLMs). O objetivo central foi avaliar o impacto tanto da melhoria da qualidade dos dados quanto da incorporação de evidências externas na tarefa de classificação de veracidade.}

\correcao{O processo iniciou-se com a seleção e validação de três conjuntos de dados distintos — \textbf{Fake.br}, \textbf{COVID19.BR} e \textbf{MuMiN-PT} —, escolhidos por sua diversidade de fontes (notícias web, WhatsApp, X/Twitter), domínios temáticos (geral, saúde), abordagens de coleta (\textit{top-down} e \textit{bottom-up}) e períodos temporais. Foi aplicado um rigoroso fluxo de validação semi-automático para garantir a qualidade dos dados, envolvendo filtragem inicial automatizada, filtragem de idiomas, resolução de rótulos conflitantes em pares relacionados, verificação externa de rótulos e tratamento específico para cada \textit{corpus}. Como medida final, URLs explícitas foram removidas dos textos para mitigar vieses de domínio.}

\correcao{O núcleo metodológico consiste em um \textbf{fluxo de enriquecimento adaptativo} inspirado no \textit{pipeline} canônico de Verificação Semi-Automática de Fatos (SAFC). A principal inovação reside na otimização da eficiência computacional: para textos que já funcionam como consultas de alta qualidade, procede-se diretamente à busca na \textit{web}; caso contrário, aciona-se uma etapa de \textbf{extração de alegação} via LLM Gemini 1.5 Flash para identificar a afirmação factual central. A correspondência entre consulta e resultados é avaliada através da presença de termos da consulta nos fragmentos destacados pelos marcadores HTML dos resultados do Google CSE. A recuperação de evidências foi realizada por meio de duas fontes complementares: a API \textbf{Google Custom Search Engine} para buscas gerais na \textit{web} e a API \textbf{Google FactCheck Claims Search} para alegações previamente verificadas por organizações de checagem de fatos.}

\correcao{Para a extração de alegações, foi desenvolvido um \textit{prompt} otimizado que solicita a identificação do principal fato em até 20 palavras, utilizando até os três primeiros parágrafos do texto original como contexto. O pré-processamento das consultas de busca empregou heurísticas específicas baseadas no comprimento do texto, privilegiando a primeira sentença para textos longos quando esta contém informação suficiente (7 ou mais palavras), ou utilizando o texto integral para textos curtos (até 20 palavras).}

\correcao{Por fim, foi delineada uma estratégia de avaliação experimental para mensurar o impacto incremental de cada etapa metodológica. Foram definidas configurações comparativas de dados: (1) dados originais como linha de base, (2) dados validados através do fluxo semi-automático, e (3) dados validados e enriquecidos com contexto externo, incluindo variações com e sem conteúdo de redes sociais. A avaliação foi conduzida nos \textit{corpora} COVID19.BR e Fake.br através de duas abordagens distintas de aprendizado de máquina: \textit{fine-tuning} supervisionado do modelo Bertimbau e \textit{few-shot learning} com Gemini 1.5 Flash, garantindo uma avaliação abrangente da eficácia das metodologias propostas sob diferentes paradigmas de modelagem.}

\correcao{Tendo detalhado a metodologia de validação e enriquecimento, o Capítulo \ref{c:develop} apresentará a aplicação prática deste fluxo. Serão exploradas as características dos dados originais, os resultados quantitativos e qualitativos do processo de enriquecimento e, por fim, a avaliação experimental do impacto dessas etapas.}
\chapter{Desenvolvimento}
\label{c:develop}

\novo{Este capítulo detalha o processo de desenvolvimento e apresenta os resultados obtidos a partir da aplicação da metodologia descrita no Capítulo \ref{c:methods}. Inicia-se com uma análise exploratória dos \textit{corpora} originais (Seção \ref{sec:eda}), destacando suas características e limitações intrínsecas. Em seguida, a Seção \ref{sec:expanded} descreve quantitativamente os dados enriquecidos com informações externas via LLMs e mecanismos de busca e a \novo{seção \ref{sec:qualitative_analysis} aprofunda a análise através de um estudo qualitativo dos padrões encontrados nos dados enriquecidos. Os resultados da avaliação dos dados, comparando diferentes configurações de processamento, são detalhados na Seção \ref{sec:qualitative_analysis}.} Por fim, a Seção \ref{sec:partial_results} oferece uma síntese e discussão integrada dos achados.}

\section{Análise Exploratória dos Dados}
\label{sec:eda}

\novo{Antes de proceder ao enriquecimento, realizou-se uma análise exploratória detalhada dos três conjuntos de dados selecionados (Fake.br, COVID19.BR, MuMiN-PT) após a etapa de pré-processamento e validação descrita na Seção \ref{subsec:validation}.}

\subsection{Balanceamento e Características Textuais}
\label{subsec:eda_main}

\begin{figure}[ht]
    \centering
    \includesvg[width=0.6\textwidth]{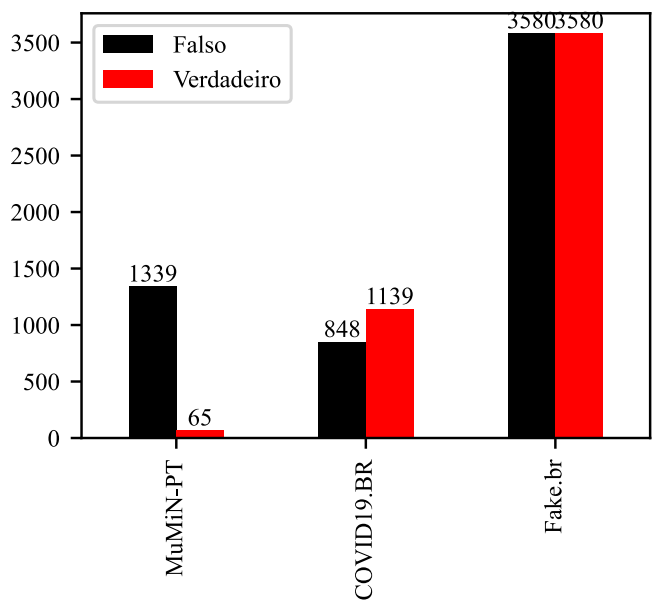}
    \caption{Balanceamento dos dados após pré-processamento.}
    \label{fig:balancing}
\end{figure}

A Figura \ref{fig:balancing} mostra o número de exemplos verdadeiros e falsos de cada conjunto de dados processado. Observa-se que o Fake.br apresenta um balanceamento razoável, enquanto MuMiN-PT e COVID19.BR exibem desbalanceamentos distintos: MuMiN-PT possui consideravelmente mais exemplos falsos, e COVID19.BR, mais exemplos verdadeiros. 

Uma hipótese para esses desbalanceamentos seria da natureza da obtenção dos dados. Como o MuMiN-PT é uma abordagem \textit{bottom-up}, ou seja, parte das agências que verificam fatos, é muito provável que as agências foquem em notícias falsas e, portanto, possuam mais exemplos falsos.
Isso pode ser reforçado pela Tabela \ref{tab:rotulos}, em que a contagem de rótulos utilizada pelas entidades jornalísticas era, em sua maioria, falsos.

\novo{Já o COVID19.BR, embora coletado via abordagem \textit{top-down} em um contexto de alta circulação de mensagens sobre a pandemia via WhatsApp, resultou em uma maior quantidade de exemplos verdadeiros. Isso sugere que a coleta capturou não apenas \textit{fake news}, mas também um volume significativo de mensagens informativas, alertas oficiais ou discussões factuais que foram classificadas como verdadeiras, ou que o processo de rotulação original teve um critério específico que levou a este resultado.}

\begin{table}[ht]
    \centering
\begin{tabularx}{\textwidth}{Xrrrrrrrrr}
\toprule
\small
 \multirow{2}{*}{Estatística} & \multicolumn{2}{c}{\textbf{MuMiN-PT}} & & \multicolumn{2}{c}{\textbf{Fake.br}} & & \multicolumn{2}{c}{\textbf{COVID19.BR}} \\
 \cmidrule{2-3} \cmidrule{5-6} \cmidrule{8-9}
 & \texttt{fake} & \texttt{true} && \texttt{fake} & \texttt{true} && \texttt{fake} & \texttt{true} \\
\midrule
{Média núm. de palavras} & 18,9 & 16,3 && 181,4 & 183,1 && 167,7 & 111,1 \\
\makecell[l]{Média. tam. das palavras\\(em caracteres)} & 5,0 & 4,9 && 4,8 & 5,0 && 4,9 & 6,6 \\
{Média de núm. de frases} & 1,4 & 1,4 && 10,4 & 9,0 && 10,9 & 5,8 \\
{Média núm de palavras por frase} & 14,5 & 12,3 && 18,6 & 22,1 && 19,2 & 22,9 \\
\makecell[l]{Presença de URLs}  & 0,3\% & 0,0\% && 1,0\% & 0,7\% && 28,9\% & 56,9\% \\
\bottomrule
\end{tabularx}
    \caption{Estatísticas textuais por conjunto de dados e rótulo.}
    \label{tab:dataset_words}
\end{table}

A Tabela \ref{tab:dataset_words} resume as estatísticas textuais. O tamanho médio dos textos varia significativamente, refletindo a plataforma de origem: \textit{tweets} (MuMiN-PT) são os mais curtos, seguidos por mensagens de WhatsApp (COVID19.BR) e notícias web (Fake.br). \novo{O tamanho médio das palavras, contudo, mostrou-se notavelmente estável entre os conjuntos e rótulos, pairando em torno de 4,8-5,0 caracteres.}

\novo{Em termos de rótulos, o MuMiN-PT não varia as estatísticas das palavras e frases entre os rótulos verdadeiro e falso, possivelmente por conta do limite baixo de caracteres imposto no X. Já no Fake.br, a notícia verdadeira, vinda de sites confiáveis, tende originalmente a ser maior. Para garantir uma classificação justa, o artigo original do Fake.br normaliza o tamanho dos textos em número de palavras, truncando os textos maiores em relação aos menores \cite{monteiro2018fakebr}. Conforme explicado na Seção \ref{sec:data}, a versão normalizada é utilizada neste trabalho.}

\novo{No COVID19.BR, por sua vez, tem a tendência inversa do Fake.br, uma hipótese seria por conta dos "textões" do WhatsApp comuns em notícias falsas, além do típico de mensagens menores do WhatsApp. Os autores do conjunto de dados mencionam que os exemplos \textit{fake} tendem a possuir tamanho maior e mais variável, quanto aos diferentes estilos de escrita \cite{martins2021covid19br}.}

O COVID19.BR é o conjunto de dados que mais possui links nos textos, tanto de forma absoluta \novo{em 889 exemplos ou relativa (44,7\%)}. Uma possível explicação seria que no ambiente do WhatsApp se compartilhem mais links em textos do que sites e \textit{tweets}. Tanto no COVID19.BR quanto no Fake.br as notícias verdadeiras possuem mais links associados. O MuMiN-PT não possui praticamente links associados nos exemplos, o que é esperado para \textit{tweets}.

\correcao{
\begin{table}[ht]
    \centering
    \begin{tabular}{lrrr}
    \toprule
    Domínios de URLs mencionadas   &  \texttt{fake} (\%)  & \texttt{true} (\%) & Total \\
    \midrule
    \textbf{\texttt{gazetabrasil}} & 0 (0,0\%) & 259 (\textbf{100,0\%}) & 259 \\
    \textbf{\texttt{bit.ly}} & 6 (6,8\%) & 82 (\textbf{93,2\%}) & 88 \\
    \texttt{youtube} & 47 (67,1\%) & 23 (32,9\%) & 70 \\
    \textbf{\texttt{globo}} & 6 (11,3\%) & 47 (\textbf{88,7\%}) & 53 \\
    \texttt{facebook} & 19 (38,0\%) & 31 (62,0\%) & 50 \\
    \texttt{dunapress} & 0 (0,0\%) & 46 (\textbf{100,0\%}) & 46 \\
    \texttt{twitter} & 25 (56,8\%) & 19 (43,2\%) & 44 \\
    \textbf{\texttt{whatsapp}} & 1 (3,1\%) & 31 (\textbf{96,9\%}) & 32 \\
    \texttt{uol} & 11 (37,9\%) & 18 (62,1\%) & 29 \\
    \texttt{conexaopolitica} & 16 (59,3\%) & 11 (40,7\%) & 27 \\
    \texttt{gov.br}\tnote{*}  & 6 (26,1\%) & 17 (73,9\%) & 23 \\
    \texttt{instagram} & 5 (21,7\%) & 18 (78,3\%) & 23 \\
    \texttt{jornaldacidadeonline} & 14 (63,6\%) & 8 (36,4\%) & 22 \\
    \textbf{\texttt{japinaweb}} & 0 (0,0\%) & 21 (\textbf{100,0\%}) & 21 \\
    \texttt{atrombetanews} & 7 (35,0\%) & 13 (65,0\%) & 20 \\
    \bottomrule
    \end{tabular}
         \caption{\correcao{Distribuição dos 15 domínios de URL mais frequentes no \textit{corpus} COVID19.BR por rótulo. A tabela apresenta a contagem absoluta de menções para cada rótulo (\texttt{fake}/\texttt{true}) e o total por domínio. As porcentagens indicam a proporção de cada rótulo dentro do total de menções daquele domínio. Domínios em negrito são aqueles em que mais de 80\% das menções ocorrem em textos com rótulo \texttt{true}.}}
    \label{tab:domain_covid}
\end{table}
}

\novo{No contexto de \textit{links} no conjunto de dados COVID19.BR, a Tabela \ref{tab:domain_covid} mostra os domínios de URLs mais mencionados nos exemplos. Os domínios da Gazeta Brasil, do bit.ly, da Globo, do WhatsApp e o JapinaWeb representam quase sempre exemplos verdadeiros. No entanto, a presença de um domínio governamental não é, por si só, garantia de veracidade. Domínios do governo brasileiro, como \texttt{gov.br}, podem ser instrumentalizados em contextos enganosos, como ilustrado por um exemplo do \textit{corpus} MuMiN-PT na Figura \ref{fig:example}.}

\begin{figure}[ht]
    \centering
\begin{tcolorbox}[colback=black!5, boxrule=0.5pt, arc=1mm, title=Exemplo de Uso Enganoso de URL Oficial (MuMiN-PT)]
\small
Pessoal, todo mundo precisa se cadastrar no conectesus para vacinar. Sugiro fazer já. Provavelmente o site nao aguentará os acessos quando for o momento. \url{https://conectesus-paciente.saude.gov.br/} É um cadastro no SUS. Quem tomou a vacina da Febre Amarela em 2018 já tem. Ou quem usou o SUS nos últimos anos. O aplicativo funciona mais ou menos como esses apps de carteira de motorista ou título de eleitor.
\end{tcolorbox}
    \caption[]{Exemplo do MuMiN-PT onde uma URL oficial é usada em contexto de \textit{fake news}. \footnotemark}
    \label{fig:example}
\end{figure}

\footnotetext{\url{https://lupa.uol.com.br/jornalismo/2021/01/12/verificamos-cadastro-sus}}

\subsection{Tópicos Predominantes e Temporalidade}
\label{subsec:eda_topics}

\begin{figure}[ht]
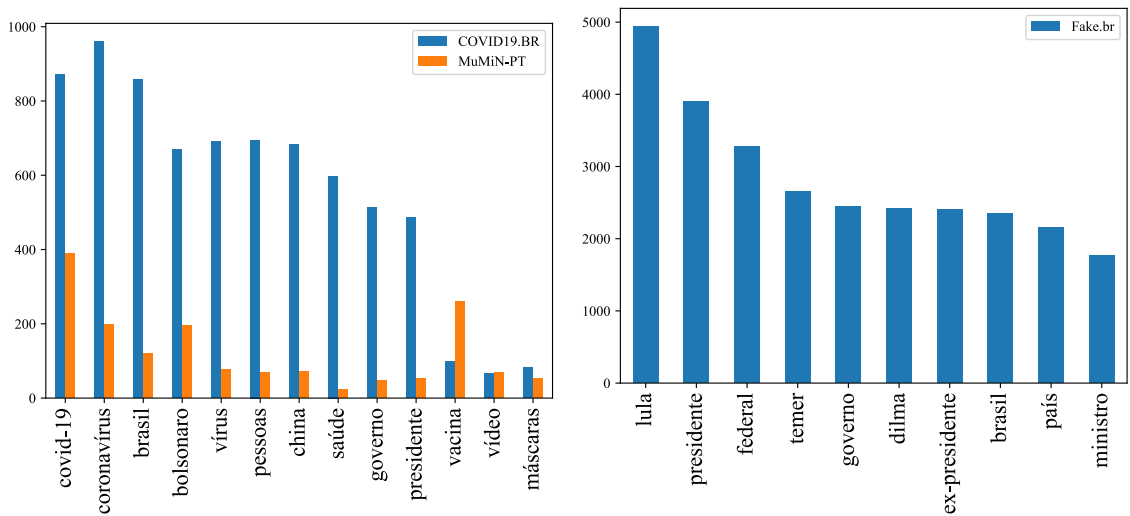

\centering
\includesvg[width=.496\linewidth, keepaspectratio]{figures/covid_topic.svg}
\hfill
\includesvg[width=.496\linewidth, keepaspectratio]{figures/fakebr_topic.svg}

\caption{10 Termos mais comuns (sem \textit{stopwords}) nos \textit{corpora} COVID19.BR/MuMiN-PT (esquerda) e Fake.br (direita).}
\label{fig:topics}
\end{figure}

A análise dos termos mais frequentes (após remoção de \textit{stopwords}), ilustrada na Figura \ref{fig:topics}, revela a forte dependência temporal dos conjuntos de dados. O Fake.br (coletado entre 2016-2018) reflete o cenário político pré-pandêmico brasileiro, com menções frequentes ao então presidente Temer.

Por outro lado, COVID19.BR e MuMiN-PT, coletados durante 2020-2022, compartilham tópicos centrados na pandemia de COVID-19, como "covid-19", "vacina", "máscaras") e no cenário político da época (menções ao então presidente Bolsonaro). \novo{Embora o MuMiN-PT se declare de domínio geral, a predominância de tópicos relacionados à COVID-19 sugere que eventos de grande impacto global dominam a conversação online, mesmo em conjuntos de dados que não são explicitamente focados neles.}

Analisando os termos mais mencionados nos gráficos da Figura \ref{fig:topics}, percebe-se que os conjuntos de dados de detecção de notícias falsas são altamente dependentes do cenário temporal. Isso é característico em tarefas de análise de sentimentos de redes sociais digitais, em que os termos em alta são voláteis, o que gera um vocabulário mais dinâmico e temporal.

\begin{figure}[ht]
    \centering
    \includesvg{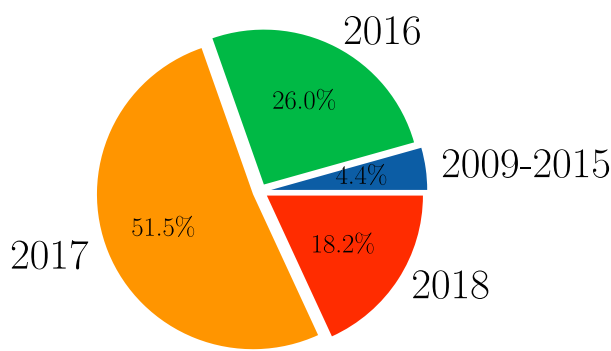}
    \caption{Tempo de publicação dos dados do Fake.br}
    \label{fig:fakebr_time}
\end{figure}

O Fake.br é o único que possui data associada a cada exemplo. A distribuição da data dos exemplos é mostrada na Figura \ref{fig:fakebr_time}. A metade dos dados é de 2017, um quarto é de 2016, um pouco menos de um quinto de 2018 e menos de 5,0\% é entre 2009 a 2015. 

\subsection{\novo{Análise de duplicatas}}
\label{sec:correpondence}

\novo{A presença de textos quase idênticos (quase duplicatas) foi investigada utilizando a técnica MinHash LSH, uma técnica em que trata a semelhança textual como um problema de interseção de caracteres ou palavras, e estima o tamanho relativo das interseções utilizando amostragem aleatória} \cite{broder_minhash_2000, har-peled_lsh_2012}.\novo{Foi utilizada a biblioteca Akin}\footnote{\url{https://github.com/justinbt1/Akin/tree/v0.1.0}} \novo{na versão 0.1.0. Os algoritmos de MinHash com LSH tiveram os seguintes parâmetros: semente aleatória} \texttt{seed=3}, \novo{tipo de n-grama como sendo carácter, tamanho do n-grama} \texttt{n\_gram = 5}, \novo{bits do HASH} \texttt{hash\_bits = 128}, \novo{número de bandas} \texttt{no\_of\_bands = 50} \novo{e distância mínima de Jaccard} \texttt{near\_duplicates} \novo{de 0,7.}

\begin{figure}[ht]
    \centering
    \includegraphics[width=0.45\textwidth,trim=20mm 20mm 20mm 15mm,clip]{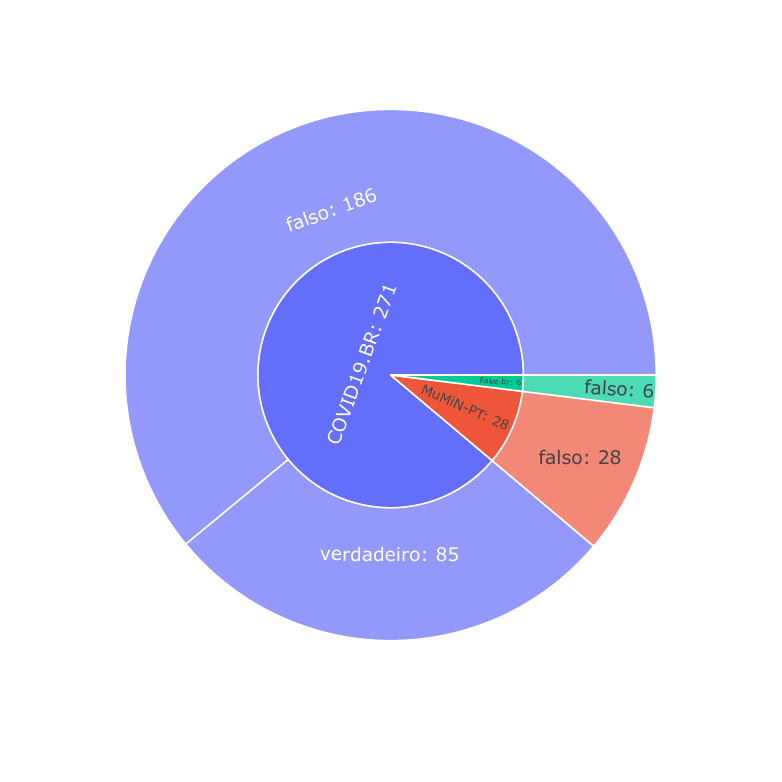}
    \caption{Contagem de quase duplicatas entre os \textit{corpora} e rótulos.}
    \label{fig:duplicates}
\end{figure}

\novo{A Figura \ref{fig:duplicates} mostra um gráfico de explosão solar dos exemplos que foram encontrados outros exemplos quase duplicatas do mesmo \textit{corpus}. O COVID19.BR possui mais exemplos quase duplicados, 271 exemplos (13,6\%). Uma hipótese seria o domínio do WhatsApp, em que as conversas curtas podem ser parecidas e redundantes e o compartilhamento de mensagens é alto e fácil. No Fake.br são 6 exemplos, o que corresponde a 0,08\% dos dados e o MuMiN-PT são 28 exemplos, correspondendo a 2,00\%.}

\novo{Analisando manualmente os exemplos, notou-se que a mudança entre quase duplicatas seria no contexto de: (1) espaçamento, (2) acréscimos de caracteres como traços "-" ou asteriscos "*" e (3) trocas ou acréscimos de poucas palavras. Na Figura \ref{fig:correspondence}, são destacados dois exemplos correspondentes do \textit{corpus} COVID19.BR, a diferença entre os dois textos é destacada em amarelo. À esquerda o texto possui duas quebras de linha a mais e à direita possui as palavras "– Conexão Política -".}

\begin{figure}[ht]
\begin{multicols}{2}
\centering
\begin{tcolorbox}[colback=black!5, boxrule=0.5pt, arc=1mm, title=Exemplo 1 (COVID19.BR)]
\footnotesize
China força países atingidos por vírus chinês a se ajoelharem diante da Huawei: “Nós lhe daremos máscaras se aceitar a Huawei 5G”\colorbox{yellow}{\phantom{..............................................................}}\\ \colorbox{yellow}{\phantom{.....................................................................}}\\
\url{https://conexaopolitica.com.br/ultimas/china-forca-paises-atingidos-por-virus-chines-a-se-ajoelharem-diante-da-huawei-nos-lhe-daremos-mascaras-se-aceitar-a-huawei-5g/}
\end{tcolorbox}
\columnbreak
\begin{tcolorbox}[colback=black!5, boxrule=0.5pt, arc=1mm, title=Exemplo 2 (COVID19.BR)]
\footnotesize
China força países atingidos por vírus chinês a se ajoelharem diante da Huawei: “Nós lhe daremos máscaras se aceitar a Huawei 5G” \hl{– Conexão Política -} \url{https://conexaopolitica.com.br/ultimas/china-forca-paises-atingidos-por-virus-chines-a-se-ajoelharem-diante-da-huawei-nos-lhe-daremos-mascaras-se-aceitar-a-huawei-5g/}
\end{tcolorbox}
\end{multicols}
\caption{\textit{Fake news} quase duplicatas no COVID19.BR (diferenças destacadas).}
\label{fig:correspondence}
\end{figure}

\begin{figure}
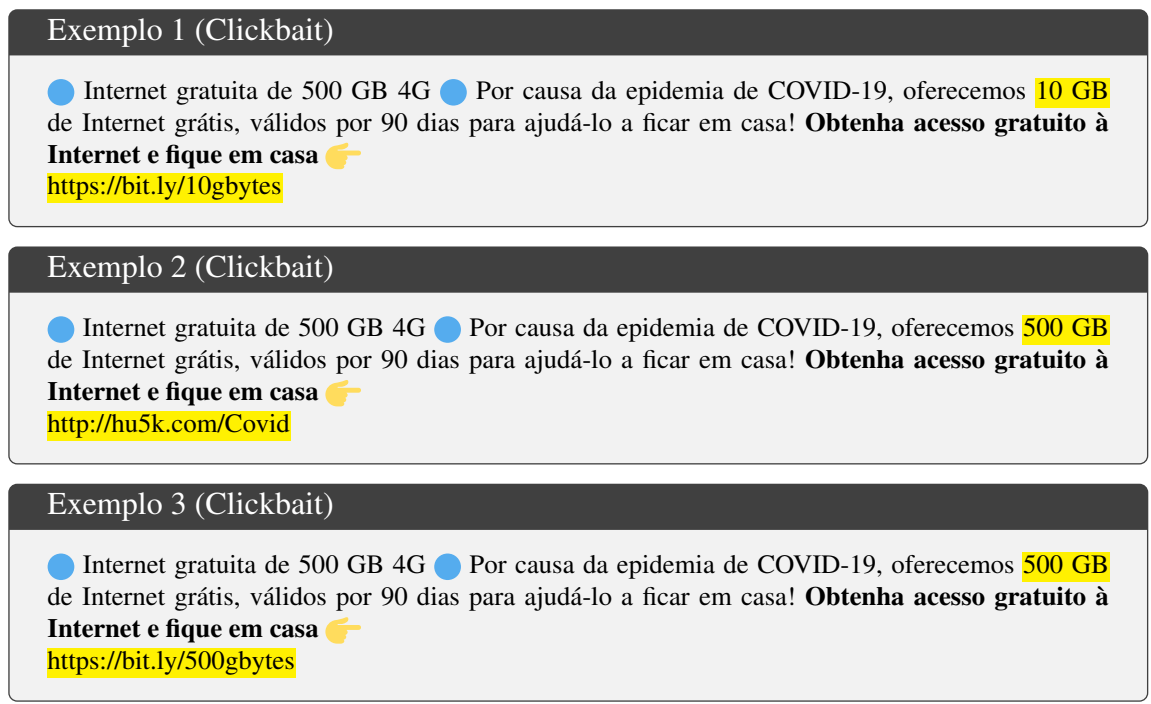

\centering
\begin{tcolorbox}[colback=black!5, boxrule=0.5pt, arc=1mm, title=Exemplo 1 (Clickbait)]
\footnotesize
\scalerel*{\includesvg{figures/blue-circle.svg}}{\textrm{\textbigcircle}} Internet gratuita de 500 GB 4G   \scalerel*{\includesvg{figures/blue-circle.svg}}{\textrm{\textbigcircle}} Por causa da epidemia de COVID-19, oferecemos \hl{10 GB} de Internet grátis, válidos por 90 dias para ajudá-lo a ficar em casa! \textbf{Obtenha acesso gratuito à Internet e fique em casa} \scalerel*{\includesvg{figures/pointing-right.svg}}{\textrm{\textbigcircle}}  \\ \hl{https://bit.ly/10gbytes}
\end{tcolorbox}

\begin{tcolorbox}[colback=black!5, boxrule=0.5pt, arc=1mm, title=Exemplo 2 (Clickbait)]
\footnotesize
\scalerel*{\includesvg{figures/blue-circle.svg}}{\textrm{\textbigcircle}} Internet gratuita de 500 GB 4G   \scalerel*{\includesvg{figures/blue-circle.svg}}{\textrm{\textbigcircle}} Por causa da epidemia de COVID-19, oferecemos \hl{500 GB} de Internet grátis, válidos por 90 dias para ajudá-lo a ficar em casa! \textbf{Obtenha acesso gratuito à Internet e fique em casa} \scalerel*{\includesvg{figures/pointing-right.svg}}{\textrm{\textbigcircle}}  \\ \hl{http://hu5k.com/Covid}
\end{tcolorbox}

\begin{tcolorbox}[colback=black!5, boxrule=0.5pt, arc=1mm, title=Exemplo 3 (Clickbait)]
\footnotesize
\scalerel*{\includesvg{figures/blue-circle.svg}}{\textrm{\textbigcircle}} Internet gratuita de 500 GB 4G   \scalerel*{\includesvg{figures/blue-circle.svg}}{\textrm{\textbigcircle}} Por causa da epidemia de COVID-19, oferecemos \hl{500 GB} de Internet grátis, válidos por 90 dias para ajudá-lo a ficar em casa! \textbf{Obtenha acesso gratuito à Internet e fique em casa} \scalerel*{\includesvg{figures/pointing-right.svg}}{\textrm{\textbigcircle}}  \\ \hl{https://bit.ly/500gbytes}
\end{tcolorbox}
\caption{Exemplos de  \textit{clickbait} quase duplicados no COVID19.BR (diferenças destacadas).}
\label{fig:clickbait}
\end{figure}

Na Figura \ref{fig:clickbait}, é um exemplo no COVID19.BR de \textit{clickbait}, sites usados para atrair cliques, termo definido na Seção \ref{subsec:taxonomy}. É divulgada uma internet gratuita 4G de gigas grátis, e varia de 10GB e 50 GB no tweet e o link. O texto é desmentido pela agência Boatos.org\footnote{\tiny \url{https://www.boatos.org/tecnologia/site-da-internet-gratuita-500-gb-4g-compartilhar-link-whatsapp.html}}.

\begin{figure}
    \centering
    \includesvg[width=0.7\textwidth]{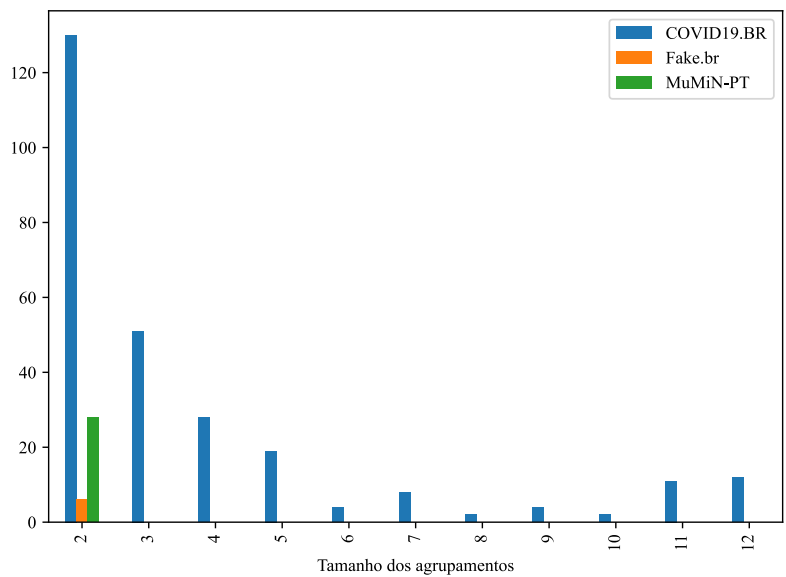}
    \caption{Histogramas de tamanho de agrupamentos}
    \label{fig:clusters}
\end{figure}

\novo{A análise dos tamanhos dos agrupamentos de duplicatas (Figura \ref{fig:clusters}) mostra que a maioria forma pares (tamanho 2), mas o COVID19.BR apresenta agrupamentos maiores (até tamanho 12), reforçando a ideia de replicação massiva de certas mensagens falsas nessa plataforma.}

\novo{A detecção e análise de quase duplicatas, incluindo os casos contraditórios removidos durante a validação (Seção \ref{sec:data}), foram fundamentais. Elas não apenas ajudaram a limpar os dados, mas também revelaram padrões de propagação de informação (atualizações vs. mutações virais) e destacaram a sensibilidade da classificação a pequenas variações textuais, um desafio importante para modelos de detecção.}

\subsection{Limitações Identificadas nos Dados Originais}
\label{subsec:limitations}
\noindent 
A análise exploratória expôs limitações inerentes aos conjuntos de dados, que motivam a busca por enriquecimento com informações externas:

\begin{itemize}
    \item \textbf{Temporalidade dos Tópicos e Rótulos:} Como visto (\ref{subsec:eda_topics}), os assuntos são voláteis. Além disso, a veracidade de uma alegação pode mudar com o tempo (e.g., obrigatoriedade de máscaras). Modelos estáticos enfrentam dificuldades com essa dinâmica.
    \item \textbf{Presença de Quase Duplicatas:} A redundância (\ref{sec:correpondence}) pode enviesar o treinamento e a avaliação. \novo{As pequenas variações entre duplicatas falsas também sugerem táticas de evasão de detecção.}
    \item \textbf{Granularidade dos Rótulos:} A classificação binária (verdadeiro/falso) simplifica a complexidade da \textit{fake news}, omitindo nuances como "enganoso" ou "fora de contexto" usadas por agências (ver Tabela \ref{tab:fact_checker_labels}).
    \item \textbf{Verificabilidade da Alegação:} Alguns textos contêm opiniões ou relatos pessoais não passíveis de verificação factual externa (e.g., "Hj faleceu um senhor com corona vírus aqui").
    \item \textbf{Dependência de URLs:} Muitos exemplos contêm URLs (especialmente no COVID19.BR). A análise do texto isolado ignora o conteúdo dessas páginas vinculadas, que pode ser crucial para a verificação. \novo{Um exemplo notório é o uso de um URL oficial do governo (Conecte SUS) para disseminar \textit{fake news}, como mostrado na Figura \ref{fig:example}.} \novo{Além disso, alguns domínios estão enviesados no conjunto de dados COVID19.BR como mostra a Tabela \ref{tab:domain_covid}, de forma que o domínio por si só poderia fornecer uma classificação de veracidade.}
\end{itemize}

\novo{Estas limitações, em conjunto, reforçam a necessidade de ir além do texto original, buscando evidências externas e contexto atualizado para uma avaliação mais robusta da veracidade, objetivo central do enriquecimento proposto.}

\pagebreak
\section{\novo{Ambiente de Avaliação}}
\label{sec:ambiente_avaliacao}

\novo{Nesta seção, detalha-se o ambiente computacional e as ferramentas empregadas na condução das avaliações experimentais, seguindo a metodologia delineada na Seção~\ref{sec:estrategia_avaliacao}. Os experimentos foram implementados em \texttt{Python}, com o auxílio da biblioteca \texttt{LiteLLM}\footnote{\url{https://github.com/BerriAI/litellm}}. Esta biblioteca foi utilizada para realizar as inferências com o modelo \texttt{Gemini 1.5 Flash}, facilitando a interação com a respectiva API.}

\novo{Para o \textit{few-shot learning}, todos os exemplos recebem os mesmos 15 exemplos aleatórios de entrada. A Figura~\ref{fig:prompt_used} apresenta o \textit{prompt} base utilizado para a inferência pelo modelo.}

\begin{figure}[ht] 
    \centering
    \begin{tcolorbox}[colback=black!1, boxrule=0.5pt, sharp corners]
        A seguir são apresentados textos de mensagens e notícias em português. Sua tarefa é classificar cada texto como contendo uma \textit{Fake News} ou como sendo VERDADEIRO.

        \ul{Para auxiliar na classificação, será também fornecido um contexto extra,} \ul{correspondente} \ul{a uma busca no Google pelos termos do texto a ser classificado.}

        Responda \textbf{apenas} com uma das seguintes \textit{tags}: \texttt{"FAKE NEWS"} ou \texttt{"VERDADEIRO"}.
    \end{tcolorbox}
    \caption{\textit{Prompt} base para a detecção de \textit{fake news}. A seção sublinhada é incluída quando o contexto extra (oriundo da busca externa) é considerado na análise.}
    \label{fig:prompt_used}
\end{figure}

\novo{Os experimentos de ajuste fino (\textit{fine-tuning}) foram executados em um ambiente com acesso a GPUs NVIDIA V100 (32 GB de VRAM) ou NVIDIA A100 (80 GB de VRAM). O \textit{framework} PyTorch foi utilizado, abstraído pela biblioteca SimpleTransformers \cite{rajapakse2024simpletransformers}. Cada execução experimental de ajuste fino demandou, tipicamente, entre 8 GB e 12 GB de memória VRAM da GPU, variando conforme a configuração de dados e o tamanho do lote (\textit{batch size}) utilizado.}

\novo{A Tabela \ref{tab:grid_search} apresenta o espaço de busca de hiperparâmetros configurado para o processo de ajuste fino do modelo Bertimbau base. Esta busca em grade (\textit{grid search}) foi aplicada a cada uma das configurações de dados de treinamento especificadas na Seção \ref{sec:estrategia_avaliacao}. A seleção da melhor combinação de hiperparâmetros para cada configuração de dados e \textit{corpus} foi realizada com base no maior valor de F1-Score obtido no conjunto de validação.}

\begin{table}[ht!]
\centering
\begin{tabular}{lr}
\hline
\textbf{Hiperparâmetros} & \textbf{Espaço de busca / Valor} \\
\hline
{Batch size} & \{8, 16\} \\
\multirow{2}{*}{Learning rate} & \multirow{2}{*}{\begin{tabular}[c]{@{}c@{}}\{1e-6, 5e-6, 1e-5,2.5e-5, \\  5e-5, 1e-4, 2.5e-5, 5e-5, 1e-4\}\end{tabular}} \\
\\
{Dropout of task layer} & \{0.1, 0.2, 0.3\} \\
& \\
{Seed} & 2025 \\
{Weight decay} & 0.01 \\
{Maximum training epochs} & 10 \\
{Maximum Sequence Length} & 512 \\
{Learning rate scheduler} & Linear com 6\% de \textit{warmup} \\
{Optimizer} & AdamW \\
AdamW $\epsilon$ & 1e-8 \\
AdamW $\beta_{1}$ & 0.9 \\
AdamW $\beta_{2}$ & 0.999 \\
{Early stopping patience} & 3 épocas \\ 
{Early stopping threshold (F1-score)} & 0.001  \\\hline
\end{tabular}
\caption{Espaço de busca de hiperparâmetros para o ajuste fino do modelo Bertimbau base.}
\label{tab:grid_search}
\end{table}

\novo{A definição do número máximo de épocas de treinamento em 10 baseou-se na observação de trabalhos anteriores com modelos BERT e RoBERTa em tarefas de classificação de texto com volumes de dados ou domínios análogos \cite{devlin2019bert, liu2019roberta, fake_bert, talendar2021fakebr_bertimbau, gusmao2023tcc, souza2020bertimbau}. O valor de \textit{weight decay} (0.01) adotado é consistente com as configurações padrão sugeridas para os modelos BERT e RoBERTa \cite{devlin2019bert, souza2020bertimbau, liu2019roberta}.}

\novo{O otimizador AdamW foi escolhido por ser o padrão para modelos da família BERT, utilizando-se os valores canônicos para AdamW $\beta_{1}$ (0.9) e AdamW $\beta_{2}$ (0.999) \cite{devlin2019bert, souza2020bertimbau, rajapakse2024simpletransformers}. O parâmetro AdamW $\epsilon$ (1e-8) corresponde ao valor padrão implementado na biblioteca SimpleTransformers \cite{rajapakse2024simpletransformers}. O comprimento máximo da sequência (\textit{Maximum Sequence Length}) foi definido em 512 \textit{tokens}, que é o limite padrão para modelos Bertimbau base e BERT base.}

\novo{A faixa de valores para a taxa de abandono (\textit{dropout}) na camada de tarefa (0.1, 0.2, 0.3) foi selecionada considerando-se que, em cenários com volumes de dados limitados, taxas de \textit{dropout} ligeiramente mais elevadas podem contribuir para a regularização do modelo e mitigar o sobreajuste (\textit{overfitting}) \cite{el2021dropout, griesshaber2020low_resource}. Os tamanhos de lote (\textit{batch sizes}) de 8 e 16 foram explorados, alinhando-se com as práticas de trabalhos que aplicaram modelos BERT, RoBERTa e Bertimbau base em \textit{corpora} de dimensão similar ou para detecção de notícias falsas \cite{devlin2019bert, liu2019roberta, souza2020bertimbau, fake_bert, talendar2021fakebr_bertimbau, gusmao2023tcc}.}

\novo{Por fim, o escalonador da taxa de aprendizado (\textit{learning rate scheduler}) com decaimento linear precedido por uma fase de aquecimento (\textit{warmup}) de 6\% das etapas de treinamento foi adotado em conformidade com as recomendações de \cite{liu2019roberta}. A estratégia de parada antecipada (\textit{early stopping}) foi configurada para monitorar o F1-Score no conjunto de validação, interrompendo o treinamento caso não houvesse melhoria superior a 0.001 por 3 épocas consecutivas, prevenindo o sobreajuste e otimizando o tempo de treinamento.}

\section{Conjunto de dados enriquecido}
\label{sec:expanded}
Esta seção apresenta os resultados da aplicação do fluxo de enriquecimento (Figura \ref{fig:process}) aos três \textit{corpora} pré-processados.

\begin{figure}[ht]
    \centering
    \includegraphics[width=\textwidth]{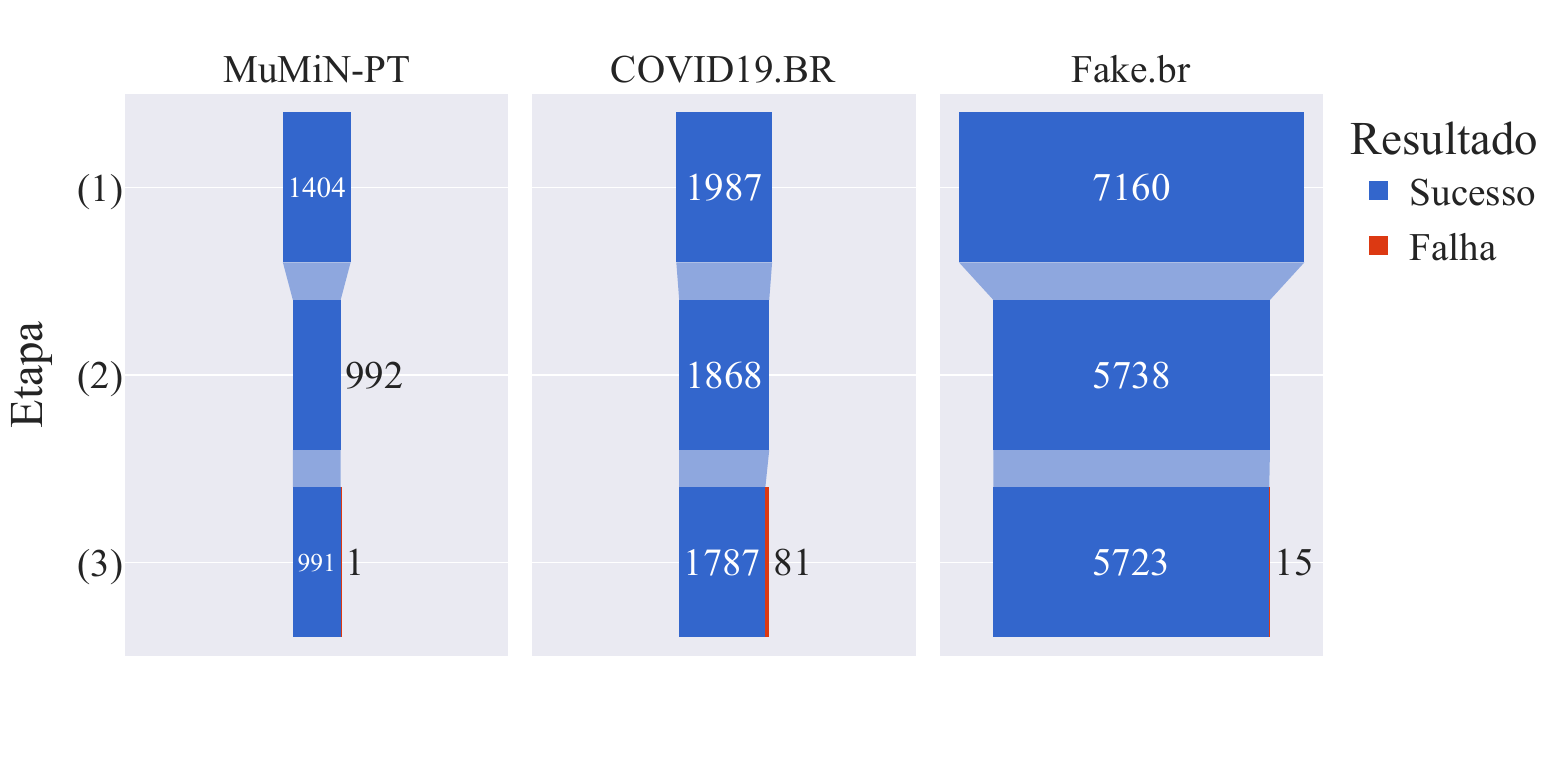}
    \caption{Funil do processo de enriquecimento para cada \textit{corpus}, mostrando o número de exemplos em cada etapa principal (Busca Inicial, Extração de Alegação, Busca pela Alegação).}
    \label{fig:funnel}
\end{figure}

A Figura \ref{fig:funnel} sumariza o fluxo de processamento e a quantidade de exemplos em cada etapa para os três \textit{corpora}. A etapa inicial envolve a busca na web pelo texto pré-processado (conforme Seção \ref{sec:search}). Se não houver correspondência forte (critério de 80\% de sobreposição de palavras, desconsiderando \textit{stopwords}), prossegue-se para a extração de alegação via LLM (Gemini 1.5 Flash, com o prompt da Figura \ref{fig:prompt_extraction}) e uma nova busca na web com a alegação extraída. \novo{Adicionalmente, realiza-se uma busca pela alegação na API Google FactCheck.}

A necessidade de extração de alegação variou consideravelmente: 94,0\% para COVID19.BR, 80,1\% para Fake.br e 70,7\% para MuMiN-PT. Uma hipótese para a menor necessidade no MuMiN-PT é pela sua característica \textit{bottom-up}, ou seja, a coleta do \textit{corpus} iniciou a partir de notícias verificadas por agências e, com isso, a correspondência da \textit{fake news} original é mais alta com os resultados e não precisa da extração de alegação.\novo{Em contrapartida, a alta taxa no COVID19.BR pode refletir a natureza mais informal e fragmentada das mensagens de WhatsApp, que exigiriam sumarização (extração de alegação) para uma busca focada.}

O processo foi robusto, com apenas 97 erros em mais de 10 mil exemplos processados, ou seja, a taxa de erro foi menor que 0,97\%. Todos os erros ocorreram na etapa de busca pela alegação extraída, onde a API não retornou resultados. \novo{Os dados enriquecidos resultantes são detalhados no Apêndice \ref{append:1}.}

\novo{Os resultados da análise qualitativa dos dados enriquecidos são apresentados na Seção \ref{sec:qualitative_analysis}, divididos entre os casos em que houve correspondência direta na busca inicial (Subseção \ref{subsec:match}) e aqueles que necessitaram de extração de alegação (Subseção \ref{subsec:no_match}).}

\subsection{Análise da Correspondência Direta na Busca Inicial}
\label{subsec:direct_match}

\begin{figure}[ht]
\centering
\includesvg[width=0.68\columnwidth]{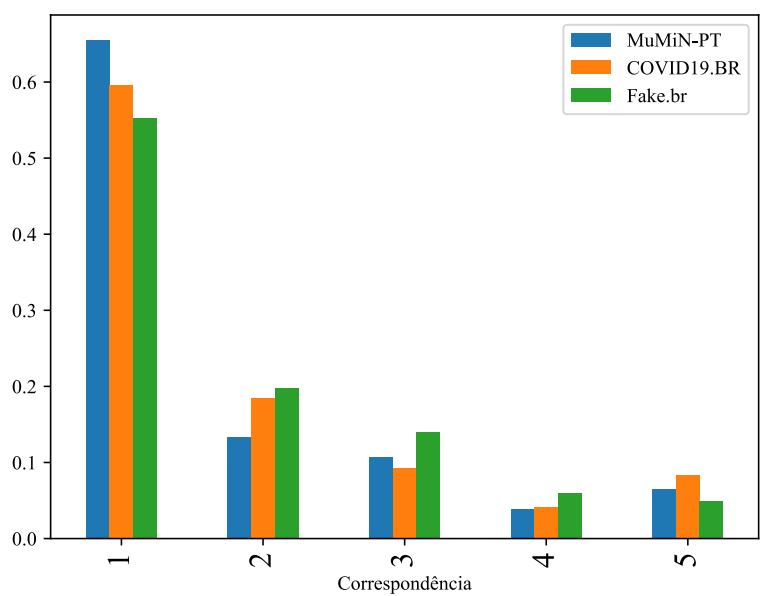}
\caption{Distribuição da posição do primeiro resultado de busca (Google CSE) com correspondência forte com a consulta inicial.}
\label{fig:match_idx}
\end{figure}

\novo{Quando a busca inicial na web (usando a API do CSE) retornava um trecho com alta similaridade lexical com a consulta (texto original pré-processado), a extração de alegação era dispensada. A Figura \ref{fig:match_idx} mostra em qual posição (do 1º ao 5º resultado) ocorreu a primeira correspondência forte.}

\novo{Observa-se que, na vasta maioria dos casos de correspondência (cerca de 60\%), ela ocorre já no primeiro resultado da busca. As taxas diminuem progressivamente para as posições seguintes. A distribuição é relativamente consistente entre os \textit{corpora}, com o MuMiN-PT apresentando uma taxa ligeiramente maior de correspondência no primeiro resultado.}  

\subsection{Análise da Extração de Alegações}
\label{subsec:claim_analysis}
Nos casos sem correspondência direta, procedeu-se à extração da alegação principal usando o LLM Gemini 1.5 Flash. Em média, as alegações extraídas continham entre 11 e 12 palavras, formando uma única frase, com tamanho médio de palavra em torno de 4.9 caracteres.

\begin{figure}[ht]
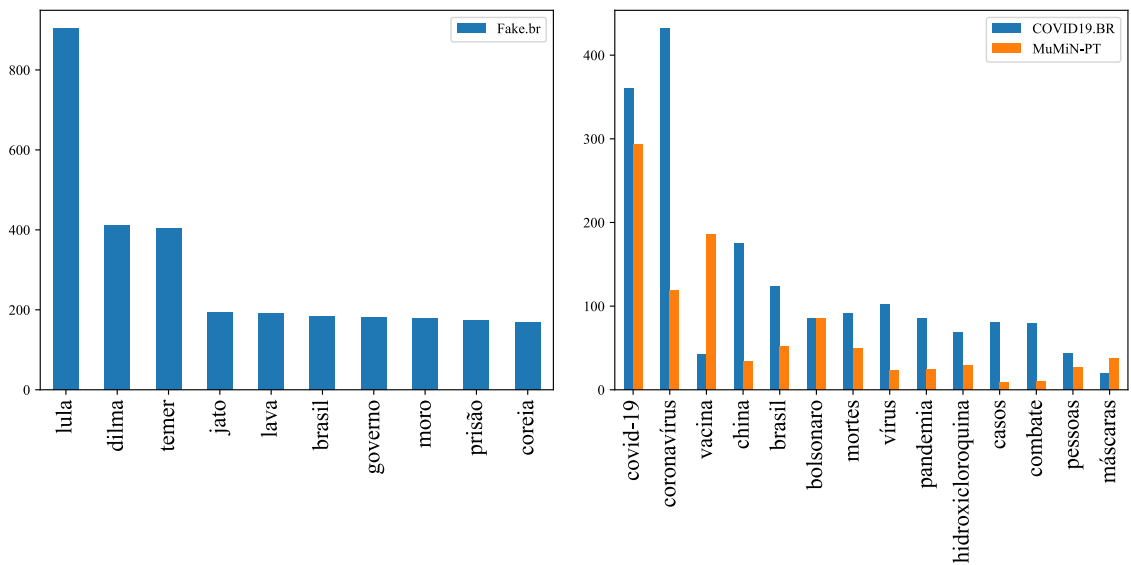

\centering
\begin{minipage}[t]{.496\textwidth}
    \vspace{-\topskip}
    \includesvg[width=\linewidth]{figures/fakebr_claim.svg}
\end{minipage}
\hfill
\begin{minipage}[t]{.496\textwidth}
    \vspace{-\topskip}
    \includesvg[width=\linewidth]{figures/covid_claim.svg}
\end{minipage}
\caption{10 Termos mais comuns (sem \textit{stopwords}) nas alegações extraídas para Fake.br (esquerda) e COVID19.BR/MuMiN-PT (direita).}
\label{fig:claim}
\end{figure}

A Figura \ref{fig:claim} apresenta os termos mais comuns nas alegações extraídas. Comparando com os termos mais comuns nos textos originais (Figura \ref{fig:topics}), nota-se uma manutenção geral dos tópicos, mas com algumas substituições.

No Fake.br, termos como "justiça" e "ex-presidente" deram lugar a "lava", "jato", "Moro" e "Dilma", refletindo um foco maior do LLM em entidades e eventos específicos (Operação Lava Jato, ex-presidenta Dilma Rousseff) ao sumarizar. No COVID19.BR/MuMiN-PT, "presidente", "saúde", "vídeo" foram parcialmente substituídos por "Bolsonaro", "hidroxicloroquina" e "casos", indicando uma sumarização focada em aspectos mais factuais ou controversos da pandemia.

Três hipóteses para mudança das palavras mais comuns das alegações podem ser a mudança temporal, o encurtamento do texto e o estilo de escrita. A passagem do tempo, por exemplo, no Fake.Br seria a palavra "Dilma" na alegação ser mais utilizada que ex-presidente no texto original, porque na coleta dos dados a Dilma era a última ex-presidente enquanto no treino do Gemini não é mais. O encurtamento do texto original para gerar a alegação pode ter contribuído para a remoção de algumas palavras mais frequentes e o estilo de escrita do Gemini pode ter contribuído para a troca de alguns termos equivalentes.

\subsection{Análise dos Resultados da Busca Web (Google CSE)}
\label{subsubsec:main_search}

\begin{figure}[ht]
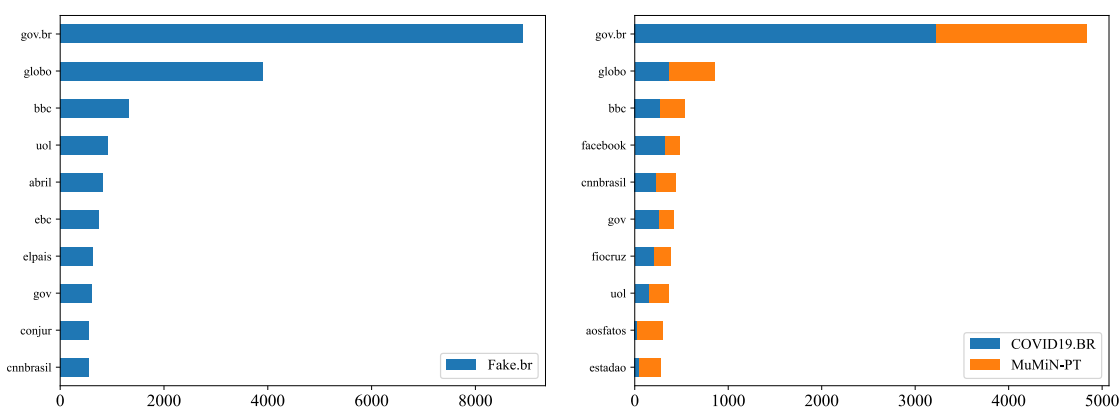

\centering
\includesvg[width=.496\linewidth, keepaspectratio]{figures/fakebr_domain.svg}
\hfill
\includesvg[width=.496\linewidth, keepaspectratio]{figures/covid_domain.svg}
\caption{10 Domínios mais frequentes nas URLs dos resultados da Google CSE para Fake.br (esquerda) e COVID19.BR/MuMiN-PT (direita).}
\label{fig:domain}
\end{figure}

Analisaram-se os domínios das URLs retornadas pela API do Google CSE, tanto na busca inicial quanto na busca pela alegação extraída. A Figura \ref{fig:domain} mostra os domínios mais frequentes para cada \textit{corpus}, os domínios governamentais brasileiros foram aglutinados em "gov.br" e os do governo americano em ".gov". 

Os principais domínios (gov.br, globo.com, bbc.com, uol.com.br) são consistentes entre os \textit{corpora}, indicando a proeminência de fontes governamentais e grandes veículos de mídia nos resultados de busca. Domínios como facebook.com aparecem com mais destaque para COVID19.BR e MuMiN-PT, refletindo a origem desses dados em redes sociais.

\novo{Similarmente, a presença de fiocruz.br e cnnbrasil.com.br nos resultados para os \textit{corpora} relacionados à pandemia (COVID19.BR, MuMiN-PT) e sua ausência no Fake.br (pré-pandemia e pré-CNN Brasil) demonstra a sensibilidade da busca ao contexto temporal e temático dos dados originais.}

Uma parcela dos resultados da CSE remetia a páginas indexadas pela ferramenta Google FactCheck, contendo verificações jornalísticas. \novo{Essa categoria foi especialmente relevante para o MuMiN-PT (21,3\% dos retornos), reforçando a conexão deste \textit{corpus} com alegações já verificadas, em comparação com COVID19.BR (1,0\%) e Fake.br (0,6\%).} A Tabela \ref{tab:agent_domain} lista os domínios dessas agências identificadas indiretamente via CSE.

\begin{table}[ht]
\centering
\begin{tabular}{lccc}
\toprule
\textbf{Domínio da Agência (via CSE)} & \textbf{COVID19.BR} & \textbf{Fake.br} & \textbf{MuMiN-PT} \\ \midrule
\texttt{afp.com}              & 8                  & 10                & 154               \\
\texttt{uol.com.br}           & 5                  & 11                & 95                \\ 
\texttt{observador.pt}        & 4                  & 4                 & 80                \\
\texttt{estadao.com.br}       & 4                  & 17                & 11                \\
\texttt{e-farsas.com}         & 0                  & 1                 & 8                 \\
\texttt{sbt.com.br}           & 3                  & 5                 & 0                 \\
\texttt{globo.com}            & 0                  & 0                 & 7                 \\ 
\texttt{projetocomprova.com.br} & 0                & 0               & 7                 \\
\texttt{sapo.pt}              & 0                  & 1                 & 0                 \\ \bottomrule
\end{tabular}
\caption{Contagem de domínios de agências de checagem identificados nos resultados da Google CSE que continham marcação `ClaimReview` (domínios principais aglutinados).}
\label{tab:agent_domain}
\end{table}

Os domínios mais frequentes foram AFP, UOL, Observador e Estadão. \novo{É importante notar que, embora essas páginas fossem indexadas como contendo verificações, a API da CSE por si só não fornece o rótulo de veracidade atribuído pela agência.}

\begin{figure}[ht]
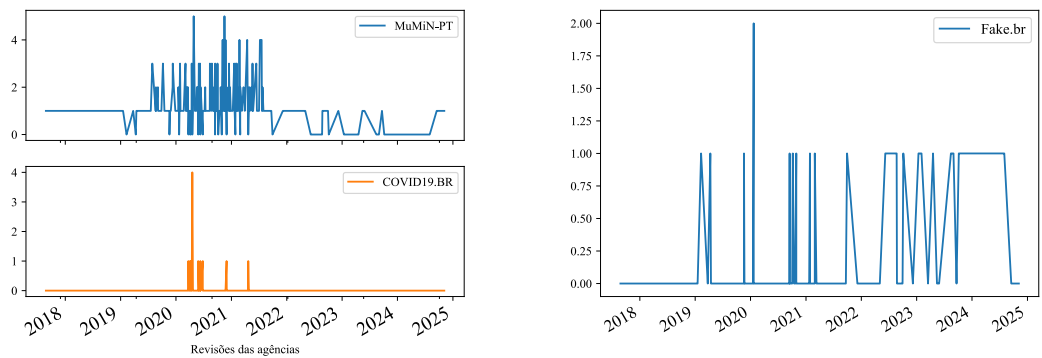

\centering
\includesvg[width=.496\linewidth, keepaspectratio]{figures/covid_dates.svg}
\hfill
\includesvg[width=.496\linewidth, keepaspectratio]{figures/fakebr_agency_date.svg}
\caption{Distribuição das datas de publicação das páginas de agências encontradas via Google CSE para COVID19.BR/MuMiN-PT (esquerda) e Fake.br (direita).}
\label{fig:agency_date}
\end{figure}

A análise temporal das datas de publicação dessas páginas de verificação (Figura \ref{fig:agency_date}) mostra alinhamento com os períodos de coleta dos dados originais para COVID19.BR (pico em 2020) e MuMiN-PT (2020-2022). \novo{Para o Fake.br, as datas dos resultados de verificação não mapeiam diretamente com a distribuição temporal dos dados originais (Figura \ref{fig:fakebr_time}), sugerindo que as buscas atuais recuperam verificações mais recentes ou que as verificações da época (2016-2018) têm menor visibilidade hoje. Isso reforça a sugestão de realizar buscas com restrição temporal para simular melhor a verificação em tempo real.}

\novo{Por fim, observou-se que, apesar dos parâmetros para restringir a busca ao português do Brasil (\texttt{gl=pt-BR}, \texttt{lr=lang\_pt}), um pequeno número de resultados (<0.05\%) ainda retornou páginas em outros idiomas (principalmente espanhol). Tentativas de restringir ainda mais pela localização do servidor (\texttt{cr=countryBR}) levaram à omissão de retornos relevantes, indicando um desafio na filtragem geográfica/linguística perfeita via API.}

\subsection{\novo{Análise dos Resultados da Busca de Alegações (Google FactCheck API)}}
\label{subsubsec:claim_search}

\novo{Paralelamente à busca geral (CSE), utilizou-se a API de busca de alegação do Google FactCheck, especificamente projetada para recuperar alegações verificadas. A taxa de sucesso (exemplos que retornaram pelo menos uma verificação) variou drasticamente: 58,8\% para MuMiN-PT, 4,8\% para COVID19.BR e apenas 0,7\% para Fake.br. A Tabela \ref{tab:agent_fact_checker_domain} mostra os domínios das agências cujas verificações foram recuperadas diretamente por esta API.}

\begin{table}[ht]
\centering
\begin{tabular}{lccc}
\toprule
\textbf{Domínio da Agência (via Fact Check API)} & \textbf{COVID19.BR} & \textbf{Fake.br} & \textbf{MuMiN-PT} \\
\midrule
\texttt{aosfatos.org} & 25 & 15 & 243 \\
\texttt{uol.com.br} & 27 & 24 & 219 \\ 
\texttt{observador.pt} & 4 & 3 & 140 \\
\texttt{boatos.org} & 20 & 2 & 77 \\
\texttt{afp.com} & 2 & 5 & 64 \\
\texttt{estadao.com.br} & 6 & 7 & 34 \\
\texttt{projetocomprova.com.br} & 3 & 0 & 42 \\
\texttt{globo.com} & 1 & 0 & 10 \\ 
\bottomrule
\end{tabular}
\caption{Contagem de domínios das agências fontes dos resultados da busca de alegações do Google FactCheck (domínios principais aglutinados).}
\label{tab:agent_fact_checker_domain}
\end{table}

\begin{figure}[ht]
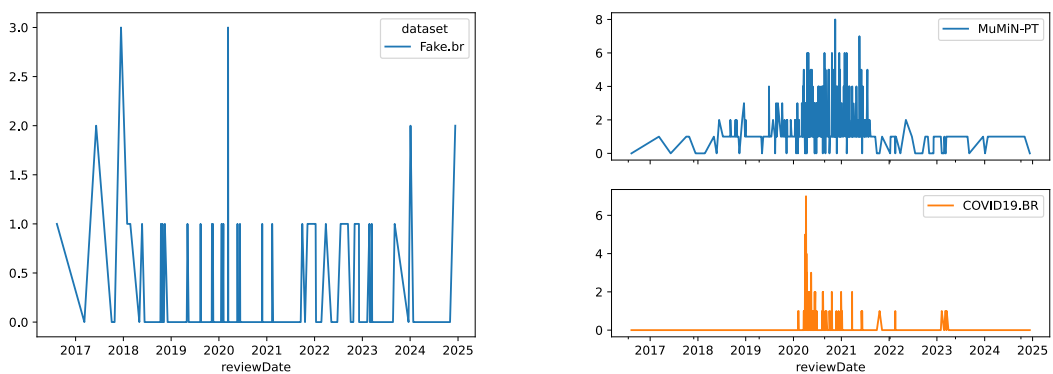

\centering
\includesvg[width=.496\linewidth, keepaspectratio]{figures/fakebr_agency_fact_checker_dates.svg}
\hfill
\includesvg[width=.496\linewidth, keepaspectratio]{figures/covid_fact_checker_dates.svg}
\caption{Distribuição das datas de publicação das verificações encontradas via Google FactCheck API para Fake.br (esquerda) e COVID19.BR (direita).}
\label{fig:fact_checker_date}
\end{figure}

\novo{Comparando com a Tabela \ref{tab:agent_domain} (agências via CSE), observa-se uma mudança na proeminência: 'Aos Fatos' emerge como a principal fonte direta na Fact Check API, enquanto 'AFP' era mais visível indiretamente na CSE. 'UOL' e 'Observador' mantêm-se relevantes em ambas. Notavelmente, 'Boatos.org', importante fonte para o COVID19.BR (Tabela \ref{tab:datasets}), aparece aqui, mas não estava entre os principais da CSE com marcação `ClaimReview`. A distribuição temporal das verificações recuperadas (Figura \ref{fig:fact_checker_date}) é similar à observada na Figura \ref{fig:agency_date}, alinhada aos períodos dos \textit{corpora}.}

\novo{A busca de alegações do Google FactCheck, como explicado na Figura \ref{fig:claim_search}, mostra o rótulo que a entidade jornalística associou a alegação. Dos 986 resultados, somente 11 foram catalogados como verdadeiros. Isso sustenta a hipótese de que as organizações verficam mais notícias como falsas do que verdadeiras.}

\begin{table}[ht]
    \centering
\begin{tabularx}{\textwidth}{lX}
\toprule
\textbf{Agências} & \textbf{Rótulos} \\
\midrule
Aos Fatos & \textbf{falso (246)}, distorcido (29), não é bem assim (2), insustentável (2), exagerado (2), contraditório (2) \\
UOL Notícias & \textbf{falso (130)}, \textbf{enganoso (10)}, insustentável (7), distorcido (2), sátira (1) \\
Observador & \textbf{errado (124)}, \textbf{enganador (23)} \\
Agência Lupa - UOL & \textbf{falso (100)}, verdadeiro (5), verdadeiro mas (5), exagerado (2), de olho (1), ainda é cedo para dizer (1) \\
Boatos.org & \textbf{falso (99)} \\
AFP Checamos & \textbf{falso (48)}, \textbf{enganoso (17)}, sem registro (2), verdadeiro (1), sem indícios (1), sem contexto (1), falta contexto (1) \\
Estadão Verifica & \textbf{enganoso (25)}, \textbf{falso (20)}, fora de contexto (2) \\
Projeto Comprova & \textbf{falso (24)}, \textbf{enganoso (21)} \\
G1: Fato ou Fake & \textbf{fake (11)} \\
BOL - UOL & \textbf{falso (2)} \\
Folha - UOL & \textbf{enganoso (1)}, \textbf{falso (1)} \\
Revista Piauí - UOL & \textbf{falso (2)} \\
\bottomrule
\end{tabularx}
    \caption{Distribuição dos rótulos de veracidade atribuídos pelas agências, conforme recuperados pela API do Google FactCheck (rótulos mais comuns destacados em negrito).}
    \label{tab:fact_checker_labels}
\end{table}

\novo{Os rótulos associados às agências estão apresentados na Tabela \ref{tab:fact_checker_labels}. Os mais comuns são "falso/fake/errado" e "enganoso/enganador", destacados em negrito. A diversidade de rótulos além do binário simples (e.g., "distorcido", "sem contexto", "exagerado") também ilustra a granularidade perdida nos conjuntos de dados originais e parcialmente recuperada através desta API.}

\novo{Esses resultados podem ser comparados com os da Tabela \ref{tab:rotulos} de \cite{couto2021central_de_fatos}, com a ressalva de que o trabalho mencionado extraiu aproximadamente quatro vezes mais alegações, o que aumenta as chances de capturar categorias minoritárias. Além disso, na pesquisa de \cite{couto2021central_de_fatos}, a extração foi obtidas diretamente dos sites via \textit{crawler}, enquanto neste trabalho foram extraídas da API de verificação de fatos do Google.}

\novo{Embora os rótulos estejam distribuídos de forma semelhante, eles aparecem em ordens diferentes. Alguns termos não foram encontrados neste estudo, incluindo: "verdadeiro" e "impreciso" da agência Aos Fatos; "contraditório" e "subestimado" da Agência Lupa; "fato" e "não é bem assim" do G1: Fato ou Fake; e "Fato" do Estadão Verifica. Além disso, observou-se uma mudança na categorização atribuída pelo Boatos.org, que passou a ser "falso".}

\section{\novo{Análise Qualitativa dos Padrões nos Dados Enriquecidos}}
\label{sec:qualitative_analysis}
\novo{Para complementar a análise quantitativa, realizou-se uma análise qualitativa focada em identificar padrões recorrentes nos resultados do enriquecimento, especialmente em relação à natureza das evidências encontradas. Essa análise foi organizada considerando os dois cenários principais do fluxo da Figura \ref{fig:process}: busca com correspondência direta e busca após extração de alegação.}

\subsection{\novo{Padrões em Casos de Correspondência Direta}}
\label{subsec:match}
\novo{Quando a busca inicial (CSE) encontrava alta similaridade com o texto original, dispensando a extração de alegação, observaram-se padrões distintos dependendo da veracidade do texto original (verdadeiro/falso) das alegações analisadas:}

\noindent \textbf{\novo{Para alegações originalmente rotuladas como VERDADEIRAS:}}
\begin{itemize}[noitemsep, topsep=0pt]
\item[\textbf{V1:}] \textbf{Corroboração:} \novo{Os resultados de busca frequentemente contêm links para a fonte original da notícia ou outros veículos confiáveis que reportam a mesma informação factual. Este padrão fornece evidências que confirmam a veracidade da alegação por meio de corroboração independente, oferecendo suporte verificável para a informação apresentada.}
\item[\textbf{V2:}] \textbf{Ausência de Confirmação Explícita:} \novo{Raramente se encontraram resultados de agências de checagem de fatos que confirmassem explicitamente uma alegação verdadeira por meio da CSE ou da API Fact Check. Esta observação alinha-se com a tendência observada na Tabela \ref{tab:fact_checker_labels}, na qual as agências focam primariamente em desmentir informações falsas ou enganosas em vez de afirmar declarações verdadeiras, sugerindo um viés sistemático no ecossistema de verificação de fatos.}
\end{itemize}
\noindent \textbf{\novo{Para alegações originalmente rotuladas como FALSAS:}}
\begin{itemize}[noitemsep, topsep=0pt]
\item[\textbf{F1:}] \textbf{Refutação Direta:} \novo{O resultado ideal para fins de verificação, incluem artigos ou checagens de fatos de fontes confiáveis (frequentemente agências de checagem) que refutam diretamente a alegação, fornecendo evidências contrárias claras e específicas que desmentem a informação falsa.}
\item[\textbf{F2:}] \textbf{Reforço da \textit{fake news}:} \novo{Um resultado problemático para a verificação automatizada, apresentam outras instâncias da mesma \textit{fake news} sendo compartilhada em redes sociais, blogs ou sites não confiáveis. Este padrão inadvertidamente amplifica a alegação falsa em vez de corrigi-la, criando um desafio significativo para sistemas que dependem apenas da correspondência textual sem avaliação da credibilidade da fonte.}
\item[\textbf{F3:}] \textbf{Reconhecimento Acadêmico como \textit{fake news}:} \novo{Um padrão distinto no qual os resultados de busca apontam para artigos acadêmicos, teses ou publicações científicas que discutem a \textit{fake news} específica como um exemplo em seu contexto de pesquisa. Isso serve como meta-evidência, indicando que a alegação é reconhecida como falsa ou enganosa pela comunidade de pesquisa, proporcionando um tipo indireto mas valioso de verificação.}
\end{itemize}

\novo{Para alegações verdadeiras, os resultados de busca tipicamente correspondem à própria notícia original (V1), como exemplificado pela mensagem de WhatsApp apresentada na Figura \ref{fig:true} sobre uma \textit{startup} desenvolvendo teste de COVID-19. A ausência significativa do padrão V2 (verificação explícita de notícia verdadeira por agência) reforça a conclusão da análise quantitativa (Tabela \ref{tab:fact_checker_labels}) de que o foco primário das agências de verificação é identificar e desmentir a \textit{fake news}, não confirmar informações verdadeiras.}

\begin{figure}[ht]
    \centering
\begin{tcolorbox}[colback=black!5, boxrule=0.5pt, arc=1mm, title=Exemplo de Alegação Verdadeira (COVID19.BR - Padrão V1)]
\small
Projeto foi um dos seis primeiros selecionados em edital lançado pelo Pipe-Fapesp, para apoiar pesquisas sobre inovações. Pesquisadores da Biolinker – uma startup de biotecnologia (biotech) incubada no Centro de Inovação, Empreendedorismo e Tecnologia (Cietec) – estão desenvolvendo, com apoio do Programa Fapesp Pesquisa Inovativa em Pequenas Empresas (Pipe), um teste de diagnóstico da COVID-19 (doença causada pelo novo coronavírus) de baixo custo e alto desempenho, com insumos totalmente nacionais.
\url{https://dunapress.org/2020/06/11/covid-19-startup-busca-desenvolver-teste-de-diagnostico-totalmente-nacional/}
\end{tcolorbox}
    \caption{\novo{Exemplo de texto verdadeiro do COVID19.BR. Buscas por trechos relevantes retornam a notícia original em fontes confiáveis como a Agência FAPESP e outros veículos de comunicação que reportaram o mesmo fato. A presença de múltiplas fontes independentes reportando a mesma informação serve como forte evidência de veracidade (Padrão V1).}}
    \label{fig:true}
\end{figure}

\begin{figure}[ht]
    \centering
    \includegraphics[width=0.75\linewidth]{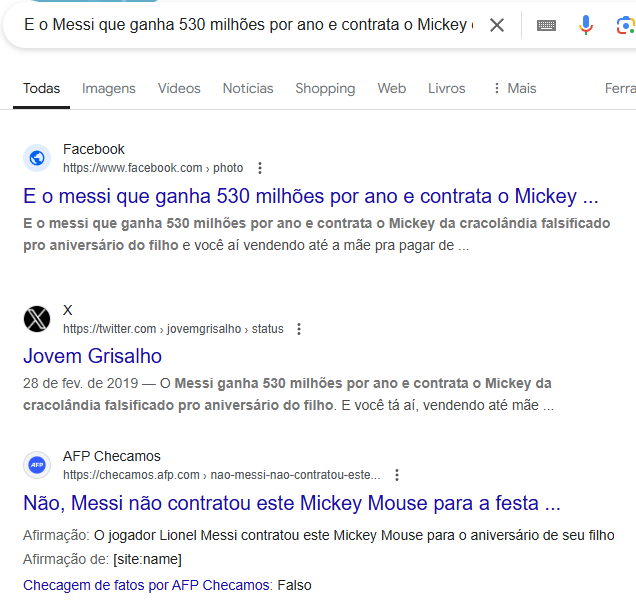}
    \caption{\novo{Exemplo de retorno de busca (Padrão F2) onde o primeiro link aponta para a própria \textit{fake news} sendo compartilhada no Facebook. Este tipo de resultado representa um desafio significativo para sistemas automatizados de verificação, pois a alta correspondência textual pode ser interpretada erroneamente como validação da alegação, quando na verdade representa apenas mais uma instância da mesma \textit{fake news} sendo propagada.}}
    \label{fig:fake_match}
\end{figure}

\novo{Para alegações falsas (\textit{fake news}), os três padrões identificados (F1, F2, F3) foram observados com frequências variáveis. O cenário ideal (F1), onde a busca retorna diretamente uma verificação de fonte confiável desmentindo a alegação, ocorreu em parte dos casos, mas não foi o padrão predominante. Frequentemente, o padrão F2 prevaleceu: a busca retornava a própria \textit{fake news} sendo propagada em outras plataformas, como posts de Facebook (Figura \ref{fig:fake_match}) ou blogs não confiáveis, sem qualquer sinalização de que se tratava de conteúdo falso.}

\novo{O padrão F2 é particularmente problemático para sistemas automáticos de verificação, pois uma correspondência forte com um resultado que replica a \textit{fake news} pode ser interpretada erroneamente como validação se a credibilidade da fonte não for rigorosamente avaliada. Esta observação ressalta a importância crítica de incorporar avaliação de credibilidade de fontes em qualquer sistema automatizado de verificação de fatos, além da mera correspondência textual.}

\novo{O padrão F3, onde a \textit{fake news} é citada em publicações acadêmicas ou análises, também foi identificado com frequência significativa. Conforme detalhado na Tabela \ref{tab:articles}, foram encontrados 37 resultados de busca apontando para 23 publicações acadêmicas distintas (artigos, dissertações, TCCs) que utilizavam exemplos dos \textit{corpora} analisados (majoritariamente do Fake.br e COVID19.BR) em suas análises sobre \textit{fake news}. Este padrão representa uma forma interessante e não antecipada de verificação indireta, onde a menção da alegação em trabalhos acadêmicos que a classificam como \textit{fake news} serve como meta-evidência de sua falsidade.}

\begin{table}[htbp]
\centering
\scriptsize 
\resizebox{\textwidth}{!}{%
\begin{tabular}{@{}lcllll@{}}
\toprule
Referência & Área & Ano & Instituição & Tipo de Publicação & Tema Principal \\ \midrule
\cite{ribeiro2018sociais} & C. Sociais & 2018 & USP & Periódico & Política \\
\midrule
\cite{melo2019dissertacao} & Comunicação & 2019 & PUC-RIO & Dissertação Mestrado & Política \\
\cite{sakurai2019tcc} & Computação & 2019 & UEL & TCC & Geral \\
\cite{ituassu2019comunicacao} & Comunicação & 2019 & PUC-RIO & Congresso & Política \\
\midrule
\cite{santos2020si} & Computação & 2020 & Unichristus & TCC & Geral \\
\midrule
\cite{melo2021livro} & Comunicação & 2021 & PUC-RIO & Livro & Política \\
\cite{barbieri2021mestrado} & Comunicação & 2021 & UTP & Dissertação Mestrado & COVID-19 \\
\cite{quintanilha2021mestrado} & Linguística & 2021 & UNL & Dissertação Mestrado & COVID-19 \\
\cite{lima2021comunicacao} & Comunicação & 2021 & UFOP & Periódico & COVID-19 \\
\cite{lima2021tcc} & Comunicação & 2021 & USP & TCC & Política \\
\cite{arndt2021psicologia} & Política & 2021 & UFSC & Periódico & COVID-19 \\
\cite{leurquin2021linguistica} & Linguística & 2021 & UFC & Periódico & COVID-19, Política \\
\cite{sa2021master} & Computação & 2021 & UFC & Dissertação Mestrado & COVID-19, Política \\
\cite{nascimento2021tcc} & Biblioteconomia & 2021 & UFC & TCC & COVID-19 \\
\cite{capistrano2021tcc} & Química & 2021 & UFPA & TCC & COVID-19 \\
\cite{batista2021adm} & Administração & 2021 & UFPE & Dissertação Mestrado & Geral \\
\midrule
\cite{melo2022politica} & Política & 2022 & UFPE & Dissertação Mestrado & Política \\
\cite{cunha2022educacao} & Educação & 2022 & Unioeste  & Periódico & COVID-19, Ciências \\
\cite{souza2022tcc} & Computação & 2022 & UFC & TCC & Geral \\
\cite{quessada2022politica} & Política & 2022 & UFSCAR & Dissertação Mestrado & Política \\
\midrule
\cite{pereira2023linguistica} & Linguística & 2023 & UEM & Periódico & Geral \\
\cite{gusmao2023tcc} & Computação & 2023 & UFAM & TCC & Geral \\
\bottomrule
\end{tabular}%
} 

\caption{\novo{Publicações acadêmicas/analíticas identificadas nos resultados de busca (Padrão F3) que citam exemplos dos \textit{corpora}. Esta tabela demonstra como as alegações falsas dos conjuntos de dados analisados se tornaram objetos de estudo acadêmico em diversas áreas do conhecimento, proporcionando uma forma indireta de validação da classificação dessas alegações como \textit{fake news}.}}

\label{tab:articles}

\end{table}

\novo{As publicações acadêmicas identificadas distribuem-se ao longo de um período de seis anos (2018-2023), com a seguinte distribuição temporal: 1 em 2018, 3 em 2019, 1 em 2020, 11 em 2021, 4 em 2022, e 2 em 2023. O aumento significativo de publicações em 2021 coincide com o período de maior intensidade da pandemia de COVID-19, refletindo o interesse acadêmico crescente no estudo da \textit{fake news} relacionada à saúde pública durante esse período crítico.}

\novo{Quanto às áreas do conhecimento, os trabalhos dividem-se em: 6 da área de Comunicação, 5 de Computação, 3 de Linguística, 3 de Política, 1 de Biblioteconomia, 1 de Administração, 1 de Ciências Sociais, 1 de Educação e 1 de Química. Esta diversidade disciplinar demonstra como o fenômeno das \textit{fake news} atravessa fronteiras acadêmicas tradicionais, sendo objeto de estudo tanto em campos relacionados à tecnologia quanto em ciências humanas e sociais.}

\novo{Entre as publicações da área de Computação, quatro são trabalhos de conclusão de curso (TCCs) e uma é dissertação de mestrado. Os TCCs \cite{sakurai2019tcc, santos2020si, gusmao2023tcc, souza2022tcc} abordam principalmente os aspectos técnicos da detecção automática de \textit{fake news}, utilizando o \textit{corpus} Fake.br como conjunto de dados para treinamento e avaliação de modelos. Estes trabalhos exploram desde modelos tradicionais de aprendizado de máquina até arquiteturas mais avançadas como BERT e LSTM, demonstrando a relevância dos conjuntos de dados analisados para o desenvolvimento tecnológico no campo da verificação automática.}

\novo{A dissertação de mestrado em Computação \cite{sa2021master} é de autoria do criador do \textit{corpus} COVID19.BR. Concentra-se no processo de coleta de dados denominado "farol digital de monitoramento de WhatsApp Públicos" \cite{sa2021iceis}, destacando a importância de metodologias específicas para a coleta e análise de \textit{fake news} em plataformas de mensagens privadas.}

\novo{Das publicações na área de Linguística, \cite{pereira2023linguistica} analisa diretamente o Fake.br, investigando as estratégias linguísticas de manipulação utilizadas na \textit{fake news}. O estudo identifica a evidencialidade como uma tática comum, na qual alegações falsas ganham credibilidade aparente ao atribuírem informações a figuras públicas conhecidas. Por exemplo, em "Alckmin diz que por ele PSDB 'desembarca', mas não explica se utilizará o aparelho do filme MIB", o nome de uma figura política conhecida (Geraldo Alckmin) é instrumentalizado para conferir credibilidade à informação falsa.}

\novo{Dos \textit{corpora} que são diretamente mencionados, portanto, somente o Fake.br por \cite{sakurai2019tcc, gusmao2023tcc, santos2020si, souza2022tcc, pereira2023linguistica} e o COVID19.BR por \cite{sa2021master}.}

\novo{Três publicações \cite{barbieri2021mestrado, cunha2022educacao, quintanilha2021mestrado} investigam especificamente como professores podem inadvertidamente propagar \textit{fake news} sobre COVID-19 em contextos educacionais. Barbieri \cite{barbieri2021mestrado} e Cunha \cite{cunha2022educacao} focam em professores do ensino fundamental brasileiro, enquanto Quintanilha \cite{quintanilha2021mestrado} analisa o conhecimento de professores de química sobre a COVID-19 em Portugal, sendo este o único dos 23 trabalhos analisados proveniente de uma instituição não brasileira.}

\novo{A ampla adoção dos \textit{corpora} Fake.br e COVID19.BR em pesquisas acadêmicas diversas não apenas valida a qualidade desses conjuntos de dados, mas também demonstra como o estudo da \textit{fake news} se consolidou como campo de pesquisa interdisciplinar. Este fenômeno cria um ciclo interessante onde a identificação de exemplos de \textit{fake news} alimenta estudos acadêmicos que, por sua vez, podem ser detectados por sistemas automatizados de verificação como evidência indireta da classificação dessas informações como falsas.}

\subsection{\novo{Padrões em Casos com Extração de Alegação}}
\label{subsec:no_match}

\novo{Nos casos em que a correspondência direta falhou e foi necessária a extração de alegação via LLM, a análise qualitativa focou na natureza da extração e na relevância dos resultados da segunda busca (realizada com a alegação extraída).}

\begin{figure}[ht]
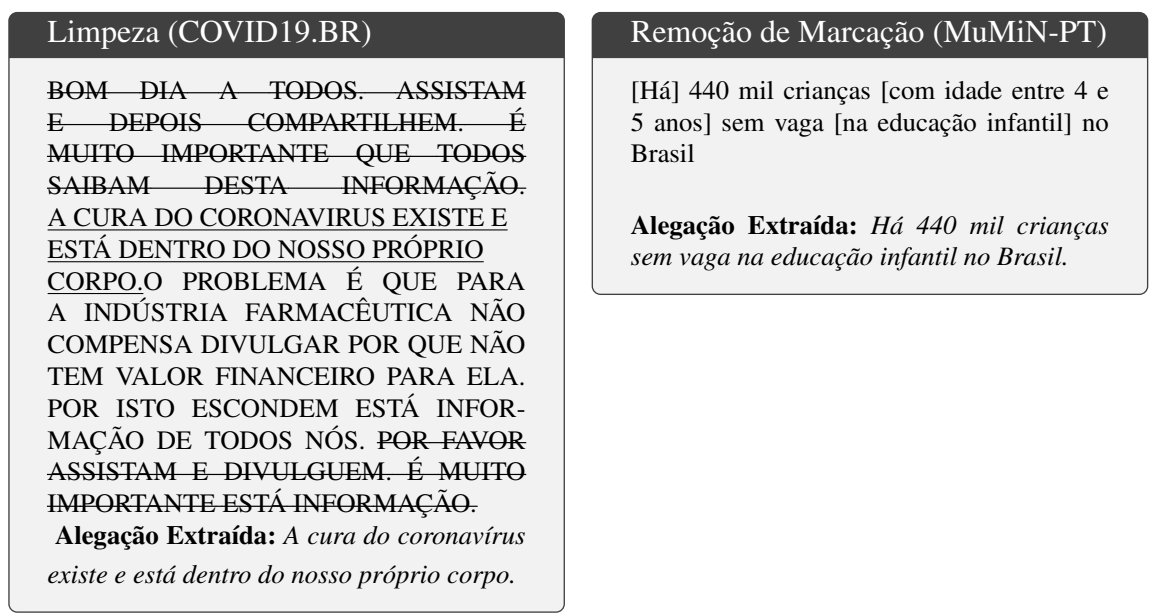

\begin{multicols}{2}
\centering
\begin{tcolorbox}[colback=black!5, boxrule=0.5pt, arc=1mm, title=Limpeza (COVID19.BR)]
\footnotesize
\st{BOM DIA A TODOS.
ASSISTAM E DEPOIS COMPARTILHEM.
É MUITO IMPORTANTE QUE TODOS SAIBAM DESTA INFORMAÇÃO.}
\underline{A CURA DO CORONAVIRUS EXISTE E} \underline{ESTÁ DENTRO DO NOSSO PRÓPRIO} \underline{CORPO.}O PROBLEMA É QUE PARA A INDÚSTRIA FARMACÊUTICA NÃO COMPENSA DIVULGAR POR QUE NÃO TEM VALOR FINANCEIRO PARA ELA. POR ISTO ESCONDEM ESTÁ INFORMAÇÃO DE TODOS NÓS. \st{POR FAVOR 
ASSISTAM E DIVULGUEM. É MUITO IMPORTANTE ESTÁ INFORMAÇÃO.} \\ \vspace{1mm} \textbf{Alegação Extraída:} \textit{A cura do coronavírus existe e está dentro do nosso próprio corpo.}
\end{tcolorbox}
\columnbreak
\begin{tcolorbox}[colback=black!5, boxrule=0.5pt, arc=1mm, title=Remoção de Marcação (MuMiN-PT)]
\footnotesize
[Há] 440 mil crianças [com idade entre 4 e 5 anos] sem vaga [na educação infantil] no Brasil\\ \vspace{1mm}

\textbf{Alegação Extraída:} \textit{Há 440 mil crianças sem vaga na educação infantil no Brasil.}
\end{tcolorbox}
\end{multicols}

\caption{\novo{Exemplos de extração de alegação envolvendo limpeza textual e remoção de marcações. À esquerda, o LLM identificou e removeu elementos paratextuais característicos de mensagens de WhatsApp (saudações, chamadas à ação), isolando a alegação central. À direita, a extração removeu marcações e reconstitui a afirmação principal em forma direta e verificável.}}
\label{fig:filter}
\end{figure}

\novo{Observou-se que, em alguns casos, a extração de alegações consistiu em operações mais simples de limpeza textual, como a remoção de saudações, despedidas ou marcações específicas do texto original (e.g., colchetes, asteriscos). Essas operações podem ser importantes, pois tais elementos poderiam interferir na eficácia da busca, especialmente se contiverem caracteres especiais interpretados pelos mecanismos de busca. A Figura \ref{fig:filter} apresenta dois exemplos: à esquerda, uma limpeza de "cabeçalho" e "rodapé" em uma mensagem de WhatsApp; à direita, a remoção de marcações (colchetes) em um texto do MuMiN-PT.}

\novo{A capacidade do LLM de realizar essa sumarização abstrata foi geralmente eficaz em produzir consultas mais direcionadas para a segunda busca.} No entanto, mesmo com uma alegação bem extraída, a busca subsequente nem sempre retornava resultados relevantes ou conclusivos. \novo{O exemplo Figura da \ref{fig:exemplo2} ilustrou um caso onde a alegação foi bem extraída, mas a busca em novembro de 2024 falhou, enquanto a busca inicial na Figura \ref{fig:search_result_cse_example} pelo texto original em dezembro de 2024 havia encontrado correspondência.}

\novo{Este fenômeno sugere a influência da temporalidade e da dinâmica dos algoritmos de busca.} A relevância e a indexação de um documento podem mudar ao longo do tempo nos índices dos mecanismos de busca. Estratégias como o uso de restrições temporais na busca, quando a data original da alegação é conhecida ou pode ser estimada (como no caso do Fake.br), poderiam mitigar parcialmente esse problema, aproximando a busca do contexto temporal original da alegação.

\section{\novo{Resultados da Avaliação dos Dados}}
\label{eval_results}

\novo{Esta seção detalha os resultados obtidos a partir da avaliação dos dados, conforme a estratégia delineada na Seção \ref{sec:estrategia_avaliacao} e utilizando o ambiente computacional especificado na Seção \ref{sec:ambiente_avaliacao}. As Tabelas \ref{tab:bertimbau_results} e \ref{tab:gemini_results} apresentam, respectivamente, os desfechos do ajuste fino do modelo Bertimbau e da abordagem \textit{few-shot} com o modelo Gemini 1.5 Flash, considerando as diferentes configurações de processamento de dados estabelecidas na Seção \ref{sec:estrategia_avaliacao}.}

\begin{table}[ht]
\centering
\begin{tabularx}{\textwidth}{Xccccc}
\toprule
\multirow{2}{*}{Processamento}  & \multicolumn{2}{c}{COVID19.BR} && \multicolumn{2}{c}{Fake.br} \\
 \cmidrule(lr){2-3} \cmidrule(lr){5-6}
 & Acurácia & F1 macro && Acurácia  & F1 macro \\
\midrule
\small 1. Original & 81,9 & 82,1 && \textbf{99,6} & \textbf{99,6} \\
\midrule
\small 2. Validado & 81,1 & 81,4 && 98,9 &  98,9 \\
\midrule
\small 3.a) Validado e Enriquecido completo & \textbf{\ul{82,1}} & \textbf{\ul{82,4}} && \ul{99,2} & \ul{99,2} \\
\small 3.b) Validado e Enriquecido filtrado & 77,9 & 78,3 && 98,7 & 98,8 \\
\bottomrule
\end{tabularx}
\caption{Resultados de Acurácia e F1-Macro para o ajuste fino do Bertimbau base nas diferentes configurações de processamento de dados. Os modelos foram selecionados com base no maior F1-Macro obtido no conjunto de validação e, subsequentemente, avaliados no conjunto de teste. As maiores pontuações estão em negrito e as melhores avaliações após-validação estão sublinhadas.}
\label{tab:bertimbau_results}
\end{table}

\novo{Os resultados indicam que os conjuntos de dados submetidos ao processo de validação (item 2 das tabelas) apresentaram desempenho inferior em termos de Acurácia e F1-Macro quando comparados aos dados originais (item 1), tanto para o Bertimbau (Tabela \ref{tab:bertimbau_results}) quanto para o Gemini nos datasets COVID19.BR e Fake.br (Tabela \ref{tab:gemini_results}). Este fenômeno pode ser atribuído às modificações realizadas durante o processo de validação, descritas na Seção \ref{subsec:validation}, podem ter tornado a tarefa de classificação mais desafiadora.} 

\begin{table}[ht]
\centering
\begin{tabularx}{\textwidth}{Xcc@{\hspace{0.5em}}c@{\hspace{0.5em}}cc@{\hspace{0.5em}}}
\toprule
  \multirow{2}{*}{Processamento}  & \multicolumn{2}{c}{COVID19.BR} && \multicolumn{2}{c}{Fake.br} \\
 
\cmidrule(lr){2-3} \cmidrule(lr){5-6}
& Acurácia & F1 macro && Acurácia  & F1 macro \\
\midrule
\small 1. Original & 75,2 & 75,2 && 81,5 & 81,0\\
\midrule
\small 2. Validado & 76,9 & 76,9 && \textbf{81,0} & \textbf{80,4}  \\
\midrule
\small 3.a) Val. e Enriq. completo & \textbf{79,3} & \textbf{79,3} && 77,6 & 76,7 \\
\small 3.b) Val. e Enriq. filtrado & 79,0 & 78,9 && 78,1 & 77,2 \\
\bottomrule
\end{tabularx}
\caption{Resultados de Acurácia e F1-Macro para a abordagem \textit{few-shot} com o Gemini 1.5 Flash nas diferentes configurações de processamento de dados. As maiores pontuações estão em negrito.}
\label{tab:gemini_results}
\end{table}

\novo{A análise comparativa entre os resultados obtidos com os dados apenas validados (item 2) e os dados validados e enriquecidos (itens 3.a e 3.b) revela que a etapa de enriquecimento de dados geralmente proporcionou ganhos de desempenho. Esta tendência foi observada para o Bertimbau no COVID19.BR (Tabela \ref{tab:bertimbau_results}) e para o Gemini 1.5 Flash no mesmo conjunto (Tabela \ref{tab:gemini_results}).} 

\novo{Para o dataset Fake.br, o enriquecimento completo (3.a) proporcionou uma ligeira melhora no desempenho do Bertimbau, mas resultou em queda de desempenho para o Gemini. Esta degradação no Gemini pode ser atribuída à qualidade dos exemplos utilizados na entrada do few-shot, onde aproximadamente um terço (5 dos 15) dos resultados de busca se mostraram irrelevantes – um problema possivelmente relacionado à temporalidade do \textit{corpus}. Conforme indica a Figura \ref{fig:fakebr_time}, os resultados de busca podem não apresentar uma concentração temporal adequada com os dados mais antigos do Fake.br. Uma possível solução seria a seleção de exemplos temporalmente mais representativos para o \textit{few-shot} e a aplicação de restrições temporais nas buscas, considerando as datas específicas do \textit{corpus}.}

\novo{A filtragem dos dados enriquecidos para excluir aqueles provenientes de redes sociais (item 3.b em comparação com 3.a) resultou em redução de desempenho para o Bertimbau em ambos os datasets, e também para o Gemini no COVID19.BR. Entretanto, para o Gemini no Fake.br, essa filtragem específica (3.b) superou o enriquecimento completo (3.a). A redução de desempenho observada na maioria dos cenários com a filtragem (excluindo redes sociais) sugere que: (i) informações relevantes para a classificação, presentes em fontes de redes sociais, foram eliminadas no processo de filtragem; ou (ii) as páginas web não classificadas como redes sociais continham características distintivas de notícias falsas/verdadeiras que foram perdidas com essa filtragem específica, ou ainda que o conteúdo de redes sociais pode ter sido mais ruidoso para esses casos.}

\novo{Os resultados demonstram consistentemente que o ajuste fino obteve resultados superiores ao \textit{few-shot learning} em todos os cenários avaliados. Esta observação corrobora a literatura científica que indica que, havendo quantidade suficiente de exemplos de treinamento, o ajuste fino tende a superar as abordagens de \textit{few-shot} \cite{qiu2024chatgpt_bert, yang2024llm_survey, vannguyen2024slm}.}

\section{Resumo dos Resultados}
\label{sec:partial_results}
\correcao{Este trabalho desenvolveu e validou uma metodologia de enriquecimento de \textit{corpora} de detecção de notícias falsas em português utilizando contexto externo. Os resultados apresentados demonstram tanto os benefícios quanto os desafios dessa abordagem, organizados em três etapas principais conforme a metodologia proposta.}

\subsection*{\correcao{Seleção e Validação dos Conjuntos de Dados}}
O levantamento inicial identificou 18 conjuntos de dados públicos de \textit{fake news} em português. Dentre os 11 \textit{corpora} que continham alegações textuais, foram selecionados três com características distintas que atendem aos critérios metodológicos estabelecidos: \textbf{Fake.br} \cite{monteiro2018fakebr} (7.200 notícias web de domínio geral, 12/2018), \textbf{COVID19.BR} \cite{martins2021covid19br} (5.635 mensagens WhatsApp sobre saúde, 10/2021) e \textbf{MuMiN-PT} \cite{nielsen2024mumin} (3.382 \textit{tweets} de domínio geral, 02/2022).

A análise exploratória dos dados (Seção \ref{sec:eda}) revelou que o Fake.br era relativamente balanceado, enquanto o MuMiN-PT apresentava mais exemplos falsos (devido à sua natureza \textit{bottom-up}) e o COVID19.BR, mais exemplos verdadeiros. \novo{Características textuais variaram com a plataforma: textos do MuMiN-PT eram mais curtos e homogêneos em tamanho entre rótulos; no Fake.br, notícias verdadeiras eram originalmente mais longas (normalizadas para o estudo); e no COVID19.BR, textos falsos tendiam a ser mais longos ("textões" de WhatsApp). A presença de URLs foi mais significativa no COVID19.BR, com alguns domínios fortemente correlacionados com o rótulo de veracidade (Tabela \ref{tab:domain_covid}), justificando sua remoção no pré-processamento.}

\subsection*{Aplicação do Fluxo de Enriquecimento}
\correcao{O fluxo de enriquecimento adaptativo demonstrou comportamento diferenciado entre os \textit{corpora}. A necessidade de extração de alegação variou conforme esperado: 95\% (COVID19.BR), 80\% (Fake.br) e 71\% (MuMiN-PT), com a menor taxa no MuMiN-PT devido à sua origem em notícias já verificadas por agências. As alegações extraídas pelo Gemini 1.5 Flash apresentaram consistência de tamanho (11-12 palavras em média) e qualidade de limpeza textual.}

\correcao{A recuperação de evidências via \textbf{Google Custom Search Engine (CSE)} retornou predominantemente fontes confiáveis (domínios governamentais e grandes mídias), com 21,3\% dos resultados para MuMiN-PT contendo verificações estruturadas (\texttt{ClaimReview}). A \textbf{Google FactCheck Claims Search API} apresentou taxas de sucesso diferenciadas: 58,8\% (MuMiN-PT), 4,8\% (COVID19.BR) e 0,7\% (Fake.br), com 'Aos Fatos' emergindo como principal fonte de verificação (Tabela \ref{tab:agent_fact_checker_domain}).}

\correcao{A análise qualitativa identificou padrões consistentes no enriquecimento. Para alegações verdadeiras, os resultados geralmente corroboravam a informação (Padrão V1), com raras confirmações explícitas por agências (V2). Para alegações falsas, emergiram três padrões: refutação direta (F1), reforço por fontes não confiáveis (F2) e reconhecimento acadêmico como \textit{fake news} (F3). O padrão F3 revelou 23 publicações acadêmicas citando exemplos dos \textit{corpora} estudados, principalmente com pico em 2021.}

\subsection*{Avaliação Experimental do Impacto}
\correcao{A estratégia de avaliação experimental, aplicando \textit{fine-tuning} do Bertimbau e \textit{few-shot learning} com Gemini 1.5 Flash nas três configurações de dados, revelou resultados nuançados. Conforme previsto na metodologia, os dados apenas validados geralmente apresentaram desempenho inferior aos originais (Tabelas \ref{tab:bertimbau_results} e \ref{tab:gemini_results})), confirmando o aumento da complexidade pela remoção de sinais discriminativos.}

\correcao{O enriquecimento com contexto externo demonstrou benefícios seletivos: melhorias consistentes para Bertimbau no COVID19.BR e para Gemini em ambos os \textit{corpora}, com exceção do Fake.br com Gemini, onde a idade do \textit{corpus} pode ter impactado a relevância das evidências recuperadas. A filtragem dos resultados de enriquecimento apresentou resultados mistos, sugerindo que a qualidade supera a quantidade de evidências.}
\correcao{Consistentemente, o \textit{fine-tuning} do Bertimbau superou o \textit{few-shot learning} com Gemini, alinhando-se com evidências da literatura sobre a eficácia de modelos menores especializados \cite{qiu2024chatgpt_bert, yang2024llm_survey, vannguyen2024slm}.}

\novo{Em suma, o enriquecimento adiciona contexto valioso, mas sua eficácia depende da qualidade da busca, da extração de alegações, da cobertura das APIs e da temporalidade da informação. Os resultados sugerem que uma abordagem híbrida, combinando análise textual com avaliação crítica de evidências externas (incluindo credibilidade da fonte e data), é promissora para uma detecção de \textit{fake news} mais robusta e adaptável.}

\correcao{A análise detalhada do desenvolvimento e dos resultados experimentais apresentada neste capítulo fornece a base para as conclusões e discussões sobre os próximos passos da pesquisa, que serão sumarizadas no Capítulo \ref{c:conclusion}.}
\chapter{Conclusão}
\label{c:conclusion}

Esta dissertação partiu da constatação, validada por um levantamento inicial (Hipótese \textbf{H1}), da escassez de \textit{corpora} em língua portuguesa adequados para a tarefa de Verificação Semi-Automática de Fatos baseada em evidências externas. \novo{A maioria dos recursos existentes concentra-se na classificação baseada em características intrínsecas do texto ou, quando incluem evidências, estas frequentemente se tornaram inacessíveis devido a restrições de APIs (como a do Twitter/X)}.

O objetivo central deste trabalho foi desenvolver um método para enriquecer conjuntos de dados de notícias em português já existentes, agregando evidências contextuais relevantes recuperadas de fontes externas, conforme proposto no Objetivo Geral e na Hipótese \textbf{H2}. A abordagem simulou o processo cognitivo de um usuário que verifica informações online, utilizando Modelos de Linguagem Grandes (LLMs) para extrair a alegação principal dos textos e, subsequentemente, empregando mecanismos de busca (API de busca do Google e API de busca de alegação do Google FactCheck) para recuperar evidências associadas.

\novo{Os resultados demonstraram a viabilidade da metodologia proposta. Três conjuntos de dados distintos (Fake.Br, COVID19.BR, MuMiN-PT), selecionados por suas diferentes características (fonte, método de coleta, temporalidade), foram estendidos com sucesso. A análise detalhada desses conjuntos estendidos constitui o resultado principal desta pesquisa.}

\novo{Confirmou-se a Hipótese \textbf{H3}, de que a extração de alegações via LLM (utilizando o Gemini 1.5 Flash) é uma etapa útil no processo, especialmente para textos mais longos ou menos diretos, como os encontrados no Fake.Br e COVID19.BR, onde a necessidade de extração foi maior (80\% e 95\%, respectivamente). Contudo, observou-se que, em alguns casos, principalmente para textos mais curtos ou já próximos de uma alegação verificável (mais comum no MuMiN-PT, com 71\% de necessidade de extração), a busca direta poderia retornar resultados relevantes sem a etapa de extração.}

\novo{A análise comparativa dos \textit{corpora} enriquecidos apoiou as Hipóteses \textbf{H4} (referente ao impacto do veículo de publicação) e \textbf{H5} (referente à natureza da coleta dos dados), revelando que o método de coleta (\textit{top-down} vs. \textit{bottom-up}) e a plataforma de publicação (notícias web, WhatsApp, Twitter) exercem influência significativa nas propriedades dos dados e, por conseguinte, no processo de enriquecimento. O MuMiN-PT (\textit{bottom-up}, Twitter) destacou-se pela maior correspondência com verificações de fatos preexistentes e menor demanda por extração de alegações. Adicionalmente, características textuais como o tamanho dos textos e a prevalência de quase duplicatas, notadamente alta no COVID19.BR (13,6\%), demonstraram variar consideravelmente entre os \textit{corpora}, impactando a análise de consistência e a interpretação dos dados enriquecidos.}

\correcaoPrevia{Os aprendizados sobre extração de alegações foram a base para um método de normalização, validada na competição CLEF CheckThat! 2025 \cite{CheckThat:ECIR2025, clef-checkthat:2025-lncs}. O sistema proposto, utilizando o ajuste fino do modelo Mono-PTT5 \cite{10.1007/978-3-031-79032-4_23}, provou-se altamente competitivo ao alcançar a terceira posição para o português (METEOR 0.5290) \cite{clef-checkthat:2025:task2}.}

\novo{A avaliação experimental, relacionada à Hipótese \textbf{H6}, indicou que, embora a etapa de validação dos dados, ao remover certos sinais (e.g., URLs explícitas), pudesse tornar a tarefa de classificação mais desafiadora para os modelos, o subsequente enriquecimento com conteúdo externo geralmente propiciou melhorias de desempenho sobre os dados apenas validados. Isso foi particularmente observado para o modelo Bertimbau e para o Gemini 1.5 Flash no \textit{corpus} COVID19.BR. Consistentemente, o ajuste fino do Bertimbau demonstrou superioridade em relação ao \textit{few-shot learning} com Gemini, sugerindo que o enriquecimento adiciona contexto valioso, mas sua eficácia final depende da qualidade da busca, da extração de alegações, da cobertura das APIs e da temporalidade da informação.} \correcao{Reconhece-se, contudo, que esta abordagem possui limitações intrínsecas, como a dependência de APIs comerciais e a ausência de uma avaliação granular de cada etapa do pipeline, discutidas em detalhe na Seção \ref{sec:limitations}.}

\novo{O processo de validação semi-automático, que incluiu a detecção de quase duplicatas e checagens de consistência de rótulos, mostrou-se fundamental para melhorar a qualidade dos dados base antes do enriquecimento.} \novo{A análise dos resultados de busca revelou domínios de fontes recorrentes (governamentais, grandes portais de notícia) e diferenças importantes entre a busca web geral e a API do Google FactCheck. Notavelmente, a última retornou predominantemente verificações classificadas como falsas ou enganosas, apoiando a ideia de que as agências de checagem focam no desmentido.} \novo{A análise qualitativa de casos com correspondência direta na busca web destacou a importância crucial da avaliação da credibilidade da fonte e identificou cenários recorrentes, como a citação de \textit{fake news} em trabalhos acadêmicos (Padrão F3), com 23 publicações identificadas.}

\novo{As contribuições desta dissertação, detalhadas na Seção \ref{sec:contribution}, podem ser sumarizadas como: (1) um levantamento e análise comparativa aprofundada de \textit{corpora} de \textit{fake news} em português, incluindo características não exploradas anteriormente como métodos de coleta e quase duplicatas; (2) um processo de validação de dados que melhora a confiabilidade dos \textit{corpora}; (3) o desenvolvimento e aplicação de uma metodologia de enriquecimento utilizando LLMs e APIs de busca; (4) uma análise detalhada dos dados estendidos, fornecendo \textit{insights} sobre o processo de extração, as fontes de evidência e as características dos diferentes tipos de \textit{fake news}; e (5) uma avaliação experimental do impacto das etapas de validação e enriquecimento no desempenho de modelos de detecção de \textit{fake news}.}

\correcao{O trabalho demonstra que é possível enriquecer sistematicamente \textit{corpora} de verificação de fatos em português com evidências contextuais, proporcionando recursos mais robustos para a pesquisa e desenvolvimento de sistemas de combate à desinformação.} A metodologia proposta, embora apresente limitações relacionadas à dependência de APIs comerciais e à ausência de avaliação granular, estabelece uma base sólida para futuras pesquisas na área.

\section{Limitações e Trabalhos Futuros}
\label{sec:limitations}

\noindent \correcao{As limitações desta pesquisa podem ser categorizadas em três dimensões principais:}

\correcao{\textbf{Limitações dos dados:}} A temporalidade dos conjuntos de dados representa um desafio, tanto pela desatualização de certos tópicos (e.g., referências políticas específicas no Fake.br) quanto pela dinâmica da relevância dos resultados de busca, que pode decair com o tempo. A dependência de rótulos binários (verdadeiro/falso) nos \textit{corpora} originais simplifica a complexidade da veracidade, embora a API do Google FactCheck tenha permitido recuperar alguma granularidade (e.g., "enganoso"). Uma restrição adicional é a falta de verificabilidade associada a determinadas afirmações, como aquelas relacionadas a experiências pessoais.

\correcao{\textbf{Limitações metodológicas:}} As limitações dos métodos incluem a dependência de tecnologias específicas (Gemini 1.5 Flash, Google APIs), o que afeta a replicabilidade e generalização dos resultados, agravada pela ausência de uma semente determinística (\textit{seed}) na API do Gemini utilizada e pela natureza personalizada dos resultados de busca do Google. A abordagem atual não considera também \textit{claim splitting}, isto é, a separação de várias alegações em um texto. \correcao{Além disso, como uma limitação central do método, não foi conduzida uma avaliação objetiva e granular de cada etapa do processo de enriquecimento. Tal avaliação permitiria quantificar o impacto isolado da qualidade da extração de alegações e da relevância dos resultados de busca no desempenho final, oferecendo uma compreensão mais profunda sobre quais componentes da metodologia contribuem mais significativamente para o resultado.}

\correcao{\textbf{Limitações de escopo:} O estudo não aborda desafios emergentes como a desinformação gerada por LLMs nem explora a aplicação prática da metodologia em cenários reais de agências de checagem de fatos.}

\correcao{Com base nas limitações identificadas e nos resultados obtidos, propõem-se as seguintes direções para pesquisas futuras, organizadas em três grandes áreas:}

\noindent \correcao{\textbf{Aprimoramento metodológico:}}
\begin{itemize}
\item \correcao{Avaliar objetivamente cada etapa do processo nos \textit{corpora} gerados, desenvolvendo métricas para aferir a qualidade dos \textit{prompts} de extração de alegação e a relevância das evidências recuperadas com e sem a extração, de forma a quantificar a eficácia da tarefa de enriquecimento como um todo.}
\item Realizar buscas contextualmente mais precisas, restringindo o período temporal para simular a verificação no momento da publicação original, sobretudo para \textit{corpora} como o Fake.br.
\item \novo{Implementar a coleta completa do conteúdo de páginas Web, superando a superficialidade dos trechos retornados por APIs de busca, a fim de assegurar uma análise contextual mais robusta e abrangente.}
\item Desenvolver e testar técnicas de \textit{claim splitting} para processar textos com múltiplas alegações, comparando-as com a abordagem focada na alegação principal.
\end{itemize}

\noindent \correcao{\textbf{Expansão teórica e técnica:}}
\begin{itemize}
\item Investigar a evidencialidade linguística e seu papel na verificação de fatos, examinando o tratamento de fontes e informações atribuídas nos dados e por modelos computacionais \cite{pereira2023linguistica}.
\item Explorar modelos de verificação mais avançados, como agentes baseados em LLMs \cite{nakano2022webgpt, yao2022react}, capazes de interação iterativa com mecanismos de busca e refinamento de consultas.
\item \correcao{Explorar a geração de texto sintético utilizando LLMs, como uma estratégia para aumentar o volume e a diversidade dos \textit{corpora} em português. Conforme demonstrado por \cite{choi2024synthetic}, essa técnica permite criar conjuntos de dados amplos e balanceados para tarefas de verificação de fatos, superando a dificuldade de encontrar exemplos naturais em quantidade suficiente.}
\item \correcao{Investigar o combate à desinformação gerada por Modelos de Linguagem Grandes LLMs, um desafio crescente que não foi objeto deste estudo. Esta linha de pesquisa futura abordaria tanto a detecção de conteúdo sintético, que pode ser mais convincente e produzido em larga escala, quanto a adaptação da metodologia de enriquecimento de evidências para enfrentar essa nova ameaça, conforme explorado por \cite{chen2024llm_fake1, liu2024llmfake2, zhang2024llmfake3}.}
\end{itemize}

\noindent \correcao{\textbf{Aplicação prática e escalabilidade:}}
\begin{itemize}
\item Mitigar desafios de reprodutibilidade e custos por meio da experimentação com mecanismos de busca alternativos (e.g., DuckDuckGo, Mojeek) e da busca por maior determinismo nos LLMs.
\item \correcao{Medir o impacto da metodologia em aplicações práticas, desenvolvendo sistemas de apoio à decisão para agências de checagem de fatos. Tal colaboração permitiria avaliar a eficácia do enriquecimento automático na otimização do fluxo de trabalho dos verificadores humanos e na aceleração do processo de desmentido.}
\item \correcao{Mensurar a escalabilidade da metodologia, tanto em termos de custo computacional (especialmente o uso de APIs comerciais) quanto de vazão de processamento, para viabilizar sua aplicação em cenários de grande volume de dados, como o monitoramento contínuo de redes sociais.}
\end{itemize}

\correcao{Essas direções futuras visam não apenas superar as limitações identificadas, mas também expandir o impacto da pesquisa para cenários práticos e desafios emergentes no combate à desinformação.}

\cleardoublepage

\arial
\bibliography{ref}

\begin{thebibliography}{100}

\bibitem{abbasiantaeb2024generateretrieve}
{\sffamily\scshape Abbasiantaeb, Z.; Aliannejadi, M.}
\newblock {\bfseries Generate then retrieve: Conversational response retrieval using llms as answer and query generators}, 2024.
\newblock Disponível em \url{https://arxiv.org/abs/2403.19302}.

\bibitem{addullah2023sent}
{\sffamily\scshape Abdullah, T.; Ahmet, A.}
\newblock {\bfseries Deep learning in sentiment analysis: Recent architectures}.
\newblock {\itshape ACM Comput. Surv.}, 55(8), dec 2022.
\newblock Disponível em \url{https://doi.org/10.1145/3548772}.

\bibitem{ahmad2022systematic}
{\sffamily\scshape Ahmad, T.; Aliaga~Lazarte, E.~A.; Mirjalili, S.}
\newblock {\bfseries A systematic literature review on fake news in the covid-19 pandemic: Can ai propose a solution?}
\newblock {\itshape Applied Sciences}, 12(24), 2022.
\newblock Disponível em \url{https://www.mdpi.com/2076-3417/12/24/12727}.

\bibitem{aimeur2023review}
{\sffamily\scshape Aimeur, E.; Amri, S.; Brassard, G.}
\newblock {\bfseries Fake news, disinformation and misinformation in social media: a review}.
\newblock {\itshape Social Network Analysis and Mining}, 13(1):30, 2023.

\bibitem{CheckThat:ECIR2025}
{\sffamily\scshape Alam, F.; Stru{\ss}, J.~M.; Chakraborty, T.; Dietze, S.; Hafid, S.; Korre, K.; Muti, A.; Nakov, P.; Ruggeri, F.; Schellhammer, S.; Setty, V.; Sundriyal, M.; Todorov, K.; V., V.}
\newblock {\bfseries The clef-2025 checkthat! lab: Subjectivity, fact-checking, claim normalization, and retrieval}.
\newblock In: Hauff, C.; Macdonald, C.; Jannach, D.; Kazai, G.; Nardini, F.~M.; Pinelli, F.; Silvestri, F.; Tonellotto, N., editors, {\itshape Advances in Information Retrieval}, p. 467--478, Cham, 2025. Springer Nature Switzerland.

\bibitem{clef-checkthat:2025-lncs}
{\sffamily\scshape Alam, F.; Struß, J.~M.; Chakraborty, T.; Dietze, S.; Hafid, S.; Korre, K.; Muti, A.; Nakov, P.; Ruggeri, F.; Schellhammer, S.; Setty, V.; Sundriyal, M.; Todorov, K.; Venktesh, V.}
\newblock {\bfseries Overview of the {CLEF}-2025 {CheckThat! Lab}: Subjectivity, fact-checking, claim normalization, and retrieval}.
\newblock In: Carrillo-de Albornoz, J.; Gonzalo, J.; Plaza, L.; García Seco~de Herrera, A.; Mothe, J.; Piroi, F.; Rosso, P.; Spina, D.; Faggioli, G.; Ferro, N., editors, {\itshape Experimental IR Meets Multilinguality, Multimodality, and Interaction. Proceedings of the Sixteenth International Conference of the CLEF Association (CLEF 2025)}, 2025.

\bibitem{ali2022survey}
{\sffamily\scshape Ali, I.; Ayub, M. N.~B.; Shivakumara, P.; Noor, N. F. B.~M.}
\newblock {\bfseries Fake news detection techniques on social media: A survey}.
\newblock {\itshape Wireless Communications and Mobile Computing}, 2022, 2022.

\bibitem{almeida2021dataset-fake-news}
{\sffamily\scshape Almeida, L.; Fuzaro, V.; Nieto, F.~V.; Santana, A.}
\newblock {\bfseries Identificação de “fake news” no contexto político brasileiro: uma abordagem computacional}.
\newblock In: {\itshape Anais do II Workshop sobre as Implicações da Computação na Sociedade}, p. 78--89, Porto Alegre, RS, Brasil, 2021. SBC.
\newblock Disponível em \url{https://sol.sbc.org.br/index.php/wics/article/view/15966}.

\bibitem{arndt2021psicologia}
{\sffamily\scshape Arndt, G.~J.; Trindade, M.~T.; de~Oliveira~Alves, J.; de~Barros Pinto~Miguel, R.}
\newblock {\bfseries Quem é de direita toma cloroquina, quem é de esquerda toma... vacina}.
\newblock {\itshape Revista Psicología Política}, 21(51):608--626, 2021.
\newblock Disponível em \url{https://dialnet.unirioja.es/servlet/articulo?codigo=8093425}.

\bibitem{barbieri2021mestrado}
{\sffamily\scshape Barbieri, A.}
\newblock {\bfseries Tem dúvida? não compartilhe! o uso de fake news por professores de língua portuguesa do ensino fundamental ii com o propósito de desenvolver habilidades em educação midiática com seus alunos}.
\newblock Dissertação (mestrado), Universidade Tuiuti do Paraná, Curitiba, 10 2021.
\newblock Orientadora: Mônica Cristine Fort. Disponível em \url{https://tede.utp.br/jspui/handle/tede/1856}.

\bibitem{barbosa2022SIRENE-news}
{\sffamily\scshape Barbosa, V.~N.; Mendes~Neto, F.~M.; Filho, S.~A.; Silva, L.}
\newblock {\bfseries A comparative study of machine learning algorithms for the detection of fake news on the internet}.
\newblock In: {\itshape Proceedings of the XVIII Brazilian Symposium on Information Systems}, SBSI '22, New York, NY, USA, 2022. Association for Computing Machinery.
\newblock Disponível em \url{https://doi.org/10.1145/3535511.3535550}.

\bibitem{batista2021adm}
{\sffamily\scshape Batista, S.~M.}
\newblock {\bfseries Onde os fatos não têm vez: uma análise foucaultiana das fake news relativas à cultura}.
\newblock Dissertação (mestrado em administração), Universidade Federal de Pernambuco, Recife, 2 2020.
\newblock Orientador: Sérgio Carvalho Benício de Mello. Disponível em \url{https://repositorio.ufpe.br/handle/123456789/39074}.

\bibitem{bojanowski2017fasttext}
{\sffamily\scshape Bojanowski, P.; Grave, E.; Joulin, A.; Mikolov, T.}
\newblock {\bfseries Enriching word vectors with subword information}.
\newblock {\itshape Transactions of the Association for Computational Linguistics}, 5:135--146, 2017.
\newblock Disponível em \url{https://aclanthology.org/Q17-1010}.

\bibitem{cunha2022educacao}
{\sffamily\scshape Borin~da Cunha, M.; Tilschneider Garcia~Rosa, B.}
\newblock {\bfseries Fake science: proposta de análise}.
\newblock {\itshape Góndola, Enseñanza y Aprendizaje de las Ciencias}, 17(3):520–538, 11 2022.
\newblock Disponível em \url{https://revistas.udistrital.edu.co/index.php/GDLA/article/view/18098}.

\bibitem{broder_minhash_2000}
{\sffamily\scshape Broder, A.~Z.}
\newblock {\bfseries Identifying and filtering near-duplicate documents}.
\newblock In: Giancarlo, R.; Sankoff, D., editors, {\itshape Combinatorial Pattern Matching}, p. 1--10, Berlin, Heidelberg, 2000. Springer Berlin Heidelberg.

\bibitem{brown2020gpt3}
{\sffamily\scshape Brown, T.~B.; Mann, B.; Ryder, N.; Subbiah, M.; Kaplan, J.; Dhariwal, P.; Neelakantan, A.; Shyam, P.; Sastry, G.; Askell, A.; Agarwal, S.; Herbert-Voss, A.; Krueger, G.; Henighan, T.; Child, R.; Ramesh, A.; Ziegler, D.~M.; Wu, J.; Winter, C.; Hesse, C.; Chen, M.; Sigler, E.; Litwin, M.; Gray, S.; Chess, B.; Clark, J.; Berner, C.; McCandlish, S.; Radford, A.; Sutskever, I.; Amodei, D.}
\newblock {\bfseries Language models are few-shot learners}, 2020.
\newblock Disponível em \url{https://arxiv.org/abs/2005.14165}.

\bibitem{cabral2021fakewhastapp}
{\sffamily\scshape Cabral, L.; Monteiro, J.~M.; da~Silva, J. W.~F.; Mattos, C. L.~C.; Mourao, P. J.~C.}
\newblock {\bfseries Fakewhastapp.br: Nlp and machine learning techniques for misinformation detection in brazilian portuguese whatsapp messages.}
\newblock In: {\itshape ICEIS (1)}, p. 63--74, 2021.

\bibitem{capistrano2021tcc}
{\sffamily\scshape Capistrano, F. F.~D.}
\newblock {\bfseries Fake news sobre a covid-19 nas aulas de química: uma abordagem didática na monitoria das práticas de ensino}.
\newblock Trabalho de conclusão de curso (licenciatura em química), Universidade Federal do Pará, Ananindeua, 8 2022.
\newblock Orientadora: Janes Kened Rodrigues dos Santos. Disponível em \url{https://bdm.ufpa.br/jspui/handle/prefix/4440}.

\bibitem{bpln2024pln}
{\sffamily\scshape Caseli, H. d.~M.; Nunes, M. d. G.~V.; Pagano, A.}
\newblock {\bfseries O que é pln?}
\newblock In: Caseli, H.~M.; Nunes, M. G.~V., editors, {\itshape Processamento de Linguagem Natural: Conceitos, Técnicas e Aplicações em Português}, book chapter~1. BPLN, 2 edition, 2024.
\newblock Disponível em \url{https://brasileiraspln.com/livro-pln/2a-edicao/parte-introducao/cap-introducao/cap-introducao.html}.

\bibitem{charles2022Fakepedia}
{\sffamily\scshape Charles, A.~C.; Ruback, L.; Oliveira, J.}
\newblock {\bfseries Faketruebr: Um corpus brasileiro de notícias falsas}.
\newblock In: Pinheiro, V.; Gamallo, P.; Amaro, R.; Scarton, C.; Batista, F.; Silva, D.; Magro, C.; Pinto, H., editors, {\itshape Anais da XVIII Escola Regional de Banco de Dados}, p. 108--117, Porto Alegre, RS, Brasil, 2023. SBC.
\newblock Disponível em \url{https://sol.sbc.org.br/index.php/erbd/article/view/24352}.

\bibitem{chase2022langchain}
{\sffamily\scshape Chase, H.}
\newblock {\bfseries {LangChain}}.
\newblock Disponível em \url{https://github.com/langchain-ai/langchain}, 8 2022.
\newblock Acesso em 6 de jul. de 2024.

\bibitem{chen2023llm_fake_survey}
{\sffamily\scshape Chen, C.; Shu, K.}
\newblock {\bfseries Combating misinformation in the age of llms: Opportunities and challenges}.
\newblock {\itshape AI Magazine}, 45(3):354--368, 2024.
\newblock Disponível em \url{https://onlinelibrary.wiley.com/doi/abs/10.1002/aaai.12188}.

\bibitem{chen2024llm_fake1}
{\sffamily\scshape Chen, C.; Shu, K.}
\newblock {\bfseries Combating misinformation in the age of llms: Opportunities and challenges}.
\newblock {\itshape AI Magazine}, 45(3):354--368, 2024.

\bibitem{chen2022gere}
{\sffamily\scshape Chen, J.; Zhang, R.; Guo, J.; Fan, Y.; Cheng, X.}
\newblock {\bfseries Gere: Generative evidence retrieval for fact verification}.
\newblock In: {\itshape Proceedings of the 45th International ACM SIGIR Conference on Research and Development in Information Retrieval}, SIGIR '22, p. 2184–2189, New York, NY, USA, 2022. Association for Computing Machinery.
\newblock Disponível em \url{https://doi.org/10.1145/3477495.3531827}.

\bibitem{cho2022qg}
{\sffamily\scshape Cho, S.; Jeong, S.; Yang, W.; Park, J.}
\newblock {\bfseries Query generation with external knowledge for dense retrieval}.
\newblock In: Agirre, E.; Apidianaki, M.; Vuli{\'c}, I., editors, {\itshape Proceedings of Deep Learning Inside Out (DeeLIO 2022): The 3rd Workshop on Knowledge Extraction and Integration for Deep Learning Architectures}, p. 22--32, Dublin, Ireland and Online, May 2022. Association for Computational Linguistics.
\newblock Disponível em \url{https://aclanthology.org/2022.deelio-1.3}.

\bibitem{choi2024claim_matching}
{\sffamily\scshape Choi, E.~C.; Ferrara, E.}
\newblock {\bfseries Automated claim matching with large language models: Empowering fact-checkers in the fight against misinformation}.
\newblock In: {\itshape Companion Proceedings of the ACM on Web Conference 2024}, WWW '24, p. 1441–1449, New York, NY, USA, 2024. Association for Computing Machinery.
\newblock Disponível em \url{https://doi.org/10.1145/3589335.3651910}.

\bibitem{choi2024synthetic}
{\sffamily\scshape Choi, E.~C.; Ferrara, E.}
\newblock {\bfseries Fact-gpt: Fact-checking augmentation via claim matching with llms}.
\newblock In: {\itshape Companion Proceedings of the ACM Web Conference 2024}, WWW '24, p. 883–886, New York, NY, USA, 2024. Association for Computing Machinery.

\bibitem{choi2021decontextualization}
{\sffamily\scshape Choi, E.; Palomaki, J.; Lamm, M.; Kwiatkowski, T.; Das, D.; Collins, M.}
\newblock {\bfseries Decontextualization: Making sentences stand-alone}.
\newblock {\itshape Transactions of the Association for Computational Linguistics}, 9:447--461, 2021.
\newblock Disponível em https://aclanthology.org/2021.tacl-1.27.

\bibitem{cordeiro2019faketweetbr}
{\sffamily\scshape Cordeiro, P.~R.; Pinheiro, V.}
\newblock {\bfseries Um corpus de not{\i}cias falsas do twitter e verifica{\c{c}}ao autom{\'a}tica de rumores em l{\i}ngua portuguesa}.
\newblock In: {\itshape Proceedings of the Symposium in Information and Human Language Technology}, p. 219--228, 2019.

\bibitem{couto2021central_de_fatos}
{\sffamily\scshape Couto, J.; Pimenta, B.; de~Araújo, I.~M.; Assis, S.; Reis, J. C.~S.; da~Silva, A.~P.; Almeida, J.; Benevenuto, F.}
\newblock {\bfseries Central de fatos: Um repositório de checagens de fatos}.
\newblock In: {\itshape Anais do III Dataset Showcase Workshop}, p. 128--137, Porto Alegre, RS, Brasil, 2021. SBC.
\newblock Disponível em \url{https://sol.sbc.org.br/index.php/dsw/article/view/17421}.

\bibitem{silva2020FakeNewsSetGen}
{\sffamily\scshape da~Silva, F. R.~M.; Freire, P. M.~S.; de~Souza, M.~P.; de~A.~B.~Plenamente, G.; Goldschmidt, R.~R.}
\newblock {\bfseries Fakenewssetgen: A process to build datasets that support comparison among fake news detection methods}.
\newblock In: {\itshape Proceedings of the Brazilian Symposium on Multimedia and the Web}, WebMedia '20, p. 241–248, New York, NY, USA, 2020. Association for Computing Machinery.
\newblock Disponível em \url{https://doi.org/10.1145/3428658.3430965}.

\bibitem{melo2022politica}
{\sffamily\scshape de~Freitas~Melo, U. M.~B.}
\newblock {\bfseries Feita sob medida: a estrutura de uma notícia falsa e seu papel no convencimento do eleitor}.
\newblock Master's thesis, Universidade Federal de Pernambuco, Recife, 2 2022.
\newblock Dissertação (Mestrado em Ciência Política). Orientador: Sérgio Carvalho Benício de Mello. Disponível em \url{https://repositorio.ufpe.br/handle/123456789/44709}.

\bibitem{melo2019dissertacao}
{\sffamily\scshape de~Melo, M.~C.}
\newblock {\bfseries A pauta da desinformação: "fake news" e categorizações de pertencimento nas eleições presidenciais brasileiras de 2018}.
\newblock Dissertação (mestrado), Pontifícia Universidade Católica do Rio de Janeiro, 4 2019.
\newblock Orientadora: Adriana Andrade Braga. Disponível em \url{https://www.dbd.puc-rio.br/pergamum/tesesabertas/1712862_2019_completo.pdf}.

\bibitem{melo2021livro}
{\sffamily\scshape de~Melo, M.~C.}
\newblock {\bfseries A pauta da desinformação: as ideias por trás das "fake news" nas eleições de 2018}.
\newblock Fafich/Selo PPGCOM/UFMG, Belo Horizonte, 1 edition, 2021.
\newblock Disponível em \url{https://seloppgcomufmg.com.br/publicacao/a-pauta-da-desinformacao/}.

\bibitem{morais2020fakenewsmultilabel}
{\sffamily\scshape de~Morais, J.~I.; Abonizio, H.~Q.; Tavares, G.~M.; da~Fonseca, A.~A.; Barbon~Jr, S.}
\newblock {\bfseries A multi-label classification system to distinguish among fake, satirical, objective and legitimate news in brazilian portuguese}.
\newblock {\itshape iSys - Brazilian Journal of Information Systems}, 13(4):126–149, Jul. 2020.
\newblock Disponível em \url{https://journals-sol.sbc.org.br/index.php/isys/article/view/833}.

\bibitem{sa2021master}
{\sffamily\scshape de~Sá, I.~C.}
\newblock {\bfseries Digital lighthouse: a platform for monitoring misinformation in whatsapp public groups}.
\newblock Dissertação (mestrado), Universidade Federal do Ceará, 2021.
\newblock Orientação: Prof. Dr. José Maria da Silva Monteiro Filho. Disponível em \url{https://repositorio.ufc.br/handle/riufc/59268}.

\bibitem{sa2021iceis}
{\sffamily\scshape de~Sá, I.~C.; Monteiro, J.; da~Silva, J.~F.; Medeiros, L.; Mourão, P.; da~Cunha, L.~C.}
\newblock {\bfseries Digital lighthouse: A platform for monitoring public groups in whatsapp}.
\newblock In: {\itshape Proceedings of the 23rd International Conference on Enterprise Information Systems - Volume 1: ICEIS,}, p. 297--304. INSTICC, SciTePress, 2021.
\newblock Disponível em \url{https://www.scitepress.org/Link.aspx?doi=10.5220/0010480102970304}.

\bibitem{deka2022medical_afc}
{\sffamily\scshape Deka, P.; Jurek-Loughrey, A.; P, D.}
\newblock {\bfseries Evidence extraction to validate medical claims in fake news detection}.
\newblock In: Traina, A.; Wang, H.; Zhang, Y.; Siuly, S.; Zhou, R.; Chen, L., editors, {\itshape Health Information Science}, p. 3--15, Cham, 2022. Springer Nature Switzerland.

\bibitem{devlin2019bert}
{\sffamily\scshape Devlin, J.; Chang, M.-W.; Lee, K.; Toutanova, K.}
\newblock {\bfseries {BERT}: Pre-training of deep bidirectional transformers for language understanding}.
\newblock In: Burstein, J.; Doran, C.; Solorio, T., editors, {\itshape Proceedings of the 2019 Conference of the North {A}merican Chapter of the Association for Computational Linguistics: Human Language Technologies, Volume 1 (Long and Short Papers)}, p. 4171--4186, Minneapolis, Minnesota, June 2019. Association for Computational Linguistics.
\newblock Disponível em \url{https://aclanthology.org/N19-1423}.

\bibitem{dhall2021blockchain}
{\sffamily\scshape Dhall, S.; Dwivedi, A.~D.; Pal, S.~K.; Srivastava, G.}
\newblock {\bfseries Blockchain-based framework for reducing fake or vicious news spread on social media/messaging platforms}.
\newblock {\itshape ACM Trans. Asian Low-Resour. Lang. Inf. Process.}, 21(1), 11 2021.
\newblock Disponível em \url{https://doi.org/10.1145/3467019}.

\bibitem{fake_bert}
{\sffamily\scshape Ding, J.; Hu, Y.; Chang, H.}
\newblock {\bfseries Bert-based mental model, a better fake news detector}.
\newblock In: {\itshape Proceedings of the 2020 6th International Conference on Computing and Artificial Intelligence}, ICCAI '20, p. 396–400, New York, NY, USA, 2020. Association for Computing Machinery.
\newblock Disponível em \url{https://doi.org/10.1145/3404555.3404607}.

\bibitem{gusmao2023tcc}
{\sffamily\scshape dos Santos~Gusmão, F.}
\newblock {\bfseries Estudo comparativo de modelos de classificação textual aplicados na classificação de fake news}.
\newblock Trabalho de conclusão de curso (bacharelado em engenharia da computação), Universidade Federal do Amazonas, Manaus, 6 2023.
\newblock Orientador: José Luiz de Souza Pio. Disponível em \url{https://riu.ufam.edu.br/bitstream/prefix/6934/2/TCC_FelipeGusm%C3%A3o.pdf}.

\bibitem{el2021dropout}
{\sffamily\scshape El~Anigri, S.; Himmi, M.~M.; Mahmoudi, A.}
\newblock {\bfseries How bert's dropout fine-tuning affects text classification?}
\newblock In: Fakir, M.; Baslam, M.; El~Ayachi, R., editors, {\itshape Business Intelligence}, p. 130--139, Cham, 2021. Springer International Publishing.

\bibitem{faustini2019bracisfake}
{\sffamily\scshape Faustini, P.; Covões, T.}
\newblock {\bfseries Fake news detection using one-class classification}.
\newblock In: {\itshape 2019 8th Brazilian Conference on Intelligent Systems (BRACIS)}, p. 592--597, 2019.

\bibitem{reddy2022zero_shot}
{\sffamily\scshape Gangi~Reddy, R.; Chinthakindi, S.~C.; Fung, Y.~R.; Small, K.; Ji, H.}
\newblock {\bfseries A zero-shot claim detection framework using question answering}.
\newblock In: Calzolari, N.; Huang, C.-R.; Kim, H.; Pustejovsky, J.; Wanner, L.; Choi, K.-S.; Ryu, P.-M.; Chen, H.-H.; Donatelli, L.; Ji, H.; Kurohashi, S.; Paggio, P.; Xue, N.; Kim, S.; Hahm, Y.; He, Z.; Lee, T.~K.; Santus, E.; Bond, F.; Na, S.-H., editors, {\itshape Proceedings of the 29th International Conference on Computational Linguistics}, p. 6927--6933, Gyeongju, Republic of Korea, Oct. 2022. International Committee on Computational Linguistics.
\newblock Disponível em \url{https://aclanthology.org/2022.coling-1.603}.

\bibitem{gangi2022claim_zero}
{\sffamily\scshape Gangi~Reddy, R.; Chinthakindi, S.~C.; Fung, Y.~R.; Small, K.; Ji, H.}
\newblock {\bfseries A zero-shot claim detection framework using question answering}.
\newblock In: Calzolari, N.; Huang, C.-R.; Kim, H.; Pustejovsky, J.; Wanner, L.; Choi, K.-S.; Ryu, P.-M.; Chen, H.-H.; Donatelli, L.; Ji, H.; Kurohashi, S.; Paggio, P.; Xue, N.; Kim, S.; Hahm, Y.; He, Z.; Lee, T.~K.; Santus, E.; Bond, F.; Na, S.-H., editors, {\itshape Proceedings of the 29th International Conference on Computational Linguistics}, p. 6927--6933, Gyeongju, Republic of Korea, Oct. 2022. International Committee on Computational Linguistics.
\newblock Disponível em \url{https://aclanthology.org/2022.coling-1.603}.

\bibitem{garcia2022FakeRecogna}
{\sffamily\scshape Garcia, G.~L.; Afonso, L. C.~S.; Papa, J.~P.}
\newblock {\bfseries Fakerecogna: A new brazilian corpus for fake news detection}.
\newblock In: Pinheiro, V.; Gamallo, P.; Amaro, R.; Scarton, C.; Batista, F.; Silva, D.; Magro, C.; Pinto, H., editors, {\itshape Computational Processing of the Portuguese Language}, p. 57--67, Cham, 2022. Springer International Publishing.

\bibitem{garg2020ml_vs_dl}
{\sffamily\scshape Garg, D.; Khan, S.; Alam, M.}
\newblock {\bfseries Integrative use of iot and deep learning for agricultural applications}.
\newblock In: Singh, P.~K.; Panigrahi, B.~K.; Suryadevara, N.~K.; Sharma, S.~K.; Singh, A.~P., editors, {\itshape Proceedings of ICETIT 2019}, p. 521--531, Cham, 2020. Springer International Publishing.

\bibitem{gomes2023erigo}
{\sffamily\scshape Gomes, J.; Neto, V.; Barbosa, J.; de~Lima, E.}
\newblock {\bfseries A rapid tertiary review at the fake news domain}.
\newblock In: {\itshape Anais da XI Escola Regional de Informática de Goiás}, Porto Alegre, RS, Brasil, 2023. SBC.

\bibitem{griesshaber2020low_resource}
{\sffamily\scshape Grie{\ss}haber, D.; Maucher, J.; Vu, N.~T.}
\newblock {\bfseries Fine-tuning {BERT} for low-resource natural language understanding via active learning}.
\newblock In: Scott, D.; Bel, N.; Zong, C., editors, {\itshape Proceedings of the 28th International Conference on Computational Linguistics}, p. 1158--1171, Barcelona, Spain (Online), dec 2020. International Committee on Computational Linguistics.
\newblock Disponível em \url{https://aclanthology.org/2020.coling-main.100/}.

\bibitem{guo2022survey_afc}
{\sffamily\scshape Guo, Z.; Schlichtkrull, M.; Vlachos, A.}
\newblock {\bfseries A survey on automated fact-checking}.
\newblock {\itshape Transactions of the Association for Computational Linguistics}, 10:178--206, 2022.
\newblock Disponível em \url{https://aclanthology.org/2022.tacl-1.11}.

\bibitem{hangloo2022survey}
{\sffamily\scshape Hangloo, S.; Arora, B.}
\newblock {\bfseries Combating multimodal fake news on social media: methods, datasets, and future perspective}.
\newblock {\itshape Multimedia Systems}, 28(6):2391--2422, Dec 2022.
\newblock Disponível em \url{https://doi.org/10.1007/s00530-022-00966-y}.

\bibitem{har-peled_lsh_2012}
{\sffamily\scshape Har-Peled, S.; Indyk, P.; Motwani, R.}
\newblock {\bfseries Approximate {Nearest} {Neighbors}: {Towards} {Removing} the {Curse} of {Dimensionality}}.
\newblock {\itshape Theory of Computing}, 8(1):321--350, 2012.
\newblock Disponível em \url{https://theoryofcomputing.org/articles/v008a014}.

\bibitem{hartmann2017embeddings}
{\sffamily\scshape Hartmann, N.; Fonseca, E.; Shulby, C.; Treviso, M.; Silva, J.; Alu{\'\i}sio, S.}
\newblock {\bfseries {P}ortuguese word embeddings: Evaluating on word analogies and natural language tasks}.
\newblock In: Paetzold, G.~H.; Pinheiro, V., editors, {\itshape Proceedings of the 11th {B}razilian Symposium in Information and Human Language Technology}, p. 122--131, Uberl{\^a}ndia, Brazil, Oct. 2017. Sociedade Brasileira de Computa{\c{c}}{\~a}o.
\newblock Disponível em \url{https://aclanthology.org/W17-6615}.

\bibitem{hu2024llm_fake}
{\sffamily\scshape Hu, B.; Sheng, Q.; Cao, J.; Shi, Y.; Li, Y.; Wang, D.; Qi, P.}
\newblock {\bfseries Bad actor, good advisor: exploring the role of large language models in fake news detection}.
\newblock 2024.
\newblock Disponível em \url{https://doi.org/10.1609/aaai.v38i20.30214}.

\bibitem{ituassu2019comunicacao}
{\sffamily\scshape Ituassu, A.; Lifschitz, S.; Capone, L.; Mannheimer, V.}
\newblock {\bfseries De donald trump a jair bolsonaro: democracia e comunicação política digital nas eleições de 2016, nos estados unidos, e 2018, no brasil}.
\newblock In: {\itshape Anais do 8º Congresso da Associação Brasileira de Pesquisadores em Comunicação e Política}, Brasília, 5 2019. Universidade de Brasília, Associação Brasileira de Pesquisadores em Comunicação e Política - Compolítica.
\newblock Disponível em \url{https://compolitica.org/novo/anais/2019_gt4_Ituassu.pdf}.

\bibitem{jakobson1961linguistics}
{\sffamily\scshape Jakobson, R.}
\newblock {\bfseries Linguistics and communications theory}.
\newblock American mathematical society, 1961.

\bibitem{kamoi2023wice}
{\sffamily\scshape Kamoi, R.; Goyal, T.; Diego~Rodriguez, J.; Durrett, G.}
\newblock {\bfseries {W}i{CE}: Real-world entailment for claims in {W}ikipedia}.
\newblock In: Bouamor, H.; Pino, J.; Bali, K., editors, {\itshape Proceedings of the 2023 Conference on Empirical Methods in Natural Language Processing}, p. 7561--7583, Singapore, Dec. 2023. Association for Computational Linguistics.
\newblock Disponível em \url{https://aclanthology.org/2023.emnlp-main.470}.

\bibitem{karpukhin2020dpr}
{\sffamily\scshape Karpukhin, V.; Oguz, B.; Min, S.; Lewis, P.; Wu, L.; Edunov, S.; Chen, D.; Yih, W.-t.}
\newblock {\bfseries Dense passage retrieval for open-domain question answering}.
\newblock In: Webber, B.; Cohn, T.; He, Y.; Liu, Y., editors, {\itshape Proceedings of the 2020 Conference on Empirical Methods in Natural Language Processing (EMNLP)}, p. 6769--6781, Online, Nov. 2020. Association for Computational Linguistics.
\newblock Disponível em \url{https://aclanthology.org/2020.emnlp-main.550}.

\bibitem{kazemi2022Fact-check_tweet}
{\sffamily\scshape Kazemi, A.; Li, Z.; P{\'e}rez-Rosas, V.; Hale, S.~A.; Mihalcea, R.}
\newblock {\bfseries Matching tweets with applicable fact-checks across languages}.
\newblock In: {\itshape Proceedings of De-Factify: Workshop on Multimodal Fact Checking and Hate Speech Detection, CEUR}, 2022.

\bibitem{kim2024llm_explain}
{\sffamily\scshape Kim, K.; Lee, S.; Huang, K.-H.; Chan, H.~P.; Li, M.; Ji, H.}
\newblock {\bfseries Can llms produce faithful explanations for fact-checking? towards faithful explainable fact-checking via multi-agent debate}, 2024.
\newblock Disponível em \url{https://arxiv.org/abs/2402.07401}.

\bibitem{kondamudi2021systematic}
{\sffamily\scshape Kondamudi, M.~R.; Sahoo, S.~R.; Chouhan, L.; Yadav, N.}
\newblock {\bfseries A comprehensive survey of fake news in social networks: Attributes, features, and detection approaches}.
\newblock {\itshape Journal of King Saud University - Computer and Information Sciences}, 35(6):101571, 2023.
\newblock Disponível em \url{https://www.sciencedirect.com/science/article/pii/S1319157823001258}.

\bibitem{kotitsas2024lora_claim}
{\sffamily\scshape Kotitsas, S.; Kounoudis, P.; Koutli, E.; Papageorgiou, H.}
\newblock {\bfseries Leveraging fine-tuned large language models with {L}o{RA} for effective claim, claimer, and claim object detection}.
\newblock In: Graham, Y.; Purver, M., editors, {\itshape Proceedings of the 18th Conference of the European Chapter of the Association for Computational Linguistics (Volume 1: Long Papers)}, p. 2540--2554, St. Julian{'}s, Malta, Mar. 2024. Association for Computational Linguistics.
\newblock Disponível em \url{https://aclanthology.org/2024.eacl-long.156}.

\bibitem{krishna2022proofver}
{\sffamily\scshape Krishna, A.; Riedel, S.; Vlachos, A.}
\newblock {\bfseries {P}roo{FV}er: Natural logic theorem proving for fact verification}.
\newblock {\itshape Transactions of the Association for Computational Linguistics}, 10:1013--1030, 2022.
\newblock Disponível em \url{https://aclanthology.org/2022.tacl-1.59}.

\bibitem{leurquin2021linguistica}
{\sffamily\scshape Leurquin, E. V. L.~F.; Leurquin, C.}
\newblock {\bfseries Fake news, desinformação e necessidade de formar leitores críticos}.
\newblock {\itshape Scripta}, 25(54):265--295, 11 2021.
\newblock Disponível em \url{https://periodicos.pucminas.br/index.php/scripta/article/view/26681}.

\bibitem{lewis2020rag}
{\sffamily\scshape Lewis, P.; Perez, E.; Piktus, A.; Petroni, F.; Karpukhin, V.; Goyal, N.; K\"{u}ttler, H.; Lewis, M.; Yih, W.-t.; Rockt\"{a}schel, T.; Riedel, S.; Kiela, D.}
\newblock {\bfseries Retrieval-augmented generation for knowledge-intensive nlp tasks}.
\newblock In: {\itshape Proceedings of the 34th International Conference on Neural Information Processing Systems}, NIPS '20, Red Hook, NY, USA, 2020. Curran Associates Inc.

\bibitem{li2024self_check}
{\sffamily\scshape Li, M.; Peng, B.; Galley, M.; Gao, J.; Zhang, Z.}
\newblock {\bfseries Self-checker: Plug-and-play modules for fact-checking with large language models}.
\newblock In: Duh, K.; Gomez, H.; Bethard, S., editors, {\itshape Findings of the Association for Computational Linguistics: NAACL 2024}, p. 163--181, Mexico City, Mexico, June 2024. Association for Computational Linguistics.
\newblock Disponível em \url{https://aclanthology.org/2024.findings-naacl.12}.

\bibitem{li2020mmcovid}
{\sffamily\scshape Li, Y.; Jiang, B.; Shu, K.; Liu, H.}
\newblock {\bfseries Mm-covid: A multilingual and multimodal data repository for combating covid-19 disinformation}, 2020.

\bibitem{lima2021tcc}
{\sffamily\scshape Lima, A.~G.}
\newblock {\bfseries A propagação de fake news e seus impactos: um estudo sobre a onda conservadora na política ocidental contemporânea}.
\newblock Trabalho de conclusão de curso (bacharelado em comunicação social com habilitação em relações públicas), Universidade de São Paulo, São Paulo, 2019.
\newblock Orientador: Luiz Alberto de Farias. Disponível em \url{https://bdta.abcd.usp.br/item/003051841}.

\bibitem{lima2021comunicacao}
{\sffamily\scshape Lima, G.; Calazans, M.; MASSI, L.}
\newblock {\bfseries Mensagens falsas sobre o novo coronavírus: legitimidade e manipulação na luta de classes}.
\newblock {\itshape Chasqui. Revista Latinoamericana de Comunicación}, 1(147):259--280, 08 2021.
\newblock Disponível em \url{https://revistachasqui.org/index.php/chasqui/article/view/4408}.

\bibitem{liu2024llmfake2}
{\sffamily\scshape Liu, A.; Sheng, Q.; Hu, X.}
\newblock {\bfseries Preventing and detecting misinformation generated by large language models}.
\newblock In: {\itshape Proceedings of the 47th International ACM SIGIR Conference on Research and Development in Information Retrieval}, SIGIR '24, p. 3001–3004, New York, NY, USA, 2024. Association for Computing Machinery.

\bibitem{liu2024llm_fake}
{\sffamily\scshape Liu, Q.; Tao, X.; Wu, J.; Wu, S.; Wang, L.}
\newblock {\bfseries Can large language models detect rumors on social media?}, 2024.
\newblock Disponível em \url{https://arxiv.org/abs/2402.03916}.

\bibitem{liu2019roberta}
{\sffamily\scshape Liu, Y.; Ott, M.; Goyal, N.; Du, J.; Joshi, M.; Chen, D.; Levy, O.; Lewis, M.; Zettlemoyer, L.; Stoyanov, V.}
\newblock {\bfseries Roberta: A robustly optimized bert pretraining approach}, 2019.
\newblock Disponível em \url{https://arxiv.org/abs/1907.11692}.

\bibitem{maia2024extension}
{\sffamily\scshape Maia, D.; da~Silva, N. F.~F.}
\newblock {\bfseries Enhancing stance detection in low-resource {B}razilian {P}ortuguese using corpus expansion generated by {GPT}-3.5}.
\newblock In: Gamallo, P.; Claro, D.; Teixeira, A.; Real, L.; Garcia, M.; Oliveira, H.~G.; Amaro, R., editors, {\itshape Proceedings of the 16th International Conference on Computational Processing of Portuguese - Vol. 1}, p. 503--508, Santiago de Compostela, Galicia/Spain, Mar. 2024. Association for Computational Lingustics.
\newblock Disponível em \url{https://aclanthology.org/2024.propor-1.51}.

\bibitem{manning2008ir_book}
{\sffamily\scshape Manning, C.~D.; Raghavan, P.; Sch\"{u}tze, H.}
\newblock {\bfseries Introduction to Information Retrieval}.
\newblock Cambridge University Press, USA, 2008.

\bibitem{martins2021covid19br}
{\sffamily\scshape Martins, A. D.~F.; Cabral, L.; Mourao, P. J.~C.; de~S{\'a}, I.~C.; Monteiro, J.~M.; Machado, J.}
\newblock {\bfseries Covid19.br: A dataset of misinformation about covid-19 in brazilian portuguese whatsapp messages}.
\newblock In: {\itshape Anais do III Dataset Showcase Workshop}, p. 138--147. SBC, 2021.

\bibitem{martins2021covid19br_modelo}
{\sffamily\scshape Martins, A.; Cabral, L.; Mourão, P.~J.; Monteiro, J.; Machado, J.}
\newblock {\bfseries Detection of misinformation about covid-19 in brazilian portuguese whatsapp messages using deep learning}.
\newblock In: {\itshape Anais do XXXVI Simpósio Brasileiro de Bancos de Dados}, p. 85--96, Porto Alegre, RS, Brasil, 2021. SBC.
\newblock Disponível em \url{https://sol.sbc.org.br/index.php/sbbd/article/view/17868}.

\bibitem{martin2022factercheck}
{\sffamily\scshape Martín, A.; Huertas-Tato, J.; Álvaro Huertas-García.; Villar-Rodríguez, G.; Camacho, D.}
\newblock {\bfseries Facter-check: Semi-automated fact-checking through semantic similarity and natural language inference}.
\newblock {\itshape Knowledge-Based Systems}, 251:109265, 2022.
\newblock Disponível em \url{https://www.sciencedirect.com/science/article/pii/S0950705122006323}.

\bibitem{meel2020survey}
{\sffamily\scshape Meel, P.; Vishwakarma, D.~K.}
\newblock {\bfseries Fake news, rumor, information pollution in social media and web: A contemporary survey of state-of-the-arts, challenges and opportunities}.
\newblock {\itshape Expert Systems with Applications}, 153:112986, 2020.
\newblock Disponível em \url{https://www.sciencedirect.com/science/article/pii/S0957417419307043}.

\bibitem{meta2024prompt}
{\sffamily\scshape Meta.}
\newblock {\bfseries How-to guides: Prompting}.
\newblock Disponível em \url{https://llama.meta.com/docs/how-to-guides/prompting/\#role-based-prompts}.
\newblock Acesso em 6 de jul. de 2024.

\bibitem{mikolov2013word2vec}
{\sffamily\scshape Mikolov, T.; Chen, K.; Corrado, G.; Dean, J.}
\newblock {\bfseries Efficient estimation of word representations in vector space}, 2013.

\bibitem{mirsarraf2017elements}
{\sffamily\scshape Mirsarraf, M.; Shairi, H.; Ahmadpanah, A.}
\newblock {\bfseries Social semiotic aspects of instagram social network}.
\newblock In: {\itshape 2017 IEEE International Conference on INnovations in Intelligent SysTems and Applications (INISTA)}, p. 460--465, 2017.

\bibitem{monteiro2018fakebr}
{\sffamily\scshape Monteiro, R.~A.; Santos, R. L.~S.; Pardo, T. A.~S.; de~Almeida, T.~A.; Ruiz, E. E.~S.; Vale, O.~A.}
\newblock {\bfseries Contributions to the study of fake news in portuguese: New corpus and automatic detection results}.
\newblock In: {\itshape Computational Processing of the Portuguese Language: 13th International Conference, PROPOR 2018, Canela, Brazil, September 24–26, 2018, Proceedings}, p. 324–334, Berlin, Heidelberg, 2018. Springer-Verlag.
\newblock Disponível em \url{https://doi.org/10.1007/978-3-319-99722-3_33}.

\bibitem{bpln2024ir}
{\sffamily\scshape Moreira, V.~P.}
\newblock {\bfseries Recuperação de informação}.
\newblock In: Caseli, H.~M.; Nunes, M. G.~V., editors, {\itshape Processamento de Linguagem Natural: Conceitos, Técnicas e Aplicações em Português}, book chapter~19. BPLN, 2 edition, 2024.
\newblock Disponível em \url{https://brasileiraspln.com/livro-pln/2a-edicao/parte-aplicacoes/cap-ir/cap-ir.html}.

\bibitem{moreno2019factck.br}
{\sffamily\scshape Moreno, J.~a.; Bressan, G.}
\newblock {\bfseries Factck.br: A new dataset to study fake news}.
\newblock In: {\itshape Proceedings of the 25th Brazillian Symposium on Multimedia and the Web}, WebMedia '19, p. 525–527, New York, NY, USA, 2019. Association for Computing Machinery.
\newblock Disponível em \url{https://doi.org/10.1145/3323503.3361698}.

\bibitem{nakano2022webgpt}
{\sffamily\scshape Nakano, R.; Hilton, J.; Balaji, S.; Wu, J.; Ouyang, L.; Kim, C.; Hesse, C.; Jain, S.; Kosaraju, V.; Saunders, W.; Jiang, X.; Cobbe, K.; Eloundou, T.; Krueger, G.; Button, K.; Knight, M.; Chess, B.; Schulman, J.}
\newblock {\bfseries Webgpt: Browser-assisted question-answering with human feedback}, 2022.
\newblock Disponível em \url{https://arxiv.org/abs/2112.09332}.

\bibitem{nascimento2021tcc}
{\sffamily\scshape Nascimento, J. G.~d.}
\newblock {\bfseries Disseminação de desinformação sobre a covid-19 em um núcleo familiar: um estudo de caso}.
\newblock Trabalho de conclusão de curso (bacharelado em biblioteconomia), Universidade Federal do Ceará, Fortaleza, 2021.
\newblock Orientador: Antonio Wagner Chacon Silva. Disponível em \url{https://repositorio.ufc.br/handle/riufc/71926}.

\bibitem{vannguyen2024slm}
{\sffamily\scshape Nguyen, C.~V.; Shen, X.; Aponte, R.; Xia, Y.; Basu, S.; Hu, Z.; Chen, J.; Parmar, M.; Kunapuli, S.; Barrow, J.; Wu, J.; Singh, A.; Wang, Y.; Gu, J.; Dernoncourt, F.; Ahmed, N.~K.; Lipka, N.; Zhang, R.; Chen, X.; Yu, T.; Kim, S.; Deilamsalehy, H.; Park, N.; Rimer, M.; Zhang, Z.; Yang, H.; Rossi, R.~A.; Nguyen, T.~H.}
\newblock {\bfseries A survey of small language models}, 2024.
\newblock Disponível em: {https://arxiv.org/abs/2410.20011}.

\bibitem{ni2024afacta}
{\sffamily\scshape Ni, J.; Shi, M.; Stammbach, D.; Sachan, M.; Ash, E.; Leippold, M.}
\newblock {\bfseries {AF}a{CTA}: Assisting the annotation of factual claim detection with reliable {LLM} annotators}.
\newblock In: Ku, L.-W.; Martins, A.; Srikumar, V., editors, {\itshape Proceedings of the 62nd Annual Meeting of the Association for Computational Linguistics (Volume 1: Long Papers)}, p. 1890--1912, Bangkok, Thailand, Aug. 2024. Association for Computational Linguistics.
\newblock Disponível em \url{https://aclanthology.org/2024.acl-long.104/}.

\bibitem{nielsen2024mumin}
{\sffamily\scshape Nielsen, D.~S.; McConville, R.}
\newblock {\bfseries Mumin: A large-scale multilingual multimodal fact-checked misinformation social network dataset}.
\newblock In: {\itshape Proceedings of the 45th International ACM SIGIR Conference on Research and Development in Information Retrieval}, p. 3141--3153, 2022.

\bibitem{talendar2021fakebr_bertimbau}
{\sffamily\scshape Nogueira, G.}
\newblock {\bfseries Br fake news detection}.
\newblock Disponível em \url{https://github.com/Talendar/br_fake_news_detection}, 2021.
\newblock Acesso em 2 de fev. de 2025.

\bibitem{nogueira2023lecture}
{\sffamily\scshape Nogueira, R.}
\newblock {\bfseries Desvendando a arquitetura transformer: Fundamentos, aplicações e perspectivas futuras}.
\newblock Disponível em \url{https://www.youtube.com/watch?v=CP2B-OWvtF8}, 12 2023.
\newblock Acesso em 3 de jul. de 2024.

\bibitem{ousidhoum2022varifocal}
{\sffamily\scshape Ousidhoum, N.; Yuan, Z.; Vlachos, A.}
\newblock {\bfseries Varifocal question generation for fact-checking}.
\newblock In: Goldberg, Y.; Kozareva, Z.; Zhang, Y., editors, {\itshape Proceedings of the 2022 Conference on Empirical Methods in Natural Language Processing}, p. 2532--2544, Abu Dhabi, United Arab Emirates, Dec. 2022. Association for Computational Linguistics.
\newblock Disponível em \url{https://aclanthology.org/2022.emnlp-main.163}.

\bibitem{bpln2024lm}
{\sffamily\scshape Paes, A.; Vianna, D.; Rodrigues, J.}
\newblock {\bfseries Modelos de linguagem}.
\newblock In: Caseli, H.~M.; Nunes, M. G.~V., editors, {\itshape Processamento de Linguagem Natural: Conceitos, Técnicas e Aplicações em Português}, book chapter~15. BPLN, 2 edition, 2024.
\newblock Disponível em \url{https://brasileiraspln.com/livro-pln/2a-edicao/parte-modelos/cap-modelos-linguagem/cap-modelos-linguagem.html}.

\bibitem{page1999pagerank}
{\sffamily\scshape Page, L.; Brin, S.; Motwani, R.; Winograd, T.}
\newblock {\bfseries The pagerank citation ranking: Bringing order to the web.}
\newblock Technical report, Stanford infolab, 1999.

\bibitem{pennington2014glove}
{\sffamily\scshape Pennington, J.; Socher, R.; Manning, C.}
\newblock {\bfseries {G}lo{V}e: Global vectors for word representation}.
\newblock In: Moschitti, A.; Pang, B.; Daelemans, W., editors, {\itshape Proceedings of the 2014 Conference on Empirical Methods in Natural Language Processing ({EMNLP})}, p. 1532--1543, Doha, Qatar, Oct. 2014. Association for Computational Linguistics.
\newblock Disponível em \url{https://aclanthology.org/D14-1162/}.

\bibitem{pereira2023linguistica}
{\sffamily\scshape Pereira, L.~P.; Antonio, J.~D.}
\newblock {\bfseries É verdade ou fake news? estratégicas linguísticas de manipulação em textos que promovem a desinformação}.
\newblock {\itshape Revista USP}, (138):27–38, 11 2023.
\newblock Disponível em \url{https://www.revistas.usp.br/revusp/article/view/218040}.

\bibitem{10.1007/978-3-031-79032-4_23}
{\sffamily\scshape Piau, M.; Lotufo, R.; Nogueira, R.}
\newblock {\bfseries ptt5-v2: A closer look at continued pretraining of t5 models for the portuguese language}.
\newblock In: Paes, A.; Verri, F. A.~N., editors, {\itshape Intelligent Systems}, p. 324--338, Cham, 2025. Springer Nature Switzerland.

\bibitem{qiu2024chatgpt_bert}
{\sffamily\scshape Qiu, Y.; Jin, Y.}
\newblock {\bfseries Chatgpt and finetuned bert: A comparative study for developing intelligent design support systems}.
\newblock {\itshape Intelligent Systems with Applications}, 21:200308, 2024.
\newblock Disponível em \url{https://www.sciencedirect.com/science/article/pii/S2667305323001333}.

\bibitem{quessada2022politica}
{\sffamily\scshape Quessada, M.}
\newblock {\bfseries Desinformação e esquerda brasileira: o discurso por trás das fake news.}
\newblock Dissertação (mestrado), Universidade Federal de São Carlos, São Carlos, 2 2022.
\newblock Orientador: Thales Haddad Novaes de Andrade. Disponível em \url{https://anpocs.org.br/wp-content/uploads/2023/07/42CPM.pdf}.

\bibitem{quintanilha2021mestrado}
{\sffamily\scshape Quintanilha, V. D.~C.}
\newblock {\bfseries Combatendo as fake news sobre o sars-cov-2: o revisor como fact-checker}.
\newblock Dissertação (mestrado), Universidade Nova, 9 2021.
\newblock Orientadora: Matilde Gonçalves. Disponível em \url{https://run.unl.pt/handle/10362/128081}.

\bibitem{rajapakse2024simpletransformers}
{\sffamily\scshape Rajapakse, T.~C.; Yates, A.; de~Rijke, M.}
\newblock {\bfseries Simple transformers: Open-source for all}.
\newblock In: {\itshape Proceedings of the 2024 Annual International ACM SIGIR Conference on Research and Development in Information Retrieval in the Asia Pacific Region}, SIGIR-AP 2024, p. 209--215, 2024.
\newblock Disponível em \url{https://doi.org/10.1145/3673791.3698412}.

\bibitem{reid2024gemini}
{\sffamily\scshape Reid, M.; Savinov, N.; Teplyashin, D.; Lepikhin, D.; Lillicrap, T.; Alayrac, J.-b.; Soricut, R.; Lazaridou, A.; Firat, O.; Schrittwieser, J.; others.}
\newblock {\bfseries Gemini 1.5: Unlocking multimodal understanding across millions of tokens of context}.
\newblock {\itshape arXiv preprint arXiv:2403.05530}, 2024.

\bibitem{reimers2019sentencebert}
{\sffamily\scshape Reimers, N.; Gurevych, I.}
\newblock {\bfseries Sentence-{BERT}: Sentence embeddings using {S}iamese {BERT}-networks}.
\newblock In: Inui, K.; Jiang, J.; Ng, V.; Wan, X., editors, {\itshape Proceedings of the 2019 Conference on Empirical Methods in Natural Language Processing and the 9th International Joint Conference on Natural Language Processing (EMNLP-IJCNLP)}, p. 3982--3992, Hong Kong, China, Nov. 2019. Association for Computational Linguistics.
\newblock Disponível em \url{https://aclanthology.org/D19-1410/}.

\bibitem{reis2019nlp}
{\sffamily\scshape Reis, J. C.~S.; Correia, A.; Murai, F.; Veloso, A.; Benevenuto, F.}
\newblock {\bfseries Supervised learning for fake news detection}.
\newblock {\itshape IEEE Intelligent Systems}, 34(2):76--81, 2019.

\bibitem{ribeiro2018sociais}
{\sffamily\scshape Ribeiro, M.~M.; Ortellado, P.}
\newblock {\bfseries O que são e como lidar com as notícias falsas}.
\newblock {\itshape Sur - Revista Internacional de Direitos Humanos}, 15(27):71--83, 2018.
\newblock Disponível em \url{https://repositorio.usp.br/item/002993181}.

\bibitem{sakurai2019tcc}
{\sffamily\scshape Sakurai, G.~Y.}
\newblock {\bfseries Processamento de linguagem natural - detecção de fake news}.
\newblock Trabalho de conclusão de curso (bacharelado em ciência da computação), Universidade Estadual de Londrina, Londrina, 2019.
\newblock Orientador: Sérgio Montazzolli Silva. Disponível em \url{https://www.uel.br/cce/dc/wp-content/uploads/TCC_GUILHERME_SAKURAI.pdf}.

\bibitem{santos2020si}
{\sffamily\scshape Santos, M.~G.}
\newblock {\bfseries DetecÇÃo de fake news: Um comparativo entre os modelos de aprendizado supervisionado passive agressive e multinomial naive bayes}.
\newblock Trabalho de conclusão de curso (bacharelado em sistemas de informação, Centro Universitário Christus - Unichristus, Fortaleza, 8 2020.
\newblock Orientador: Daniel Nascimento Teixeira. Disponível em \url{https://repositorio.unichristus.edu.br/jspui/handle/123456789/1745}.

\bibitem{schlicht2023health_systematic}
{\sffamily\scshape Schlicht, I.~B.; Fernandez, E.; Chulvi, B.; Rosso, P.}
\newblock {\bfseries Automatic detection of health misinformation: a systematic review}.
\newblock {\itshape Journal of Ambient Intelligence and Humanized Computing}, p. 1--13, 2023.

\bibitem{scirè2024fenice}
{\sffamily\scshape Scir{\`e}, A.; Ghonim, K.; Navigli, R.}
\newblock {\bfseries {FENICE}: Factuality evaluation of summarization based on natural language inference and claim extraction}.
\newblock In: Ku, L.-W.; Martins, A.; Srikumar, V., editors, {\itshape Findings of the Association for Computational Linguistics: ACL 2024}, Bangkok, Thailand, Aug. 2024. Association for Computational Linguistics.
\newblock Disponível em \url{https://aclanthology.org/2024.findings-acl.841/}.

\bibitem{shahi2020fakecovid}
{\sffamily\scshape Shahi, G.~K.; Nandini, D.}
\newblock {\bfseries FakeCovid- A Multilingual Cross-domain Fact Check News Dataset for COVID-19}.
\newblock ICWSM, Jun 2020.

\bibitem{sharma2019survey}
{\sffamily\scshape Sharma, K.; Qian, F.; Jiang, H.; Ruchansky, N.; Zhang, M.; Liu, Y.}
\newblock {\bfseries Combating fake news: A survey on identification and mitigation techniques}.
\newblock {\itshape ACM Trans. Intell. Syst. Technol.}, 10(3), 4 2019.
\newblock Disponível em \url{https://doi.org/10.1145/3305260}.

\bibitem{souza2022tcc}
{\sffamily\scshape Sousa, F. J. d.~S.}
\newblock {\bfseries Transferência de conhecimento para detecção automática de fake news com aprendizagem profunda}.
\newblock Trabalho de conclusão de curso (bacharelado em ciência da computação), Universidade Federal do Ceará, Cratéus, 2022.
\newblock Orientador: Livio Antonio de Melo Freire. Disponível em \url{https://repositorio.ufc.br/handle/riufc/70014}.

\bibitem{souza2020bertimbau}
{\sffamily\scshape Souza, F.; Nogueira, R.; Lotufo, R.}
\newblock {\bfseries Bertimbau: Pretrained bert models for brazilian portuguese}.
\newblock In: Cerri, R.; Prati, R.~C., editors, {\itshape Intelligent Systems}, p. 403--417, Cham, 2020. Springer International Publishing.

\bibitem{stahl2023lingua}
{\sffamily\scshape Stahl, P.~M.}
\newblock {\bfseries Lingua-py}.
\newblock Disponível em \url{https://github.com/pemistahl/lingua-py}, 2023.
\newblock Acesso em 3 de abril de 2024.

\bibitem{statista2024social_media}
{\sffamily\scshape Statista.}
\newblock {\bfseries Most popular social networks worldwide as of january 2024, ranked by number of monthly active users}, 2024.
\newblock Disponível em \url{https://www.statista.com/statistics/272014/global-social-networks-ranked-by-number-of-users/}.

\bibitem{sundriyal2023}
{\sffamily\scshape Sundriyal, M.; Chakraborty, T.; Nakov, P.}
\newblock {\bfseries From chaos to clarity: Claim normalization to empower fact-checking}.
\newblock In: {\itshape Findings of the Association for Computational Linguistics: EMNLP 2023}, p. 6594--6609, Singapore, 2023. Association for Computational Linguistics.

\bibitem{clef-checkthat:2025:task2}
{\sffamily\scshape Sundriyal, M.; Chakraborty, T.; Nakov, P.}
\newblock {\bfseries Overview of the {CLEF-2025 CheckThat!} lab task 2 on claim normalization}.
\newblock In: Faggioli, G.; Ferro, N.; Rosso, P.; Spina, D., editors, {\itshape Working Notes of CLEF 2025 - Conference and Labs of the Evaluation Forum}, CLEF~2025, Madrid, Spain, 2025.

\bibitem{tan2023llm_pipeline}
{\sffamily\scshape Tan, X.; Zou, B.; Aw, A.~T.}
\newblock {\bfseries Evidence-based interpretable open-domain fact-checking with large language models}, 2023.
\newblock Disponível em \url{https://arxiv.org/abs/2312.05834}.

\bibitem{tan2025coling}
{\sffamily\scshape Tan, X.; Zou, B.; Aw, A.~T.}
\newblock {\bfseries Improving explainable fact-checking with claim-evidence correlations}.
\newblock In: Rambow, O.; Wanner, L.; Apidianaki, M.; Al-Khalifa, H.; Eugenio, B.~D.; Schockaert, S., editors, {\itshape Proceedings of the 31st International Conference on Computational Linguistics}, p. 1600--1612, Abu Dhabi, UAE, Jan. 2025. Association for Computational Linguistics.

\bibitem{tang2024minicheck}
{\sffamily\scshape Tang, L.; Laban, P.; Durrett, G.}
\newblock {\bfseries {M}ini{C}heck: Efficient fact-checking of {LLM}s on grounding documents}.
\newblock In: Al-Onaizan, Y.; Bansal, M.; Chen, Y.-N., editors, {\itshape Proceedings of the 2024 Conference on Empirical Methods in Natural Language Processing}, p. 8818--8847, Miami, Florida, USA, Nov. 2024. Association for Computational Linguistics.
\newblock Disponível em \url{https://aclanthology.org/2024.emnlp-main.499/}.

\bibitem{tausczik2019liwc}
{\sffamily\scshape Tausczik, Y.~R.; Pennebaker, J.~W.}
\newblock {\bfseries The psychological meaning of words: Liwc and computerized text analysis methods}.
\newblock {\itshape Journal of Language and Social Psychology}, 29(1):24--54, 2010.
\newblock Disponível em \url{https://doi.org/10.1177/0261927X09351676}.

\bibitem{geminiteam2024gemini}
{\sffamily\scshape Team, G.; Anil, R.; Borgeaud, S.; Alayrac, J.-B.; Yu, J.; Soricut, R.; Schalkwyk, J.; Dai, A.~M.; Hauth, A.; Millican, K.; Silver, D.; Johnson, M.; Antonoglou, I.; Schrittwieser, J.; Glaese, A.; Chen, J.; Pitler, E.; Lillicrap, T.; Lazaridou, A.; Firat, O.; Molloy, J.; Isard, M.; Barham, P.~R.; Hennigan, T.; Lee, B.; Viola, F.; Reynolds, M.; Xu, Y.; Doherty, R.; etal, E.~C.}
\newblock {\bfseries Gemini: A family of highly capable multimodal models}, 2024.
\newblock Disponível em \url{https://arxiv.org/abs/2312.11805}.

\bibitem{vargas2023factnews}
{\sffamily\scshape Vargas, F.; Jaidka, K.; Pardo, T.; Benevenuto, F.}
\newblock {\bfseries Predicting sentence-level factuality of news and bias of media outlets}.
\newblock In: Mitkov, R.; Angelova, G., editors, {\itshape Proceedings of the 14th International Conference on Recent Advances in Natural Language Processing}, p. 1197--1206, Varna, Bulgaria, Sept. 2023. INCOMA Ltd., Shoumen, Bulgaria.
\newblock Disponível em \url{https://aclanthology.org/2023.ranlp-1.127}.

\bibitem{varma2021health}
{\sffamily\scshape Varma, R.; Verma, Y.; Vijayvargiya, P.; Churi, P.~P.}
\newblock {\bfseries A systematic survey on deep learning and machine learning approaches of fake news detection in the pre-and post-covid-19 pandemic}.
\newblock {\itshape International Journal of Intelligent Computing and Cybernetics}, 14(4):617--646, 2021.

\bibitem{vawasni2017attention}
{\sffamily\scshape Vaswani, A.; Shazeer, N.; Parmar, N.; Uszkoreit, J.; Jones, L.; Gomez, A.~N.; Kaiser, L.~u.; Polosukhin, I.}
\newblock {\bfseries Attention is all you need}.
\newblock In: Guyon, I.; Luxburg, U.~V.; Bengio, S.; Wallach, H.; Fergus, R.; Vishwanathan, S.; Garnett, R., editors, {\itshape Advances in Neural Information Processing Systems}, volume~30. Curran Associates, Inc., 2017.
\newblock Disponível em \url{https://proceedings.neurips.cc/paper_files/paper/2017/file/3f5ee243547dee91fbd053c1c4a845aa-Paper.pdf}.

\bibitem{villarrodríguez2024distrack}
{\sffamily\scshape Villar-Rodríguez, G.; Álvaro Huertas-García.; Martín, A.; Huertas-Tato, J.; Camacho, D.}
\newblock {\bfseries Distrack: a new tool for semi-automatic misinformation tracking in online social networks}, 2024.
\newblock Disponível em \url{https://arxiv.org/abs/2408.00633}.

\bibitem{wagner2018brwac}
{\sffamily\scshape Wagner~Filho, J.~A.; Wilkens, R.; Idiart, M.; Villavicencio, A.}
\newblock {\bfseries The br{W}a{C} corpus: A new open resource for {B}razilian {P}ortuguese}.
\newblock In: Calzolari, N.; Choukri, K.; Cieri, C.; Declerck, T.; Goggi, S.; Hasida, K.; Isahara, H.; Maegaard, B.; Mariani, J.; Mazo, H.; Moreno, A.; Odijk, J.; Piperidis, S.; Tokunaga, T., editors, {\itshape Proceedings of the Eleventh International Conference on Language Resources and Evaluation ({LREC} 2018)}, Miyazaki, Japan, May 2018. European Language Resources Association (ELRA).
\newblock Disponível em \url{https://aclanthology.org/L18-1686/}.

\bibitem{wang2018glue}
{\sffamily\scshape Wang, A.; Singh, A.; Michael, J.; Hill, F.; Levy, O.; Bowman, S.}
\newblock {\bfseries {GLUE}: A multi-task benchmark and analysis platform for natural language understanding}.
\newblock In: Linzen, T.; Chrupa{\l}a, G.; Alishahi, A., editors, {\itshape Proceedings of the 2018 {EMNLP} Workshop {B}lackbox{NLP}: Analyzing and Interpreting Neural Networks for {NLP}}, p. 353--355, Brussels, Belgium, Nov. 2018. Association for Computational Linguistics.
\newblock Disponível em \url{https://aclanthology.org/W18-5446}.

\bibitem{wang2023reasoning}
{\sffamily\scshape Wang, H.; Shu, K.}
\newblock {\bfseries Explainable claim verification via knowledge-grounded reasoning with large language models}.
\newblock In: Bouamor, H.; Pino, J.; Bali, K., editors, {\itshape Findings of the Association for Computational Linguistics: EMNLP 2023}, p. 6288--6304, Singapore, Dec. 2023. Association for Computational Linguistics.
\newblock Disponível em \url{https://aclanthology.org/2023.findings-emnlp.416}.

\bibitem{wang2024factcheckbench}
{\sffamily\scshape Wang, Y.; Gangi~Reddy, R.; Mujahid, Z.~M.; Arora, A.; Rubashevskii, A.; Geng, J.; Mohammed~Afzal, O.; Pan, L.; Borenstein, N.; Pillai, A.; Augenstein, I.; Gurevych, I.; Nakov, P.}
\newblock {\bfseries Factcheck-bench: Fine-grained evaluation benchmark for automatic fact-checkers}.
\newblock In: Al-Onaizan, Y.; Bansal, M.; Chen, Y.-N., editors, {\itshape Findings of the Association for Computational Linguistics: EMNLP 2024}, p. 14199--14230, Miami, Florida, USA, Nov. 2024. Association for Computational Linguistics.
\newblock Disponível em \url{https://aclanthology.org/2024.findings-emnlp.830/}.

\bibitem{wu2022review}
{\sffamily\scshape Wu, Y.; Ngai, E.~W.; Wu, P.; Wu, C.}
\newblock {\bfseries Fake news on the internet: a literature review, synthesis and directions for future research}.
\newblock {\itshape Internet Research}, 32:1662--1699, 1 2022.
\newblock Disponível em \url{https://doi.org/10.1108/INTR-05-2021-0294}.

\bibitem{yang2024llm_survey}
{\sffamily\scshape Yang, J.; Jin, H.; Tang, R.; Han, X.; Feng, Q.; Jiang, H.; Zhong, S.; Yin, B.; Hu, X.}
\newblock {\bfseries Harnessing the power of llms in practice: A survey on chatgpt and beyond}.
\newblock {\itshape ACM Trans. Knowl. Discov. Data}, 18(6), apr 2024.
\newblock Disponível em \url{https://doi.org/10.1145/3649506}.

\bibitem{yao2022react}
{\sffamily\scshape Yao, S.; Zhao, J.; Yu, D.; Du, N.; Shafran, I.; Narasimhan, K.; Cao, Y.}
\newblock {\bfseries React: Synergizing reasoning and acting in language models}.
\newblock {\itshape arXiv preprint arXiv:2210.03629}, 2022.

\bibitem{zeng2024justilm}
{\sffamily\scshape Zeng, F.; Gao, W.}
\newblock {\bfseries {JustiLM: Few-shot Justification Generation for Explainable Fact-Checking of Real-world Claims}}.
\newblock {\itshape Transactions of the Association for Computational Linguistics}, 12:334--354, 04 2024.
\newblock Disponível em \url{https://doi.org/10.1162/tacl\_a\_00649}.

\bibitem{zhang2024llmfake3}
{\sffamily\scshape Zhang, Y.; Sharma, K.; Du, L.; Liu, Y.}
\newblock {\bfseries Toward mitigating misinformation and social media manipulation in llm era}.
\newblock WWW '24, p. 1302–1305, New York, NY, USA, 2024. Association for Computing Machinery.

\bibitem{zhao2023LLMSurvey}
{\sffamily\scshape Zhao, W.~X.; Zhou, K.; Li, J.; Tang, T.; Wang, X.; Hou, Y.; Min, Y.; Zhang, B.; Zhang, J.; Dong, Z.; Du, Y.; Yang, C.; Chen, Y.; Chen, Z.; Jiang, J.; Ren, R.; Li, Y.; Tang, X.; Liu, Z.; Liu, P.; Nie, J.-Y.; Wen, J.-R.}
\newblock {\bfseries A survey of large language models}.
\newblock {\itshape arXiv preprint arXiv:2303.18223}, 2023.
\newblock Disponível em \url{http://arxiv.org/abs/2303.18223}.

\end{thebibliography}
\label{ref}

\apendices
\chapter{Exemplos Ilustrativos do Fluxo de Validação Semiautomático}
\label{append:validation_examples}

\novo{Este apêndice fornece exemplos concretos para ilustrar as diferentes etapas do fluxo de validação semiautomático descrito na Seção \ref{subsec:validation}. O objetivo é clarificar os critérios aplicados durante a filtragem e correção dos dados. É importante notar que a remoção de URLs dos textos, mencionada como uma etapa geral para evitar viés de domínio, foi realizada após as etapas de validação aqui exemplificadas que pudessem depender da presença dessas URLs (como a resolução de contradições baseada em URL ou a filtragem específica do Fake.br por URL de origem). Os exemplos abaixo mostram os textos antes dessa remoção final de URLs, mas com outros processamentos de validação específica sendo ilustrados.}

\section{Filtragem Inicial Automatizada}
\begin{figure}[ht]
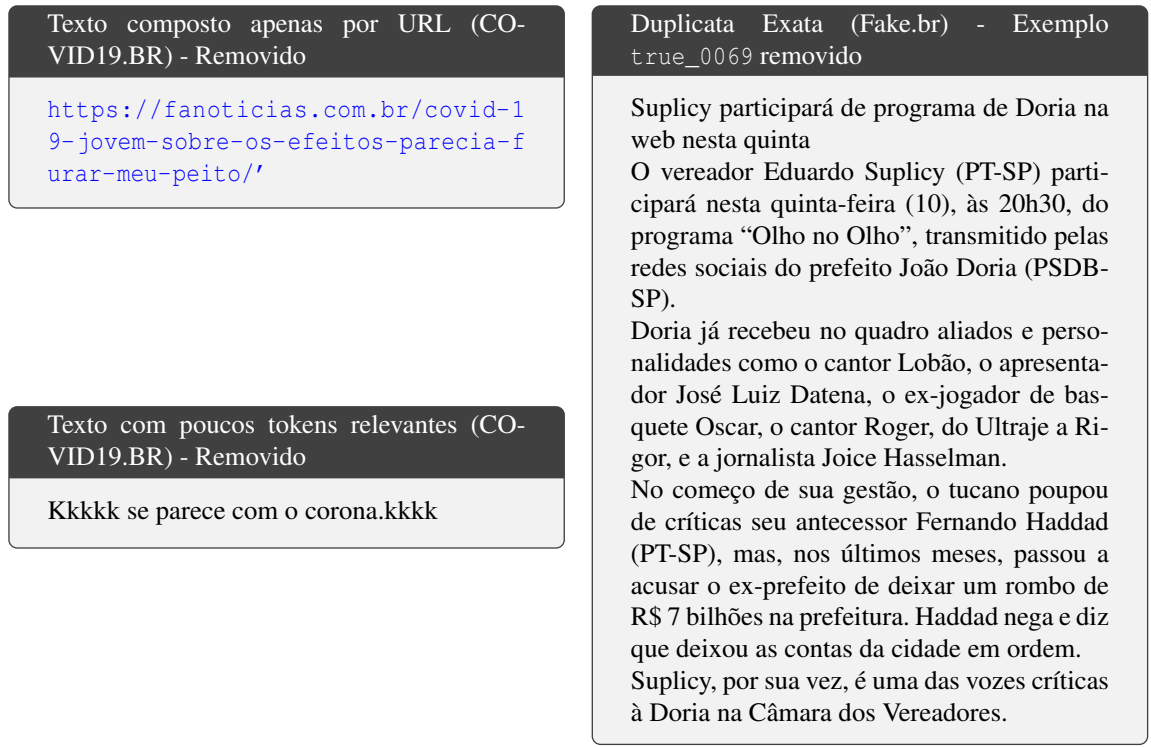

\begin{multicols}{2}
\centering
\footnotesize
\begin{tcolorbox}[colback=black!5, boxrule=0.5pt, arc=1mm, title=Texto composto apenas por URL (COVID19.BR) - Removido]
\footnotesize
\url{https://fanoticias.com.br/covid-19-jovem-sobre-os-efeitos-parecia-furar-meu-peito/'}
\end{tcolorbox}
\begin{tcolorbox}[colback=black!5, boxrule=0.5pt, arc=1mm, title=Texto com poucos tokens relevantes (COVID19.BR) - Removido]
\footnotesize
Kkkkk se parece com o corona.kkkk
\end{tcolorbox}
\columnbreak
\begin{tcolorbox}[colback=black!5, boxrule=0.5pt, arc=1mm, title=Duplicata Exata (Fake.br) - Exemplo \texttt{true\_0069} removido]
\footnotesize
Suplicy participará de programa de Doria na web nesta quinta

O vereador Eduardo Suplicy (PT-SP) participará nesta quinta-feira (10), às 20h30, do programa "Olho no Olho", transmitido pelas redes sociais do prefeito João Doria (PSDB-SP).

Doria já recebeu no quadro aliados e personalidades como o cantor Lobão, o apresentador José Luiz Datena, o ex-jogador de basquete Oscar, o cantor Roger, do Ultraje a Rigor, e a jornalista Joice Hasselman.

No começo de sua gestão, o tucano poupou de críticas seu antecessor Fernando Haddad (PT-SP), mas, nos últimos meses, passou a acusar o ex-prefeito de deixar um rombo de R\$ 7 bilhões na prefeitura. Haddad nega e diz que deixou as contas da cidade em ordem.

Suplicy, por sua vez, é uma das vozes críticas à Doria na Câmara dos Vereadores.
\end{tcolorbox}
\end{multicols}
    \caption{\novo{Exemplos de remoções realizadas na etapa de filtragem inicial automatizada.}}
    \label{fig:append_filter_auto}
\end{figure}

\novo{Esta etapa removeu exemplos que não atendiam a critérios básicos de conteúdo ou formato.} A Figura \ref{fig:append_filter_auto} ilustra três tipos de remoção nesta fase: textos compostos unicamente por uma URL, textos considerados excessivamente curtos após tokenização e remoção de \textit{stopwords}, e duplicatas exatas dentro do mesmo \textit{corpus}.

\section{Filtragem de Idioma}
\label{append:sec:lang_filter}

\novo{Exemplos identificados como não estando primariamente em português foram removidos.} A ferramenta Lingua foi usada para detecção automática, seguida de verificação manual em casos ambíguos. A Figura \ref{fig:append_filter_lang} mostra exemplos removidos por estarem em inglês e espanhol.

\begin{figure}[ht]
\begin{multicols}{2}
\centering
\footnotesize
\begin{tcolorbox}[colback=black!5, boxrule=0.5pt, arc=1mm, title=Exemplo em Inglês (COVID19.BR) - Removido]
\footnotesize
German state minister kills himself as coronavirus hits economy\\ - \url{https://www.aljazeera.com/news/2020/03/german-state-minister-kills-coronavirus-hits-economy-200329165242615.html}
\end{tcolorbox}
\columnbreak
\begin{tcolorbox}[colback=black!5, boxrule=0.5pt, arc=1mm, title=Exemplo em Espanhol (MuMiN) - Removido]
\footnotesize
Vídeo de fraude en las urnas de Flint (Michigan) durante las elecciones de Estados Unidos
\end{tcolorbox}
\end{multicols}
    \caption{Exemplos filtrados por não estarem em português.}
    \label{fig:append_filter_lang}
\end{figure}

\section{Resolução de contradições}

\novo{Esta etapa abordou inconsistências de rótulos entre exemplos semanticamente muito similares ou que referenciavam a mesma fonte.} A Figura \ref{fig:append_duplicate_contradicts} ilustra um caso de textos quase idênticos no \textit{corpus} COVID19.BR que possuíam rótulos de veracidade conflitantes (verdadeiro vs. falso). \novo{Após análise manual, baseada no contexto e fontes externas quando disponíveis, um dos rótulos foi corrigido ou, em casos irresolúveis, ambos os exemplos foram removidos para garantir a consistência.}

\begin{figure}
\begin{multicols}{2}
\centering
\footnotesize
\begin{tcolorbox}[colback=black!5, boxrule=0.5pt, arc=1mm, title=Texto Quase-Duplicado A (COVID19.BR) - Rótulo Original: Falso]
\footnotesize
Em meio à grave crise do CORONAVÍRUS, o Congresso Nacional se recusou a abrir mão do dinheiro destinado ao fundo eleitoral para auxílio ao combate da pandemia. Como sempre esses sangue-sugas demonstram se importar apenas com os próprios interesses. Vamos assinar a petição pelo fechamento do Congresso Nacional!\\ \colorbox{blue}{\scalerel*{\includesvg{figures/pointing-right.svg}}{\textrm{\textbigcircle}}} \url{https://peticaopublica.org/fechamento-congresso-nacional/}
\end{tcolorbox}

\columnbreak 

\begin{tcolorbox}[colback=black!5, boxrule=0.5pt, arc=1mm, title=Trecho dado como verdadeiro (COVID19.BR)]
\footnotesize
Em meio à grave crise do CORONAVÍRUS, o Congresso Nacional se recusou a abrir mão do dinheiro destinado ao fundo eleitoral para auxílio ao combate da pandemia. Como sempre esses sangue-sugas demonstram se importar apenas com os próprios interesses. Vamos assinar a petição pelo fechamento do Congresso Nacional! \\
\colorbox{blue}{\phantom{....................................................................}}\url{https://peticaopublica.org/fechamento-congresso-nacional/}
\end{tcolorbox}

\end{multicols}

\caption{\novo{Exemplo de par de textos quase duplicados no \textit{corpus} COVID19.BR que apresentavam rótulos de veracidade contraditórios. A resolução manual foi necessária para determinar o rótulo correto ou remover o par.}}
\label{fig:append_duplicate_contradicts}
\end{figure}

\section{Verificação Externa de Rótulos}
\label{append:sec:external_verification}

\novo{Utilizou-se a API do Google FactCheck para identificar potenciais erros de rotulação nos dados originais.} Quando a verificação externa sugeria um rótulo diferente do original, o exemplo era revisado manualmente. A Figura \ref{fig:append_change_label} demonstra um caso no \textit{corpus} MuMiN-PT onde o rótulo original (`true`) foi corrigido para `fake` após a verificação externa e análise manual confirmarem a incorreção inicial.

\begin{figure}[ht!]
\centering
\begin{tcolorbox}[colback=black!1, sharp corners, boxrule=0.5pt, title=Exemplo de Correção de Rótulo (MuMiN-PT)]
\footnotesize
\begin{description}[leftmargin=0.5cm, style=sameline, itemsep=1pt]
\item[Corpus:] MuMiN-PT
\item[Texto original:] \ul{A Associação Médica Americana retirou restrições e passou a recomendar a hidroxicloroquina contra a covid-19}
\item[Rótulo:] \st{\texttt{true}} \texttt{fake}
\item[Consulta (Alegação extraída):] A Associação Médica Americana recomenda a hidroxicloroquina contra a covid-19.
\item[Resultado\phantom{ }do\phantom{ }CSE:] 
\phantom{ }\par
\begin{tcolorbox}[colback=black!0.5, sharp corners, boxrule=0.2pt, nobeforeafter]
\begin{description}[style=sameline, itemsep=0pt]
\item[Título:] \texttt{<b>Hidroxicloroquina</b> não é recomendada como tratamento precoce ...}
\item[Trecho:] \texttt{1 day ago <b>...</b> De tempos em tempos, o medicamento <b>hidroxicloroquina</b> volta a ser apontado como tratamento precoce eficaz <b>contra a Covid</b>-<b>19</b>.}
\end{description}
\end{tcolorbox}
\item[Resultado\phantom{ }do\phantom{ }FactCheck:]
\phantom{ }\par
\begin{tcolorbox}[colback=black!0.5, sharp corners, boxrule=0.2pt, nobeforeafter]
\footnotesize
\textbf{Agência:} Aos Fatos | \textbf{Rótulo:} Falso | \textbf{Alegação revisada:} Associação Médica Americana não recomenda hidroxicloroquina ...'
\end{tcolorbox}
\end{description}
\end{tcolorbox}
\caption{\novo{Ilustração do processo de verificação externa de rótulos. Um exemplo do MuMiN-PT teve seu rótulo original (\texttt{true}) corrigido para \texttt{fake} com base em evidências da API do Google FactCheck e subsequente confirmação manual.}}
\label{fig:append_change_label}
\end{figure}

\section{Tratamento Específico ao Fake.br}

\novo{O \textit{corpus} Fake.br possui características particulares que demandaram etapas de tratamento adicionais.} A Figura \ref{fig:append_filter_fakebr} ilustra a remoção de textos quase duplicados que compartilhavam a mesma URL de origem, uma situação específica deste \textit{corpus}. Além disso, outras duas regras específicas foram aplicadas (não ilustradas com exemplos visuais detalhados por brevidade, mas descritas abaixo):

\begin{enumerate}[noitemsep, topsep=2pt, label=(\alph*)]
    \item \textbf{Remoção de textos incompletos:} Exemplos cujos textos normalizados (conforme \ref{subsec:eda_main}) foram identificados como truncados ou incompletos em relação à notícia original (abordado na Issue \#7 do repositório do Fake.br) foram removidos.
    \item \textbf{Manutenção de pares:} Dado que o Fake.br é estruturado em pares de notícias (verdadeira e falsa sobre o mesmo evento), se um dos elementos do par fosse removido por qualquer um dos critérios de validação anteriores, o elemento correspondente também era removido para preservar a integridade da estrutura pareada do conjunto de dados.
\end{enumerate}

\begin{figure}[ht]
\begin{tcolorbox}[colback=black!5, boxrule=0.5pt, arc=1mm, title=Quase-Duplicatas com Mesma URL de Origem (Fake.br) - Exemplo \texttt{true\_3023} removido]

\footnotesize
\ul{O paraíso dos infratores Para conseguir um terço dos votos dos deputados, Temer decreta perdão de multas} Os projetos de privatização para melhorar o desempenho das contas públicas estão sendo golpeados no processo de conquista de um terço dos votos da Câmara para livrar o presidente Temer de responder a investigação por crimes de organização criminosa e obstrução da Justiça. Mais agora, depois que foi atendido o condenado Waldemar da Costa Neto, vulgo Boy, que exigiu a retirada do aeroporto de Congonhas do pacote anunciado, que traria 6 bilhões de renda, em troca de manter o segundo maior aeroporto do País sob controle da Infraero. Está na cara que o esquema de corrupção dos governos anteriores é mantido. \ldots
\end{tcolorbox}
    \caption{\novo{Exemplo de remoção específica do Fake.br: textos quase duplicados (\texttt{true\_0251} e \texttt{true\_3023}) originados da mesma URL fonte. Um deles (\texttt{true\_3023}) foi removido para reduzir redundância originada da coleta.}}
    \label{fig:append_filter_fakebr}
\end{figure}

\chapter{Exemplos pós-expansão dos dados}
\label{append:1}

\novo{Neste apêndice, apresentam-se exemplos ilustrativos dos conjuntos de dados após o processo de enriquecimento descrito na Seção \ref{sec:expanded}. Cada exemplo inclui metadados do conjunto de dados original (\textit{Corpus}, rótulo, texto original com a parte da consulta sublinhada), a consulta utilizada para a busca (seja o texto sublinhado ou uma alegação extraída), o trecho principal do resultado do CSE e do Google FactCheck obtidos.}

\section{Correspondência: sem extração de alegação}
\label{sec:case1}

\novo{O primeiro cenário ilustra casos em que a busca inicial pelo texto original encontrou um resultado com alta correspondência (maior que 0,8), dispensando a extração de alegação. Os casos são organizados conforme os padrões identificados na análise qualitativa (Seção \ref{sec:qualitative_analysis}).}

\begin{figure}[ht!]
\centering
\begin{tcolorbox}[colback=black!1, sharp corners, boxrule=0.5pt, title=Exemplo com correspondência direta (Padrão F1)]
\footnotesize
\begin{description}[leftmargin=0.5cm, style=sameline, itemsep=1pt]
\item[\textit{Corpus}:] MuMiN-PT
\item[Texto original:] \ul{O novo coronavírus pode ser transmitido através de encomendas enviadas da China}
\item[Rótulo] \texttt{fake}
\item[Consulta (Texto original)] O novo coronavírus pode ser transmitido através de encomendas enviadas da China
\item[Resultado\phantom{ }do\phantom{ }CSE:] 
\phantom{ }\par
\begin{tcolorbox}[colback=black!0.5, sharp corners, boxrule=0.2pt, nobeforeafter]
\begin{description}[style=sameline, itemsep=0pt]
\item[Título:] \texttt{<b>O novo coronavírus</b> não sobrevive em <b>encomendas enviadas</b> pelo ...}
\item[Trecho:] \texttt{4 de mar. de 2020 <b>...</b> ... <b>coronavírus pode ser transmitido através de encomendas enviadas</b> pelo correio da <b>China</b> para outros países. A alegação, que serviu de base \&nbsp;...}
\end{description}
\end{tcolorbox}
\item[Resultado do FactCheck:] \texttt{None}
\end{description}
\end{tcolorbox}
    \caption{\novo{Exemplo de busca direta satisfatória (Cenário 1). A consulta encontra um resultado que refuta diretamente a alegação, sem necessidade de extração de alegação. Ilustra o Padrão F1 (Refutação Direta) onde o próprio título do resultado já indica a refutação (``não sobrevive''). O resultado origina-se do MuMiN-PT e demonstra a eficácia do sistema em localizar verificações relevantes quando a consulta contém termos precisos.}} 
    \label{fig:exemplo1}
\end{figure}

\begin{figure}[ht!]
\centering
\begin{tcolorbox}[colback=black!1, sharp corners, boxrule=0.5pt, title=Reforço da \textit{fake news} no padrão F2]
\footnotesize
\begin{description}[leftmargin=0.5cm, style=sameline, itemsep=1pt]
\item[Corpus:] Fake.br
\item[Texto original:] \ul{Governo Temer irá acabar com a farmácia popular}. Quem nunca ouviu ou até mesmo precisou da farmácia popular, não é mesmo? O sistema oferece a distribuição de remédios gratuitos para diversos tipos de doenças, esses recursos estão disponíveis até o mês de agosto de 2017 (truncado)
\item[Rótulo:] \texttt{fake}
\item[Consulta (Texto original):] Governo Temer irá acabar com a farmácia popular.
\item[Resultado\phantom{ }do\phantom{ }CSE:] 
\phantom{ }\par
\begin{tcolorbox}[colback=black!0.5, sharp corners, boxrule=0.2pt, nobeforeafter]
\begin{description}[style=sameline, itemsep=0pt]
\item[Título:] \texttt{<b>Governo Temer</b> fecha <b>Farmácia Popular</b> e pretende extinguir o SUS ...}
\item[Trecho:] \texttt{Feb 22, 2018 <b>...</b> Você <b>vai</b> deixar? Notícias \&middot; ascom ... O Ministério da Saúde <b>acaba</b> de fechar as 517 <b>farmácias</b> populares mantidas pelo <b>governo</b> federal no país.}
\end{description}
\end{tcolorbox}
\item[Resultado\phantom{ }do\phantom{ }FactCheck:]
\phantom{ }\par
\begin{tcolorbox}[colback=black!0.5, sharp corners, boxrule=0.2pt, nobeforeafter]
\footnotesize
\textbf{Agência:} Lupa - UOL | \textbf{Rótulo:} Falso | \textbf{Alegação revisada:} Temer não 'oficializa fim do projeto Farmácia Popular'
\end{tcolorbox}
\end{description}
\end{tcolorbox}

\caption{\novo{Exemplo ilustrando o Padrão F2 (Reforço da \textit{fake news}). A busca do CSE retornou um resultado que reforça a alegação falsa (``Governo Temer fecha Farmácia Popular''), amplificando a desinformação em vez de corrigi-la. No entanto, o Google FactCheck conseguiu localizar uma verificação relevante que refuta a alegação. Este caso demonstra a importância da integração de múltiplas fontes de verificação, pois o resultado do CSE isoladamente poderia levar a conclusões equivocadas.}}
\label{fig:f2}
\end{figure}

\section{Com extração de alegação}
\label{sec:case2}

\novo{Este cenário abrange os casos em que a busca inicial não encontrou resultados com correspondência suficiente (`match < 0.8` para todos os resultados). Nesses casos, uma alegação (`claim`) foi extraída via LLM, e uma segunda busca foi realizada com essa alegação. Apresentam-se exemplos onde a segunda busca retornou resultados considerados relevantes (Subseção \ref{subsec:case2.1_appendix}) e onde os resultados não foram considerados diretamente relevantes para a verificação (Subseção \ref{subsec:case2.2_appendix}).}

\subsection{Resultados relevantes após extração}
\label{subsec:case2.1_appendix}

\novo{O exemplo a seguir ilustra um caso onde a busca inicial falhou, a alegação foi extraída com sucesso, e a busca com a alegação extraída retornou resultados potencialmente úteis para verificação. Este padrão demonstra o valor da extração de alegações como etapa intermediária no processo de enriquecimento.}

\begin{table}[H]
\centering
\footnotesize
\begin{tcolorbox}[colback=black!1, sharp corners, boxrule=0.5pt, title=Exemplo com extração e resultados relevantes]
\begin{description}[leftmargin=0.5cm, style=sameline, itemsep=1pt]
\item[\textit{Corpus}] COVID19.BR
\item[Texto original] Gente ja tivemos pacientes que fizeram exame de COVID-19 , inclusive com exame feito no sabin e tendo dado negativo e repetiu aqui no Hospital Oswaldo Cruz no teste rápido e deu positivo, lembre-se o teste de swab nasal tem chance de dar negativo. Fiquem atentos, pra não estar com um falso negativo e contaminando outros. Olho se tiver sintomas não descuide pois alguns exames são falsos negativos.
\item[Rótulo] \texttt{true}
\item[Consulta (Alegação extraída)] Exames de COVID-19 podem apresentar falsos negativos, mesmo com testes em laboratórios renomados.
\item[Resultado\phantom{ }do\phantom{ }CSE:] 
\phantom{ }\par
\begin{tcolorbox}[colback=black!0.5, sharp corners, boxrule=0.2pt, nobeforeafter]
\begin{description}[style=sameline, itemsep=0pt]
\item[Match:] 0.875
\item[Título:] \texttt{<b>Covid</b>-<b>19</b>: Pesquisador da Fiocruz tira dúvidas sobre <b>testes</b> de ...}
\item[Trecho:] \texttt{Jan 15, 2021 <b>...</b> ... <b>falsos negativos podem</b> se dever à baixa especificidade e baixa sensibilidade analítica do <b>teste</b>.}
\end{description}
\end{tcolorbox}
\item[Resultado do FactCheck] \texttt{None}
\end{description}
\end{tcolorbox}
     \caption{\novo{Exemplo com extração de alegação e resultados relevantes na segunda busca (Subcenário 2.1). A busca inicial falhou em encontrar correspondência direta com o texto original (que continha detalhes específicos de relato pessoal), mas o LLM conseguiu extrair a alegação central de forma concisa e precisa. A busca com a alegação extraída encontrou informações corroborativas de fontes confiáveis que confirmam a possibilidade de falsos negativos em testes de COVID-19. Este exemplo ilustra o Padrão V1 (Corroboração) e demonstra como a extração de alegação pode simplificar textos complexos ou anedóticos para facilitar a verificação.}}
    \label{fig:exemplo2}
\end{table}

\pagebreak

\subsection{Resultados não relevantes após extração}
\label{subsec:case2.2_appendix}

\novo{O exemplo a seguir demonstra um caso onde, mesmo após a extração bem-sucedida de uma alegação, a busca não retornou resultados diretamente relevantes para a verificação da alegação específica. Este cenário representa um desafio persistente para sistemas automatizados de verificação de fatos.}

\begin{table}[H]
\centering
\footnotesize
\begin{tcolorbox}[colback=black!1, sharp corners, boxrule=0.5pt, title=Exemplo com extração e sem resultados relevantes.]
\begin{description}[leftmargin=0.5cm, style=sameline, itemsep=1pt]
\item[\textit{Corpus}] MuMiN-PT \\
\item[Texto Original] \ul{Pessoal, todo mundo precisa se cadastrar no conectesus para vacinar.} Sugiro fazer já. Provavelmente o site nao aguentará os acessos quando for o momento. \url{https://conectesus-paciente.saude.gov.br/} É um cadastro no SUS. Quem tomou a vacina da Febre Amarela em 2018 já tem. Ou quem usou o SUS nos últimos anos. O aplicativo funciona mais ou menos como esses apps de carteira de motorista ou título de eleitor \\
\item[Rótulo] \texttt{fake} \\
\item[Consulta (Alegação extraída)] Pessoal, todo mundo precisa se cadastrar no conectesus para vacinar. \\
\item[Resultado\phantom{ }do\phantom{ }CSE:] 
\phantom{ }\par
\begin{tcolorbox}[colback=black!0.5, sharp corners, boxrule=0.2pt, nobeforeafter]
\begin{description}[style=sameline, itemsep=0pt]
\item[Título:] \texttt{Obter o Certificado Nacional de <b>Vacinação</b> COVID-19}
\item[Trecho:] \texttt{Os dados já foram enviados, mas possui algum erro de informação. Quais os possíveis erros: CNS duplicado no <b>cadastro</b> do SUS.O cidadão deverá se dirigir a\&nbsp;...}
\end{description}
\end{tcolorbox}
\item[Resultado do FactCheck] \texttt{None}
\end{description}
\end{tcolorbox}
    \caption{\novo{Exemplo de busca que retorna resultados não diretamente relevantes para a alegação específica (Subcenário 2.2). Neste caso, o LLM identificou corretamente a alegação central (obrigatoriedade do cadastro para vacinação), mas a busca retornou apenas informações sobre o certificado de vacinação, sem abordar a questão da obrigatoriedade do cadastro prévio.}}
    \label{tab:final}
\end{table}

\end{document}